\documentclass[lettersize,journal]{IEEEtran}
\usepackage{float}
\usepackage{algorithm}
\usepackage{algorithmic}

\usepackage{bm}
\usepackage{multirow}
\usepackage{amsmath}

\usepackage{amsmath,amsfonts}
\usepackage{algorithmic}
\usepackage{algorithm}
\usepackage{array}
\usepackage[caption=false,font=normalsize,labelfont=sf,textfont=sf]{subfig}
\usepackage{textcomp}
\usepackage{stfloats}
\usepackage{url}
\usepackage{verbatim}
\usepackage{subcaption}
\usepackage{graphicx}
\usepackage{cite}
\usepackage{hyperref}       
\usepackage{url}            
\usepackage{amsmath}
\usepackage{booktabs}       
\usepackage{subcaption}
\usepackage{amsfonts}
\usepackage{amsfonts}       
\usepackage{nicefrac}     
\usepackage{amssymb} 
\usepackage{bm} 
\usepackage{microtype}      
\usepackage{xcolor}         
\usepackage{multirow} 
\usepackage{amsmath, amssymb}
\usepackage{amsthm}
\usepackage{picinpar}
\usepackage{footmisc}
\usepackage{amsfonts}
\usepackage{graphicx}
\usepackage{amssymb}
\usepackage{colortbl}
\usepackage[utf8]{inputenc}
\usepackage[T1]{fontenc}  
\usepackage{sidecap}
\usepackage{array}
\usepackage{wrapfig}
\usepackage{boldline}

\newtheorem{theorem}{Theorem}
\newtheorem{lemma}{Lemma}
\newtheorem{assumption}{Assumption}
\newtheorem{proposition}{Proposition}

\newtheorem{observation}{Observation}

\newenvironment{lemmanum}[1]
  {\renewcommand{\thelemma}{#1}\begin{lemma}}
{\end{lemma}}

\newenvironment{propositionnum}[1]
  {\renewcommand{\theproposition}{#1}\begin{proposition}}
{\end{proposition}}

\newcommand{\dd}{\mathrm{d}}

\begin{document}

\title{Time-reversed Flow Matching with Worst Transport in High-dimensional Latent Space for Image Anomaly Detection}

\author{Liangwei Li, Lin Liu, Hanzhe Liang, Juanxiu Liu, Jing Zhang, Ruqian Hao, Xiaohui Du,\\ Yong Liu~\IEEEmembership{Senior Member, IEEE}, Pan Li~\IEEEmembership{Fellow, IEEE}

\thanks{This work was supported in part by the National Natural Science Foundation of China (Grant No. 62574031) and the Fundamental Research Funds for the Central Universities (ZYGX2025YGLH009, ZYGX2024XJ026). (Corresponding author: Xiaohui Du.}


\thanks{Liangwei Li, Lin Liu, Juanxiu Liu, Jing Zhang, Ruqian Hao, Xiaohui Du, and Yong Liu are with School of OptoElectonic Science and Engineering, University of Electronic Science and Technology of China, Chengdu 610031, China (e-mail: vnlee@std.uestc.edu.cn; \{liulin1979, juanxiul, zhangjing, ruqianh, xiaohuie, yongliu\}@uestc.edu.cn).
}
\thanks{Hanzhe Liang is with the Mohamed bin Zayed University of Artificial Intelligence, Abu Dhabi 11058, UAE, and Shenzhen Audencia Financial Technology Institute, Shenzhen University, Shenzhen 518060, China (e-mail: lianghanzhe2023@email.szu.edu.cn).}
\thanks{Pan Li is with the School of Computer Science, Hangzhou Dianzi University, and Zhejiang Institute of Artificial Intelligence, Hangzhou, China (e-mail: lipan@ieee.org).}
}

\maketitle

\begin{abstract}

Likelihood-based deep generative models have been widely investigated for Image Anomaly Detection (IAD), particularly Normalizing Flows, yet their strict architectural invertibility needs often constrain scalability, particularly in large-scale data regimes. Although time-parameterized Flow Matching (FM) serves as a scalable alternative, it remains computationally challenging in IAD due to the prohibitive costs of Jacobian-trace estimation. This paper proposes time-reversed Flow Matching (rFM), which shifts the objective from exact likelihood computation to evaluating target-domain regularity through density proxy estimation. We uncover two fundamental theoretical bottlenecks in this paradigm: first, the reversed vector field exhibits a non-Lipschitz singularity at the initial temporal boundary, precipitating explosive estimation errors. Second, the concentration of measure in high-dimensional Gaussian manifolds induces structured irregularities, giving rise to a Centripetal Potential Field (CPF) that steers trajectories away from Optimal Transport (OT) paths. We identify these observations as the inherent dualities between FM and rFM. To address these issues, we introduce local Worst Transport Flow matching (WT-Flow), which amplifies the observed CPF of rFM to mitigate the initial singularity while circumventing the need for exact distribution transformations via density proxy. Experiments on five datasets demonstrate that WT-Flow achieves state-of-the-art performance among single-scale flow-based methods, and competitive performance against leading multi-scale approaches. Furthermore, the proposed framework enables superior one-step inference, achieving a per-image flow latency of only 6.7 ms. Our code is available on \href{https://github.com/lil-wayne-0319/FMAD}{GitHub}.

\end{abstract}   


\maketitle

\section{Introduction}
\label{sec:intro}

\begin{figure}[t]
  \centering
  \includegraphics[width=0.95\linewidth]{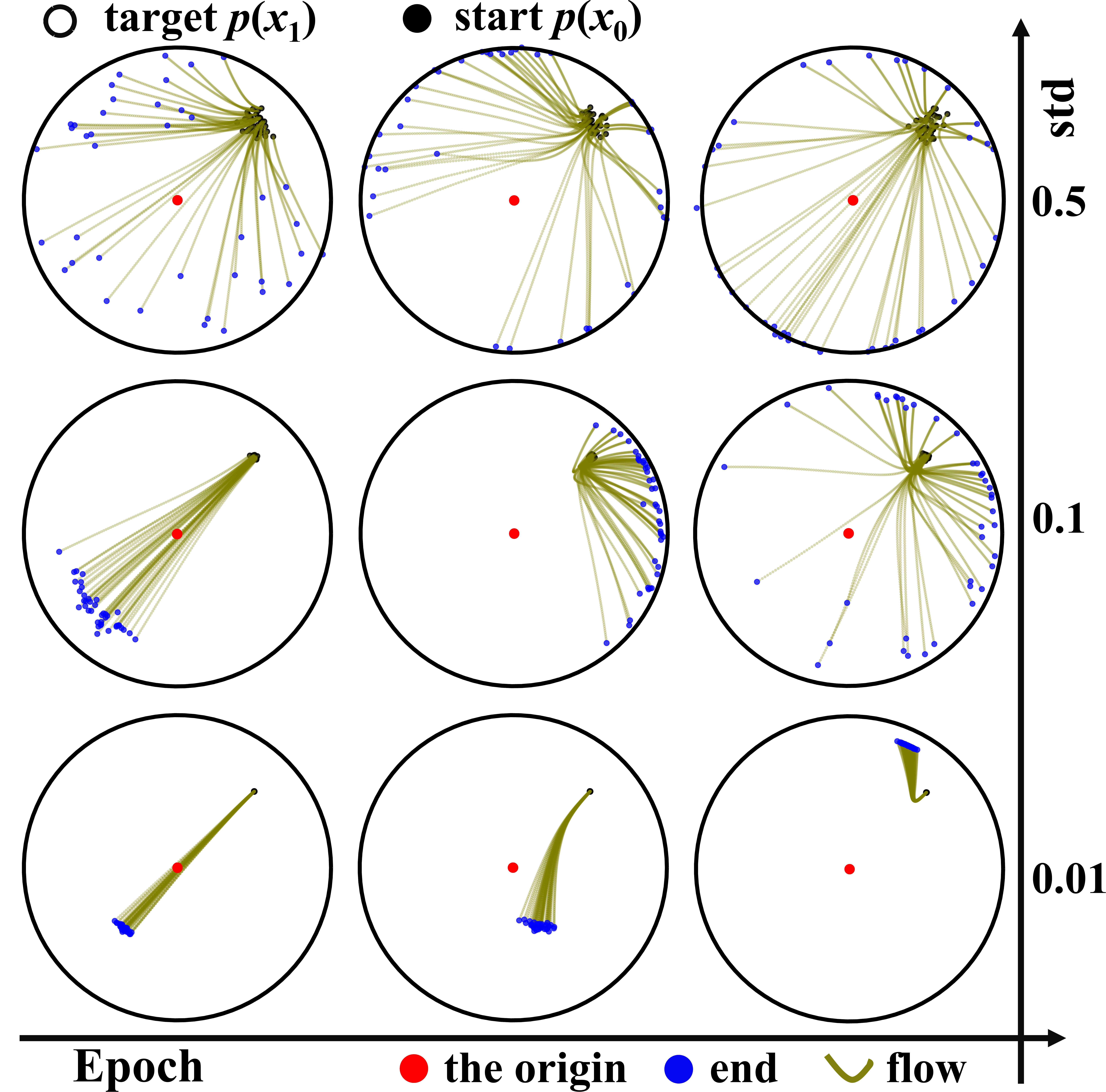}

   \caption{We simulate the described geometry in 2D by training a FM to push the black dots outward onto the black circle (heuristic visualization of high-dimensional Gaussian), which is in reverse to generative modeling. We plot trajectories of 40 initial points $x_0$ sampled from $\mathcal{N}(\boldsymbol{\mu},\sigma^2\mathbf{I})$ with different variances $\sigma$ across various training epochs (400, 4000, 8000). The target distribution is a uniform ring, designed to emulate a high-dimensional Gaussian annulus. Key observations are: first, all points first drift inward to the origin before converging to the ring; second, points closer to its distribution center experience a stronger trapping effect that hinders their reach to the target.}
   \label{fig1}
\end{figure}

Unsupervised image anomaly detection has emerged as a core research direction in industrial computer vision \cite{liu2024deep} and medical image analysis \cite{pham2025silvar}.
Likelihood-based deep generative models (DGMs) represent a principal paradigm for IAD, as they enable intuitive quantification of data regularity via probability density estimation—anomalous samples can naturally be characterized as low-likelihood observations in the learned data distribution.

Recently, FM has shown appealing promise in this field \cite{li2025scalable} due to its sampling efficiency and architectural flexibility compared to Normalizing Flows (NFs) \cite{gudovskiy2022cflow}. However, the application of vanilla FM to IAD is non-trivial, as the FM objective is tailored for generative modeling (noise-to-data). A simple remedy is to reverse the direction of the data flow in FM (data-to-noise), so that we can distinguish anomalies by inferring the corresponding noise density. We have carefully motivated the use of such paradigms in Sec.~\ref{sec:pl}. 

However, our empirical observations and theoretical analyses reveal a counter-intuitive finding formalized in Observation~\ref{ob1}: the rFM fails to push forward the data distribution $P^*(X)$ to the expected standard Gaussian distribution $P(Z)$ in high-dimensional space. The analysis indicates that the FM is inherently non-invertible, or more precisely, it cannot be exactly reversed due to (1) the uncertainty in the initial condition and (2) the emergence of a \textit{centripetal potential field} arising from manifold constraints. These correspond to two dualities in FM: the duality of initial/terminal singularities, and the space-time duality of the \textit{mean-field effect}~\cite{huang2023bridging}. The initial uncertainty stems from the fact that the reversed conditional vector field (VF) for $t \in (0,1]$ is locally Lipschitz continuous. This property exhibits a formal duality with the forward VF defined for $t \in [0,1)$. Unfortunately, the initial singularity in rFM at $t=0$ triggers an explosive accumulation of estimation errors, thereby degrading IAD performance. The centripetal effect, as captured in Observation~\ref{ob2}, originates from a mean-field game along the temporal dimension of the VF, whereas prior literature has primarily focused on its spatial counterpart.

\begin{figure}[t]
\centering
\includegraphics[width=0.99\columnwidth]{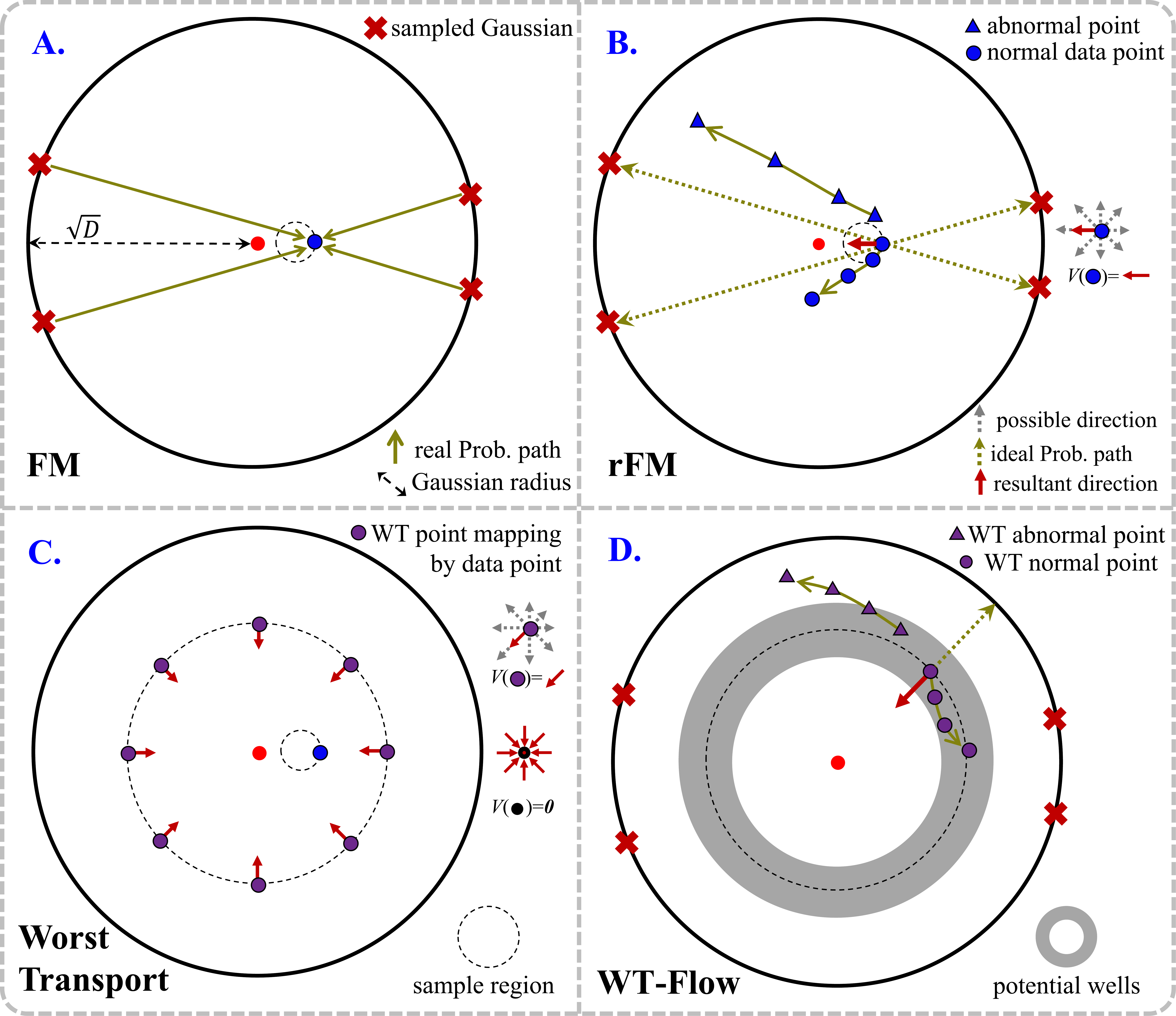} 
\caption{The illustrations of the proposed geometry relationship: (A) deterministic probability paths of FM in generative modeling; (B) the same initial point corresponds to different targets across different ``matching''; (C) data mapped via WT; and (D) sample trajectories after WT map. The gray region represents a potential well, within which anomaly-free samples are confined.}
\label{fig3}
\end{figure}

We describe the geometric intuition in Fig.~\ref{fig3}: upon reversing the flow, the interaction between the low-dimensional data manifold and the measure concentration of the high-dimensional Gaussian \cite{vershynin2018high} forms a unique geometry. The ODE-driven FM possesses the \textit{non-crossing property}~\cite{liuflow2023}, where repeated ``matching'' processes on the same sample average the predicted marginal VF over time, pulling it toward the origin. We term this the CPF, resulting from temporal averaging. We empirically validated this hypothesis on a 2D toy dataset (Fig.~\ref{fig1}) and the high-dimensional MVTec benchmark (Fig.~\ref{fig2}F), with results consistent with our theoretical expectations.

To address these issues, we formalize a density proxy based on the $\ell_2$-norm of the samples, which serves as the anomaly score.
We then amplify the observed CPF and construct the local \textit{worst transport mapping} \cite{lei2021optimal} in the Wasserstein-2 sense. The proposed WT-Flow ensures that the learned trajectories exhibits nearly trivial radial displacement---samples barely flow, as depicted in Fig.~\ref{fig3}(C,D). We evaluated WT-Flow on the challenging MVTec, VisA, MVTec LOCO, MVTec 3D, and MPDD benchmarks, achieving competitive performance with only single-scale inputs. We further report its exceptional single-step inference capability. Our contributions are as follows:

\begin{itemize}   
    \item We prove that FM with linearly interpolated Gaussian paths is not rigorously invertible in theory, and analyze the Lipschitz continuity of the reversed vector field within a truncated time interval.
    
    \item We reveal that rFM exhibits novel dynamics driven by a formalized centripetal potential field. By amplifying this field, we propose WT-Flow, which imposes a stable centripetal action within the vicinity of the normal data manifold and thus guarantees model trainability.
    
    \item We explore FM for unsupervised IAD and formalize an explainable density proxy mechanism to guide rFM for IAD, thereby expanding the design space of traditional NF-based methods. 
    
    \item On 5 image benchmarks, our method achieves state-of-the-art performance among single-scale flow-based models and is competitive with leading multi-scale alternatives.
\end{itemize}

The flow model adopted in this work mainly follows the Rectified Flow framework \cite{liuflow2023}. Relevant reversed flow models have been previously introduced in TCCM \cite{li2025scalable}, achieving promising performance on low-dimensional tabular data while struggling to scale to high-dimensional image domains. The generation quality guarantees of the generative process (noise-to-data) in FM have been fully studied \cite{zhou2025error}. To our knowledge, no work has investigated the stability of the vector field in the reverse process. This paper takes a step forward by considering the Manifold Hypothesis (Assumption \ref{assumption2}) and the Gaussian Annulus Theorem (Lemma \ref{lm:gs}), and presents the Lipschitzness analysis of the reversed vector field under time truncation. The derivations in this section, as well as the detailed proofs of Lemmas \ref{lm:vf}, \ref{lm:ubtvf}, \ref{lm:ubxvf}, and \ref{lm:ln}, are provided in \textit{Suppl. Sec. II.} The key findings of this work are outlined as follows.

\subsection{Observations}

\begin{observation}[Failure of rFM to Reach Gaussian Annulus]
    For the $D$-dimensional standard Gaussian target $\mathcal{N}(\mathbf{0}, \mathbf{I}_D)$, the rFM fails to map the data $x_0$ onto the corresponding Gaussian annulus, where $\lVert x_1 \rVert = \sqrt{D}$, yet, we observe $\lVert x_1 \rVert < \sqrt{D}$.
\label{ob1}
\end{observation}

\begin{observation}[Non-monotonic Radial Trajectory]
    Under a standard Gaussian target distribution, samples in rFM exhibit a non-monotonic radial trajectory during discrete-time inference: the Euclidean norm $\lVert x_t \rVert$ first contracts and then expands.
\label{ob2}
\end{observation}

\subsection{Assumptions}

\begin{assumption}[Bounded Support \cite{zhou2025error}]
    The data distribution $P^*(X)$ is supported on $[0, 1]^D$.
\label{assumption1}
\end{assumption}

\begin{assumption}[Manifold Hypothesis \cite{bengio2013representation}]
    High-dimensional image data inherently lies on a low-dimensional manifold.
\label{assumption2}
\end{assumption}

\begin{lemma}[Gaussian Annulus Theorem \cite{vershynin2018high}]
    For a $D$-dimensional standard Gaussian with unit variance in each direction, for any $\beta \leq \sqrt{D}$, all but at most $3e^{-c^{\beta^2}}$ of the probability mass lies within the annulus $\sqrt{D}-\beta \leq \lVert x \rVert \leq \sqrt{D}+ \beta$, where $c$ is a fixed positive constant.
    \label{lm:gs}
\end{lemma}

\subsection{Main Results}
\begin{theorem}[Initial Singularity]
    The conditional vector field ${u}(x,t|x_0)$ of rFM possesses an initial singularity w.r.t. time variable ${t}$ at $t=0$, which leaves ${u}(x,0|x_0)$ undefined.
    \label{th:singularity}
\end{theorem}

\begin{theorem}[Lipschitzness of rFM]
    If the Assumption \ref{assumption1} hold, then with truncation $t \in [T,1]$, the marginal vector field ${u}(x,t)$ of rFM is jointly $\xi$-Lipschitz continuous w.r.t. $(x,t)$ on $\mathbb{R}^D \times [T,1]$, where $\xi \leq \operatorname{max}\{ \frac{1}{T},  \frac{(1-T)D}{T^3}, \mathcal{O} (\frac{D^2}{T^4})  \} $.
    \label{th:lps-rfm}
\end{theorem}

\begin{proposition}[Centripetal Potential Field]
    If the Assumption \ref{assumption1} and \ref{assumption2} hold, the marginal vector field ${u}(x,t)$ of rFM points toward $\textbf{0}$ at $t=0$, and its magnitude is positively correlated with the sample's initial distance $d= \lVert {x_0} \rVert _2$.
    \label{pp:cpf}
\end{proposition}

\begin{proposition}[Local Worst Transport Map]
    If the Assumption \ref{assumption1} and \ref{assumption2} hold, the initial marginal vector field ${u}(x,0)$ of WT-Flow yields directions at each sample point that are either orthogonal or opposite to the OT direction at that point.
\end{proposition}

The rest of this paper is organized as follows. Sec.~\ref{sec:pl} compares different paradigms of flow-based models. Sec.~\ref{sec:methods} details our theoretical motivation and proposed methodology. Sec.~\ref{sec:Disc} presents the discussion and verification to our methodology. In Sec.~\ref{sec:exp}, extensive experiments are conducted to validate the proposed method. Sec.~\ref{sec:rw} recalls some related work. Finally, Sec.~\ref{sec:limt} and Sec.~\ref{sec:conc} draw the limitations and conclusion.

\section{Prolegomena}
\label{sec:pl}

Likelihood-based DGMs, particularly NFs \cite{dinh2017density}, have shown considerable promise in IAD \cite{gudovskiy2022cflow} due to the exact likelihood computation. Conceptually, terms such as ``rare observations,'' ``low-likelihood samples,'' and ``anomalies'' associated with probability can be quite intuitive. Once DGMs achieve backward sampling, data discrimination becomes feasible by evaluating their regularity. NFs represent one of the rare models that can perform bidirectional invertible transformations while maintaining computational efficiency. By the Change of Variables Formula \cite{dinh2017density}, the likelihood of $x$ is exactly computed:

\begin{equation}
    p_\theta(x)=p_Z(f_\theta(x))|\det(\frac{\partial f_\theta(x)}{\partial x})|,
    \label{eq:cvf1}
\end{equation}

\noindent where $p_Z(z)$ is the the Gaussian probability density function (PDF), $f_\theta(\cdot)$ is the parameterized neural network.

Despite their success, the requirement for strict invertibility often constrains the expressiveness of NFs, particularly in high-dimensional settings \cite{durkan2019neural}. Recent advances in DGMs, including Flow Matchings \cite{lipman2023flow}, Continuous Normalizing Flows (CNFs) \cite{chen2018neural}, Score Matchings (SMs) \cite{songscore}, and Diffusion-based Models (DMs) \cite{ho2020denoising}, have extended the paradigm of distribution transformation (termed indirect push-forward models \cite{salmona2022can}). They model data generation as a continuous-time dynamic process, predicting vector fields (e.g. velocity fields, diffusion noise, or score functions) to satisfy probability flow equations \cite{karras2022elucidating}. By avoiding the direct optimization of likelihood-based objectives, these methods allow manipulating the transformations of probability distributions in a more flexible way.

\begin{figure}[t]
  \centering
  \includegraphics[width=0.95\linewidth]{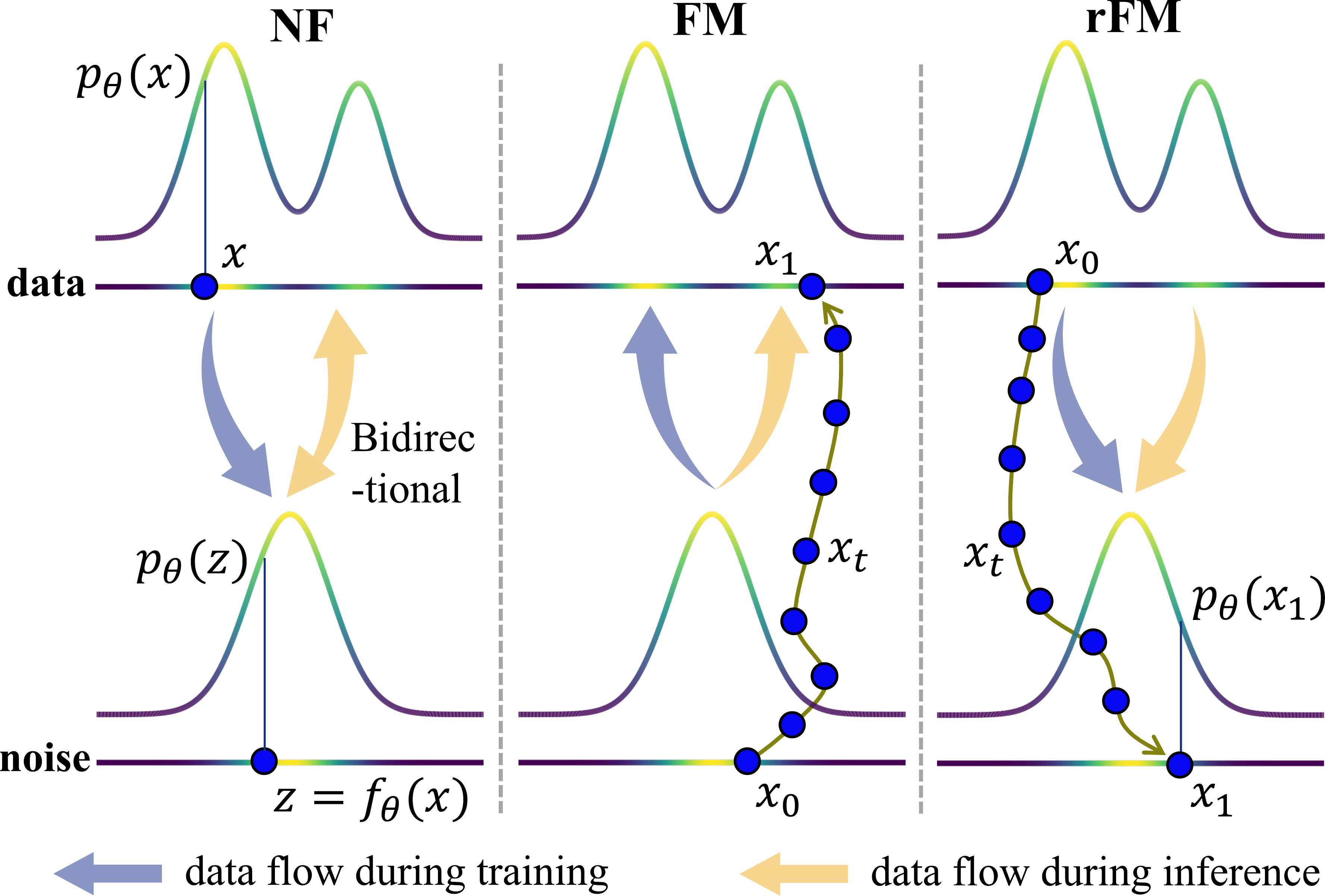}

   \caption{Difference between NF, FM and rFM. NF can directly compute the latent density $p_\theta(z)$ or data likelihood $p_\theta(x)$ via Eq.~\eqref{eq:cvf1}. FM/rFM only support unidirectional distribution transformation. The rFM can compute $p_\theta(x_1)$.}
   \label{img-parad}
\end{figure}

However, obtaining exact likelihoods within such models requires a prohibitively expensive Jacobian-trace estimation \cite{song2021maximum}. More critically, the correlation between low likelihoods and anomalies in high-dimensional space may not be as reliable as previously assumed \cite{kamkari2024geometricexplanation}. On the flip side, despite the lack of a formal explanation, recent NF-based methods \cite{rudolph2022fully, lei2023pyramidflow, zhou2024msflow} implicitly bypass direct likelihood estimation with $p_\theta(x)$, instead identifying anomalies in the latent space using $p_\theta(z)$. We formally term this paradigm density proxy estimation, which permits unidirectional distribution transformations. This perspective eliminates Jacobian-trace needs, enhancing tractability for modern DGMs in anomaly detection. We compare FM, rFM, and NF in Fig. \ref{img-parad}. Notably, although the NF is trained unidirectionally (data-to-noise), its invertible design based on affine coupling layers \cite{dinh2017density} allows for both likelihood estimation and latent sampling during inference.

We trained rFM using density proxy in a manner similar to NF, but observed massive false positive patterns in the predictions.  Further observations reveal that the outputs of the NF typically lie near the Gaussian annulus, whereas the results of the rFM are one to two orders of magnitude smaller than their corresponding Gaussian radius. In addition, we observe that the trajectories generated by the rFM tend to converge toward the origin during the early time steps.

\section{Methodology}
\label{sec:methods}

In this paper, the unknown data distribution $P^*(X)$ under consideration refers to the encoded feature distribution. For clarity in exposition, we refer to the generation procedure ($p(z) \rightarrow p^{*}(x)$) in FM as the forward process, while defining the density proxy estimation  ($p^{*}(x)\rightarrow p(z)$) as the reverse process. The $p_0(x)\rightarrow p_1(x)$ invariably denotes from the start point to the end point, no mater what is $p_0(x)$. The terms differ from some existing literature.

\begin{figure*}[htbp]
\centering
\includegraphics[width=0.99\textwidth]{imgs/main.pdf} 
\caption{(A) The end-to-end pipeline of the proposed WT-Flow. Only the smallest feature map is utilized, without incorporating any external knowledge. (B) Unique geometric relationship between the data manifold and the high-dimensional Gaussian manifold in rFM. (C) Schematic of the data manifold following BatchNorm application. (D) Schematic of the data manifold following LayerNorm application. (E) 2D cross-section illustration of LayerNorm. (F) High-dimensional empirical evidence of the evolution of radial displacement $\lVert x_t \rVert _2$ across 15 MVTec classes. The black dashed line marks the initial sample position. The height of the shade region indicates the magnitude of constrained axial displacement.  Red arrows denote the initial movement of samples toward the origin, while black boxes highlight samples where BN-based WT failed to achieve confinement. Our analysis of these failure cases is provided in the Sec. \ref{sec:Disc}. Quantitative results for the curves are detailed in \textit{Suppl. III-G}.
}
\label{fig2}
\end{figure*}

\subsection{Preliminaries}

\textbf{Density Proxy Estimation:} Let $\mathcal{D}_{\mathrm{tr}} = \{x_{\mathrm{tr}}\} \subset \mathbb{R}^D$ be a normal training set. The objective is to learn a parametric mapping $\phi_\theta: \mathbb{R}^D \to \mathbb{R}^D$, trained exclusively on $\mathcal{D}_{\mathrm{tr}}$, that pushes the normal samples $x_{\mathrm{tr}}$ toward the vicinity of a target distribution support. The anomaly score function is then defined as $\mathcal{A}(x) = f(\phi_\theta(x))$, where $f(\cdot)$ is the corresponding latent density. Specifically, element-wise ``densities'' are aggregated across channels and upsampled to the original resolution. In this paper, we adopt a Gaussian PDF as the target density,

\begin{equation}
    \mathcal{A}(x) = 1-\frac{1}{\sqrt{2\pi}} \exp (-\frac{1}{2}\phi^2_\theta(x)).
\end{equation}

Note that, although Gaussian PDF is used to normalize the output in rFM, this by no means implies that rFM is capable of achieving such a density transformation, due to Observation~\ref{ob1}.

\noindent \textbf{Flow Matching:} Flow Matching characterizes the temporal evolution of a probability distribution $P_t(X)$ through a marginal vector field $u_t(x)$. By parameterizing a model $v_\theta(x_t|x_1)$, FM defines a conditional probability path $p_t(x|x_1)$. As established in \cite{lipman2023flow}, the conditional vector field $u_t(x|x_1)$ ensures that the synthesized marginal vector field $u_t(x)$ strictly adheres to the desired marginal probability path $p_t(x)$:

\begin{equation}
    {u}({x},t)=\int {u}({x}|{x_1})\frac{p_t({x}|{x_1})p^{*}({x_1})}{p_t({x})} \dd {x_1},
\end{equation}

\noindent where ${x_1} \sim p^{*}({x_1})$ represents the unknown real data. 

In particular, ${u}_t({x})$ and $p_t({x})$ satisfy the continuity equation \cite{lipman2023flow}, ensuring local conservation of the probability mass. Provided that ${u}_t({x})$ is smooth and without singularities, the evolution of the probability flow is typically diffeomorphic.

The equivalence between conditional probability paths and marginal probability paths enables the generation of $p_t({x})$ via $p_t({x}|{x_1})$, thereby avoiding the intractable integration required to directly handle the marginal probability path. The optimization objective is formulated as follows \cite{lipman2023flow}:

\begin{equation}
    \mathcal{L}_\text{cfm}(\theta)=\mathbb{E}_{t,p^*(x_1),p(x_0)} \lVert v_{\theta}(\psi(x_0),t)-u_t(\psi_t(x_0)|x_1)\rVert,
    \label{eq:loss-cfm}
\end{equation}

\noindent where ${v}_\theta(\cdot)$ denotes a neural network parameterized by $\theta$ that takes ${x_t}=\psi_t({x_0})$ at time $t \sim \mathcal{U}(0,1)$ as input and predicts the conditional VF ${u}_t({x}|{x_1})$. In the following section, we derive the specific form of conditional VF ${u}_t({x}|{x_0})$ in rFM.

\subsection{Non-Invertibility in Reversed Flow Matching}
\label{sec:sub-nirf}
We use the probability paths defined by Rectified Flow (RF) \cite{liuflow2023}, a special variant of Conditional Flow Matching (CFM) that eliminates the noise term. This formulation enables exact transformations between the data distribution and the Gaussian prior and yields straighter trajectories. In the forward generative process, RF derives the conditional VF $u_t(x|x_1) = x_1 - x_0$. To delineate the theoretical divergences between generative FM and the proposed rFM, we retain the following critical derivation step:

\begin{equation}
 {u}_t({x}|{x_1})=\frac{\sigma_t^\prime({x_1})}{\sigma_t({x_1})}({x}-\mu({x_1}))+\mu_t^\prime({x_1})={x_1}-{x_0},
\end{equation}

\noindent where $\mu_t(x_1)=tx_1$, and $\sigma_t(x_1)=1-t$, with $t\sim \mathcal{U}(0,1)$. 

For the reverse process, the algebraic sign of $x$ is preserved while redefining its semantic interpretation: $x_0\sim p^*(x)$ is anchored to the data distribution, whereas $x_1 \sim \mathcal{N}(\mathbf{0}, \mathbf{I})$ is set as the Gaussian target. Note that the reverse process inverts the endpoint definitions. This generates a reversed conditional Gaussian probability path and the corresponding vector field:

\begin{equation}
 u_t(x|x_0)=\frac{\sigma_t^\prime(x_0)}{\sigma_t(x_0)}(x-\mu_t(x_0))+\mu_t^\prime(x_0),
\label{eq:3.2-5}
\end{equation}

\noindent where $\mu_t(x_0)=(1-t)x_0$, $\sigma_t(x_0)=t$. Under this construction, the conditional vector field is formally derived. Please refer to \textit{Suppl. II-A} for the complete proof. Then, we have

\begin{equation}
u_t(x|x_0)=x_1-x_0,
\label{eq:ut1-0}
\end{equation}

\noindent which defines deterministic trajectories pointing from data samples $x_0$ to Gaussian targets $x_1$. 

Remarkably, a mathematical divergence occurs at $t=0$, where the denominator $\sigma_0(x_0)=t$ in Eq. \eqref{eq:3.2-5} vanishes. This results in a division-by-zero singularity that leaves the initial vector field $u_0(x|x_0)$ undefined. Conversely, the vector field in the forward process remains well defined---FM occurs in division-by-zero singularities only as $t\rightarrow1$, which is sufficient to transport Gaussian samples proximal to the data manifold.
The presence of the singularity at the initial time $t=0$ renders the vector field regression ill-posed at that instant. We then provide a Lipschitzness analysis of the vector field under a truncated time regime $t \in[T,1]$, i.e., for $0 < T \leq t$.

\subsection{Lipschitzness of the Reversed Vector Field}
\label{sec:sub-lrvf}

The initial singularity induces explosive early-time estimation errors, leading to numerous false positives in IAD. Existing forward FMs mitigate this either via smooth approximations \cite{liuflow2023} or the target distribution perturbation to yield truncated-error solutions \cite{lipman2023flow}. However, under the reversed design in this paper, these strategies are either incompatible with our reversed formulation or lack rigorous error guarantees. In this section, we introduce a small truncation threshold $t_{min}=T$ and analyze the regularity of the marginal vector field $u(x,t)$.

The exact minimum of Eq. \eqref{eq:loss-cfm} is analytically achieved by

\begin{equation}
    u(x,t)= \mathbb{E}[X_1-X_0|X_t=x].
    \label{eqee}
\end{equation}

By reformulating this conditional expectation in terms of the score function, we establish the following Lemma \ref{lm:vf}:
    
\begin{lemma}
    The marginal vector field $u(x,t)$ can be written as:
    \begin{equation}
        u(x,t)=-\frac{1}{1-t} x -\frac{t}{1-t} \nabla_x \log p_t(x),
    \label{eq:vf10}
    \end{equation}
    \noindent where $p_t(\cdot)$ is the density of $X_t$, and $X_t = (1-t)X_0 + tX_1$.
    \label{lm:vf}
\end{lemma}

We further derive the upper bounds for the partial derivative of $u(x,t)$ with respect to time $t$ and its spatial gradient with respect to $x$, as summarized in Lemmas \ref{lm:ubtvf} and \ref{lm:ubxvf}.

\begin{lemma}
    The upper bound of the partial derivative of the marginal vector field $u(x,t)$ w.r.t $t \in [T,1]$ is:

    \begin{equation}
        \sup \limits_{t \in [T,1]}  \sup \limits_{x \in [-R,R]^D}  \lVert \partial_t u(x,t) \rVert =\mathcal{O}(\frac{D^2}{T^4}),
    \end{equation}
    \label{lm:ubtvf}
    \noindent where $t_{min}=T$ is the truncation time, and $R \leq \sqrt{D}$.
\end{lemma}

\begin{lemma}
    The upper bound of the gradient of the spatial variable of the marginal vector field $u_t$ w.r.t $x \in [-R,R]^D$ is:

    \begin{equation}
        \sup \limits_{t \in [T,1]}  \sup \limits_{x \in [-R,R]^D} \lVert  \nabla_x u(x,t) \rVert \leq \max \left\{ \frac{1}{T}, \frac{(1-T)D}{T^3}\right\}
    \end{equation}
    \label{lm:ubxvf}
\end{lemma}

Lemmas \ref{lm:ubtvf} and \ref{lm:ubxvf} collectively support Theorem \ref{th:singularity}, confirming that the marginal vector field $u$ is jointly $\xi$-Lipschitz continuous on the domain $\mathbb{R}^D \times [T,1]$. The above conclusion implies that when $T$ is sufficiently small, it becomes extremely difficult for the model $v_\theta$ to estimate the real  vector field. When $T=0$, the Lipschitz condition is not satisfied. Fig. \ref{img-visad} illustrates that the rFM incorrectly classifies normal patterns as anomalies. The proofs for the Lemmas above can be found in \textit{Suppl. II-B}.

\subsection{Local Worst Transport Flow Matching}
\label{sec:sub-lwtfm}

In this section, we present that the initial singular can be mitigated by refining the data manifold. We first elaborate on the centripetal potential field, deriving it via the definition of marginal VF and a geometric perspective, respectively. Next, we clarify a novel anomaly separation mechanism by amplifying the CPF. Finally, we establish the theoretical connection with OT theory. Rigorous mathematical definitions for these specialized terms are provided in \textit{Suppl. II-C}.

\noindent \textbf{Centripetal Potential Field:} We reformulate the CPF as the extension of the marginal VF at $t=0$. The initial conditional VF of rFM is undefined. However, The posterior expectation of the conditional VF, i.e., the marginal VF, admits a well-defined limit at $t=0$. This result can be derived from Eq.~\eqref{eqee} under independent coupling assumption, with

\begin{equation}
    \lim_{t\rightarrow 0} u(x,t)=\mathbb{E}[X_1 \mid X_t = x] - \mathbb{E}[X_0 \mid X_t = x] =-x.
    \label{eq:limu}
\end{equation}


Eq.~\eqref{eq:limu} characterizes the initial centripetal contraction tendency of the marginal VF, motivating us to explore how this behavior emerges. Our analysis is grounded in the properties of both the source (Assumption~\ref{assumption1}) and target (Assumption~\ref{assumption2}), as well as the non-crossing property \cite{liuflow2023} of flow trajectories.

FMs define a non-causal probability path, requiring points to be pre-sampled from the target for matching \cite{liuflow2023}. Once the target is a high-dimensional Gaussian, Lemma \ref{lm:gs} reveals that the $D$-dimensional Gaussian noise is concentrated around a thin spherical shell at radius $\sqrt{D}$ (Gaussian radius $R$). Under the manifold hypothesis in \ref{assumption2} and existing work \cite{popeintrinsic} indicates that the intrinsic dimension of common image datasets typically ranges from ten to several dozen dimensions. We may thus reasonably assume that the data manifold and the Gaussian manifold form the geometric relationship illustrated in Fig. \ref{fig2}B. 

To simplify this geometric intuition, we present a 2D conceptual diagram in Fig. \ref{fig3}B. We consider a uniform ring of radius $R$, and a particle located within this ring. In this setup, each training sample corresponds to different targets across various matching processes. Then, the weighted average over conditional VF funnels these samples into an average resultant direction (red arrow). Symmetry implies that for any point mass at distance $d$ from the origin, the resultant velocity direction is radial. And by some simple curve integrals, we can qualitatively characterize the behavior of such effect:

\begin{equation}
\label{eq:vd}
    V(d)=-\frac{1}{2\pi}\int_{0}^{2\pi} (R\cos\theta-d,R\sin\theta) \mathrm{d}\theta = -d,
\end{equation}

\noindent where $\mathrm{d}\theta=\mathrm{d}s/R$ denotes the angular infinitesimal, and $s$ is the arc length of the ring. It can be shown that $|V(d)|$ is monotonically increasing w.r.t. $d$. 

We interpret $V(d)$ as the temporal mean-field effect over trajectories of an individual sample in different matching process. This establishes a space-time duality with the previously identified spatial mean-field effect, which averages trajectories over different training samples \cite{liuflow2023}. Both phenomena follow the same fundamental principle: ODE-driven methods comply with the non-crossing property, which forbids two distinct solutions from emerging at the same point $(x,t)$.

\noindent \textbf{Degenerate Potential Wells:} 
We note that Eqs.~\eqref{eq:limu} and \eqref{eq:vd} share the same physical nature, i.e. the CPF at $t=0$, whereas $V(d)$ only considers the effect of temporal averaging and ignores spatial influences. This correspondence suggests that the vector field's behavior in the local vicinity of a training instance is primarily governed by the sample's own geometric identity. Eq.~\eqref{eq:vd} further implies that, for a single training data, the closer its initial position is to the sphere, the stronger the influence exerted by $V(d)$. Our key insight is that, despite the initial estimation error in the reversed VF, the disturbance from the initial singularity can be eliminated by properly regularizing and amplifying the observed CPF. 
 
The solution is a simple extension to rFM. We apply a normalization layer to standardize the input, which yields dual effects: mean centering and variance scaling (Fig. \ref{fig2}(C,D)). The former ensures that the expected means of the source and target distributions coincide at the origin; the latter intensifies the centripetal potential $|V(d)|$ by increasing the initial radial distance $d$. Furthermore, the re-scaling brings an additional benefit: training samples are unfolded from a compact structure, so that the neighborhood of a training data is less likely to be disturbed by trajectories of other samples (as it is infeasible to sample a same straight line from one data point to another training data in high-dimensional space).

Note that the CPF is averaged per individual sample across different ``matching''. This temporal effect acts only within the local neighborhood of each sample, i.e., near the train data manifold. We term this property \textit{self-limit}. Thus,  normal samples exhibit stronger energy dissipation in inference and remain confined within a \textit{degenerate potential wells}. In contrast, off-manifold anomalous samples are governed by spatial averaging, enabling explicit separation.

\begin{figure}[t]
\centering
\includegraphics[width=0.99\columnwidth]{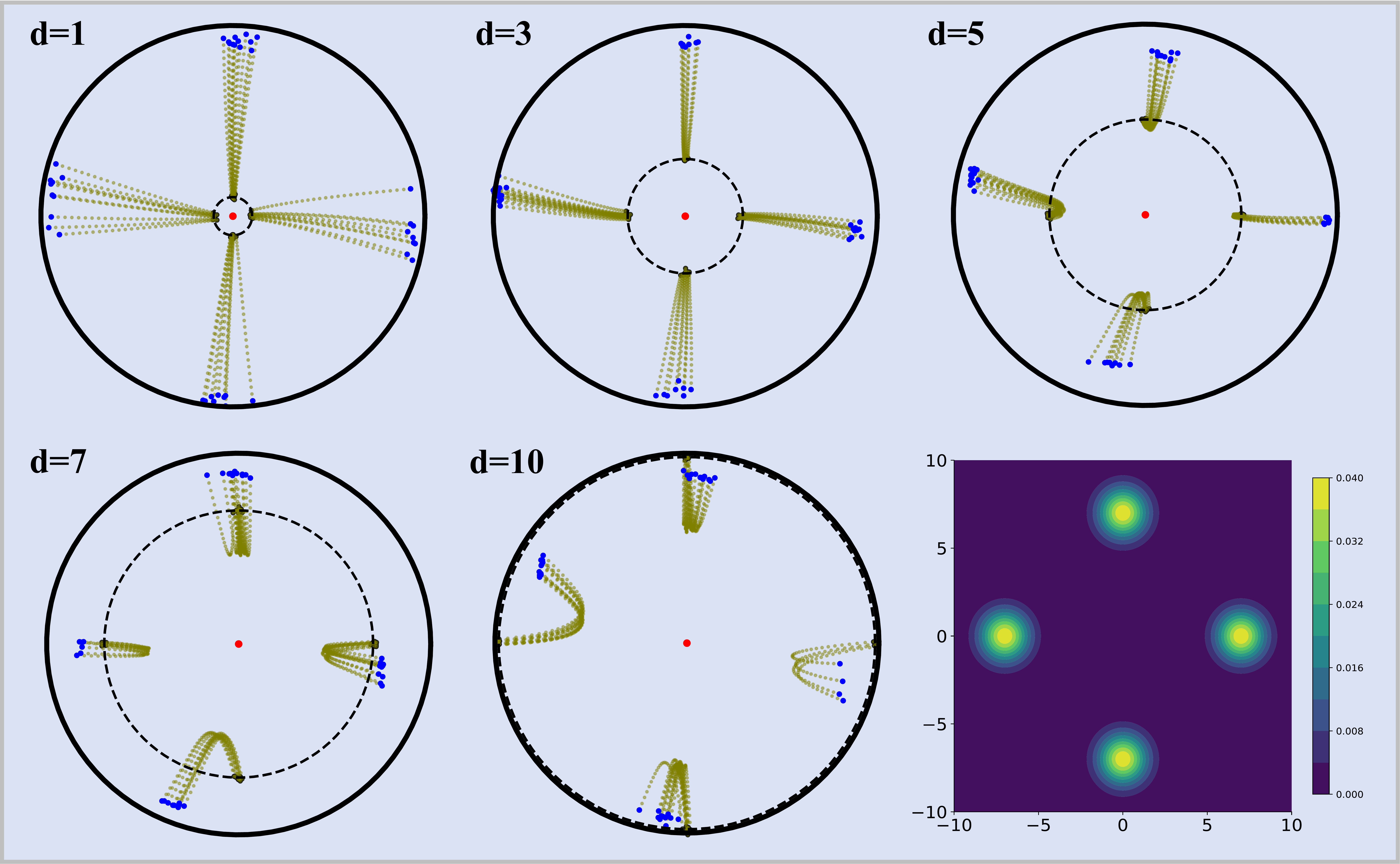} 
\caption{Visualization of the CPF mechanism with varying initial radii $d$. The source is a four-component mixture Gaussian.}
\label{figcpf}
\end{figure}

Fig.~\ref{figcpf} simulates the dynamic evolution of different source distributions under the CPF mechanism in a 2D plane. The target is set as a uniform ring with $R=10$, representing the high-dimensional Gaussian annulus. Considering that data manifolds typically exhibit sparse and structured supports, the source is defined as a symmetric four-component Gaussian mixture ($K=4$) confined within a small circle of radius $d$. This setup intuitively illustrates that $V(d)$ strengthens as $d$ increases, driving stable centripetal contraction of flow trajectories and mitigating early estimation uncertainty through more concentrated initial flow paths.

\begin{lemma}
Assume the Assumptions \ref{assumption1} and \ref{assumption2} hold and the data are non-trivial. Then LayerNorm \cite{ba2016layer} maps the input data $x \in \mathbb{R}^D $ onto a Gaussian annulus with same dimension $D$.

\begin{equation}
        \lVert \text{LN}(x) \rVert _2=R=\sqrt{D}.
    \end{equation}
    \label{lm:ln}
\end{lemma}

\noindent \textbf{Local Worst Transport Map:} While previous research shows that the conditional VF is optimal, this does not necessitate the optimality of the resulting marginal VF \cite{lipman2023flow}. We posit that the marginal probability paths in WT-Flow generate a locally worst-case transport in the Wasserstein-2 sense. To provide geometric intuition, we consider two 2D scenarios based on BatchNorm (BN) \cite{ioffe2015batch} and LayerNorm (LN).

In the BN-based case (Fig. \ref{fig3}C), batch-level statistical coupling often constrains samples within a concentric sub-ring residing inside the Gaussian sphere ($d < \sqrt{D}$), as analyzed in the \textit{Suppl. II-C}. For transport between two concentric measures, the optimal transport is uniquely defined as a radial expansion that shifts mass along the shortest Euclidean distance toward the target boundary. However, the CPF drives samples in the opposite direction toward the origin.

In the LN-based case, samples are projected directly onto the Gaussian annulus ($d = \sqrt{D}$), as formally established in Lemma \ref{lm:ln}. Such a projection induces a geometric coincidence between the source and target supports, as illustrated in Fig.~\ref{fig2}D. This configuration can be idealized as the 2D geometric representation depicted in Fig.~\ref{fig2}E, whose theoretical premises are further elaborated in \textit{Suppl. II-C}. In this regime, the OT map between these identical supports ideally reduces to an identity mapping—or a tangential mass rearrangement along the manifold surface if density differences persist. The transport cost degenerates to zero in the normal direction, confining the OT problem to the tangent space of the manifold. As a result, normal dynamics become dominated by CPF, with the maximum $|V(d)|=|V(R)|$. Conversely, the centripetal gradient remains orthogonal to this tangent space. This geometric misalignment inhibits the intended distribution relaxation and traps anomaly-free samples within degenerate potential wells. Therefore, the learned VF yields directions that are either opposite or orthogonal to the OT direction, establishing a theoretically grounded mechanism for anomaly separation.

\noindent \textbf{U-Net:} In our experiments, we use a simple model as $v_\theta$, a vanilla mini-U-Net with two downsampling and two upsampling layers, paired with a basic Euler ODE solver.

\begin{algorithm}[h]
\small
\caption{\small WT-Flow: Main Algorithm}
\label{alg:algorithm}
\textbf{Input}: Draws from $X_0\sim P^*$ of real data\\
\textbf{Parameter}: vector field model $v_\theta$ with parameter $\theta$\\
\textbf{Output}: predicted endpoint $\hat{X}_1$\\
\textbf{Training}:
\begin{algorithmic}[1] 
\STATE \textbf{Feature Extract}: $X_{0}^{\prime} = \text{encoder}(X_0)$, with $X_0 \in \mathbb{R}^D$
\STATE \textbf{WT mapping}: $X_{0}^{\prime\prime} =  (X_{0}^{\prime} - \mu) / \sigma$ (normalization)
\STATE \textbf{Matching}: a coupling $(X_{0}^{\prime\prime},X_1)$, with $x_1\sim \mathcal{N}(\mathbf{0},\mathbf{I_D})$
\STATE \textbf{Sampling } $X_t=(1-t)X_{0}^{\prime\prime}+tX_1$, with $t\sim\mathcal{U}(0,1)$
\STATE $\hat{\theta} = \arg\min_{\theta} \mathbb{E} [\| X_1 - X_{0}^{\prime\prime} - v_\theta(X_t, t) \|^2 ]$
\STATE \textbf{return} $v_{\hat{\theta}}$
\end{algorithmic}
\textbf{Sampling}: $\hat{X}_1=\text{ODE}(X_{0}^{\prime\prime}, v_{\hat{\theta}} )$\\
\textbf{Anomaly Score}: $\text{score} = \mathcal{A}(\hat{X}_1)$
\end{algorithm}

\section{Discussion and Verification}
\label{sec:Disc}

The inherent irreversibility of rFM arises from the non-Lipschitz continuity of the vector field a consequence of the intricate coupling between the dimensionality $D$ and the time $t$. This lack of regularity precipitates an explosive amplification of estimation errors during the early-time regime, elevating false positive rates as anomaly-free samples are assigned high anomaly scores. While augmenting model capacity may partially alleviate these symptoms by enhancing representational power, it fails to address the underlying regularity issues rooted in the vector field's dynamics. Instead, WT-Flow resolves this bottleneck by amplifying the self-limiting CPF. This mechanism enables the model to bypass early-time instability, ensuring that out-of-distribution anomalies remain excluded from the potential field, thereby establishing a robust separation between normal and anomalous instances.

Empirical analysis of the radial evolution $\lVert x_t \rVert_2$ on the MVTec dataset confirms these dynamics, as shown in Fig. \ref{fig2}F). Trajectories initially exhibit a convergent trend toward the origin. However, as samples deviate from the source distribution, this tendency diminishes, allowing the OT mapping to dominate the subsequent dynamics and redirect mass toward the target. Samples proximal to the data manifold experience intensified constraints and dissipating ``energy,'' which manifests as systematic deviations from the ideal Gaussian. We thus formally propose that the reverse process must integrate both conditional vector field guidance and the structural influence of the manifold topology. Furthermore, BN-based WT demonstrates diminished radial constraints on four specific simple-texture categories (Fig. \ref{fig2}F, black box), where the reduced initial norm $d$ leads to insufficient potential energy dissipation $V(d)$. Finally, We emphasize that although WT is implemented via a simple BN or LN, its design intent and emergent behavior are fundamentally distinct from those of standard normalization schemes. A detailed ablation study is provided in Section \ref{sec5.5}.

\begin{figure}[t]
  \centering
  \includegraphics[width=0.82\linewidth]{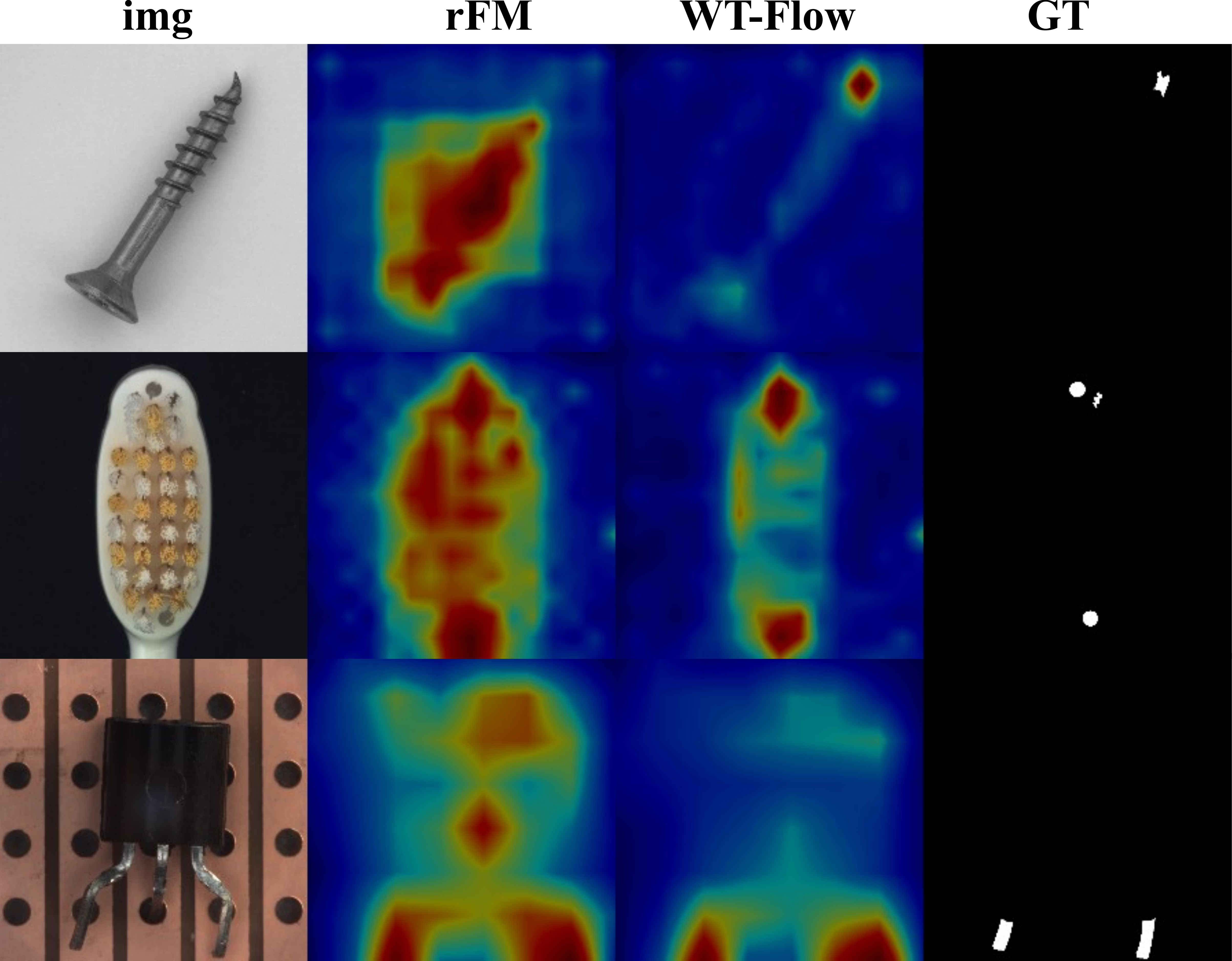}

   \caption{Visualization of activations from rFM and WT-Flow. The estimation error induced by non-Lipschitz results in high false positive rates for rFM}
   \label{img-visad}
\end{figure}

\section{Experiments}
\label{sec:exp}

\subsection{Experimental Setup}

\noindent \textbf{Datasets:} The efficacy of the proposed rFM and WT-Flow is evaluated on five prevalent benchmarks. In alignment with the unsupervised IAD paradigm, training sets comprise exclusively anomaly-free samples, whereas test sets encompass both normal and anomalous instances. Beyond resizing input resolutions, no data augmentation is utilized. We benchmark the performance against established flow-based and non-flow-based methods.

\begin{table*}[t]
\centering
\footnotesize
\setlength{\arrayrulewidth}{0.8pt}
\begin{tabular}{ccccccccc}
    \hline
    \hline
               &DTE-NP\cite{livernochediffusion}    &ADSPR\cite{shin2023anomaly}        &PaDiM\cite{defard2021padim}  &UniAD\cite{you2022unified}  &RD\cite{deng2022anomaly}   &DeSTSeg\cite{zhang2023destseg}         &RD++$^\dagger$\cite{tien2023revisiting}         	    \\
            
    \hline
       Det.    &76.11    &97.67    &97.5  &96.6   &98.46     &98.60  &98.59$_{\pm0.39}$        \\
       Loc.    &-        &97.36    &97.9  &96.6   &97.81     &97.90  &97.81$_{\pm0.17}$          \\
    \hline
    \hline
            &DifferNet\cite{rudolph2021same}  &CSFlow$^\dagger$\cite{rudolph2022fully}  & FastFlow$^\dagger$\cite{yu2021fastflow}         & CFlow$^\ddagger$\cite{gudovskiy2022cflow}   & MSFlow$^\ddagger$\cite{zhou2024msflow}        & rFM$^\ddagger$(ours)  & WT-Flow$^\ddagger$(ours)	    \\
    \hline
       Det. &94.9   &97.06$_{\pm0.26}$    &97.88$_{\pm0.39}$     &95.39$_{\pm0.01}$   &98.50$_{\pm0.20}$  &94.54$_{\pm0.14}$  &\textbf{99.05}$_{\pm\textbf{0.01}}$   \\
       Loc. &-      &-                    &97.68$_{\pm0.21}$     &96.81$_{\pm0.01}$   &97.06$_{\pm0.03}$  &96.55$_{\pm0.04}$  &\textbf{97.95}$_{\pm\textbf{0.01}}$    \\
    \hline
    \hline
\end{tabular}
\caption{Anomaly detection and localization performance on the MVTec AD dataset with mAUC. Methods are grouped into non-flow (top) and flow-based (bottom). Keys: $\dagger$ (evaluated under their multi-scale setting), $\ddagger$ (our single-scale setup).}
\label{tab1}
\end{table*}

\begin{table*}[t]
\centering
\footnotesize
\setlength{\arrayrulewidth}{0.8pt}
\begin{tabular}{ccccccccc}
    \hline
    \hline
            &DTE-NP\cite{livernochediffusion}           &PO-VQVAE\cite{yu2025attention}    &UniAD\cite{you2022unified}      &PNI\cite{bae2023pni}       &CRAD\cite{lee2024continuous}       &RD$^\dagger$\cite{zhang2023destseg}   &RD++$^\dagger$\cite{tien2023revisiting} 	    \\
    \hline
       Det. &83.36   &91.38  &89.0   &95.2      &95.4   &95.53$_{\pm1.35}$      &95.59$_{\pm0.32}$     \\
       Loc. &-       &-      &98.2   &98.8      &98.5   &98.35$_{\pm0.14}$      &98.30$_{\pm0.07}$    \\
    \hline
    \hline
              &DifferNet$^\dagger$\cite{rudolph2021same}   &CSFlow$^\dagger$\cite{rudolph2022fully}       &FastFlow$^\dagger$\cite{yu2021fastflow}       & CFlow$^\ddagger$\cite{gudovskiy2022cflow}   & MSFlow$^\ddagger$\cite{zhou2024msflow}        & rFM$^\ddagger$(ours)  & WT-Flow$^\ddagger$(ours)    \\
    \hline
       Det.   &84.17$_{\pm0.89}$  &93.94$_{\pm0.37}$   &91.52$_{\pm1.39}$   &87.60$_{\pm0.01}$  &95.06$_{\pm0.27}$     &90.62$_{\pm0.21}$  &\textbf{96.52}$_{\pm0.09}$  \\
       Loc.   &-                      &-               &97.49$_{\pm0.68}$   &95.23$_{\pm0.01}$  &97.79$_{\pm0.08}$     &98.10$_{\pm0.04}$  &\textbf{98.90}$_{\pm0.03}$   \\
    \hline
    \hline
\end{tabular}
\caption{Anomaly detection and localization performance on the VisA dataset with mAUC. Methods are grouped into non-flow (top) and flow-based (bottom). Keys: $\dagger$ (evaluated under their multi-scale setting), $\ddagger$ (our single-scale setup).}
\label{tab2}
\end{table*}

\begin{table*}[t]
\centering
\footnotesize
\setlength{\arrayrulewidth}{0.8pt}
\begin{tabular}{cccccccccc}
    \hline
    \hline
            &SPADE\cite{cohen2020sub}           &PaDim\cite{defard2021padim}       &PatchCore\cite{roth2022towards}  &GCAD\cite{bergmann2022beyond} &GRAD\cite{dai2024generating} &EfficientAD$^\dagger$\cite{batzner2024efficientad}     &RD\cite{deng2022anomaly}   &RD++$^\dagger$\cite{tien2023revisiting} 	    \\
    \hline
       Det. &68.8   &68.9    &81.0   &83.3  &87.5  &87.5$_{\pm0.33}$   &79.7       &75.3{$_{\pm0.93}$}     \\
    \hline
    \hline
              &generalAD\cite{strater2024generalad}   &THFR\cite{guo2023template}       &SINBAD\cite{cohen2023set}    &Fastflow$^\dagger$\cite{yu2021fastflow}   & CFlow\cite{gudovskiy2022cflow}   & MSFlow$^\dagger$\cite{zhou2024msflow}        & rFM$^\ddagger$(ours)  & WT-Flow$^\ddagger$(ours)    \\
    \hline
       Det.   &84.9  &86.0   &86.8 &79.7$_{\pm0.86}$  &80.8   &81.6$_{\pm0.83}$     &69.7$_{\pm0.46}$  &\textbf{87.7}$_{\pm\textbf{0.24}}$   \\
    \hline
    \hline
\end{tabular}
\caption{Anomaly detection performance on the MVTec LOCO dataset with mAUC. Keys: $\dagger$ (evaluated under their multi-scale setting), $\ddagger$ (our single-scale setup).}
\label{tabloco}
\end{table*}

\begin{table*}[!t]
\centering
\footnotesize
\setlength{\arrayrulewidth}{0.8pt}
\begin{tabular}{c|c|c|c|c|c|c|c|c|c|c|c|c|c|c}
    \hline
    \hline
    WT map        &\multicolumn{5}{c|}{w/o}   &\multicolumn{3}{c|}{BatchNorm}                     &\multicolumn{6}{c}{LayerNorm}\\     
    
    \hline
    NC          &128    &256  &512  &\multicolumn{2}{c|}{768}   &512  &\multicolumn{2}{c|}{768}   &128    &256  &\multicolumn{2}{c|}{512}  &\multicolumn{2}{c}{768}    \\
    
    \hline
    pos         &\multicolumn{4}{c|}{}   &$\bm{\checkmark}$   &\multicolumn{2}{c|}{}  &$\bm{\checkmark}$  &\multicolumn{3}{c|}{}    &$\bm{\checkmark}$  & &$\bm{\checkmark}$   \\ 
    \hline
    \multicolumn{1}{c}{Det.}       &\multicolumn{1}{c}{70.52}   &\multicolumn{1}{c}{82.11}  &\multicolumn{1}{c}{90.28}  &\multicolumn{1}{c}{92.20}  &\multicolumn{1}{c}{94.54}  &\multicolumn{1}{c}{96.68}  &\multicolumn{1}{c}{97.79}  &\multicolumn{1}{c}{98.57}  &\multicolumn{1}{c}{92.06}  &\multicolumn{1}{c}{94.96}  &\multicolumn{1}{c}{98.36}  &\multicolumn{1}{c}{98.32}  &\multicolumn{1}{c}{99.05}  &\multicolumn{1}{c}{99.04}   \\
    \multicolumn{1}{c}{Loc.}       &\multicolumn{1}{c}{89.85}   &\multicolumn{1}{c}{92.84}  &\multicolumn{1}{c}{95.65}  &\multicolumn{1}{c}{96.71}  &\multicolumn{1}{c}{96.55}  &\multicolumn{1}{c}{97.61}  &\multicolumn{1}{c}{97.63}  &\multicolumn{1}{c}{97.64}  &\multicolumn{1}{c}{96.84}  &\multicolumn{1}{c}{97.48}  &\multicolumn{1}{c}{97.74}  &\multicolumn{1}{c}{97.76}  &\multicolumn{1}{c}{97.95}  &\multicolumn{1}{c}{97.75}   \\

    \hline
    \hline
\end{tabular}
\caption{Ablation study of the proposed WT-Flow on the MVTec AD dataset is presented with image-level and pixel-level mAUC. ``NC'': the number of base channel in U-Net; ``WT map'' and ``pos'': wether to use the WT and positional embedding.}
\label{tab3}
\end{table*}

\noindent \textit{MVTec AD} \cite{bergmann2019mvtec}: A comprehensive real-world industrial dataset for UAD, consisting of over 5,000 high-resolution images across 5 texture and 10 object categories.

\noindent \textit{VisA} \cite{zou2022spot}: A large-scale industrial AD dataset characterized by its complex object structures, consisting of 10,821 high-resolution images (9,621 normal and 1,200 anomalous samples) across 12 objects in 3 domains.

\noindent \textit{MVTec LOCO AD} \cite{bergmann2022beyond}: A specialized dataset targeting logical and structural anomalies, containing 3,644 images from 5 categories. The logical anomalies violate inherent object constraints and structural anomalies denote local defects, with pixel-precise ground truth annotations.

\noindent \textit{MPDD} \cite{jezek2021deep}: A real-world industrial benchmark for defect detection on metallic components under complex acquisition conditions, covering six categories of sprayed metal parts standardized to a 1024 $\times$ 1024 resolution.

\noindent \textit{MVTec 3D AD} \cite{bergmann2022mvtec}: A multi-modal dataset for 3D IAD, providing both 3-channel RGB images and corresponding 3D coordinate data for each sample, with pixel-precise ground truth for anomalous regions. Only the RGB data of this dataset is used in our experiments.

\noindent \textbf{Metrics}: Performance is evaluated using the area under the receiver operating characteristic curve (AUROC) for both image-level anomaly detection (Det.) and pixel-level anomaly localization (Loc.), with the mean AUC (mAUC) across categories as the aggregate metric.

\noindent \textbf{Settings}: An individual Flow model is trained per category in the benchmarks with shared hyperparameters. A pre-trained Wide-ResNet50 \cite{zagoruyko2016wide} serves as feature extractor, utilizing the output from the third residual stage. The input image resolution is standardized to $512 \times 512$. Optimization is conducted using the AdamW with an initial learning rate of $2 \times 10^{-4}$ and a weight decay of $0.1$. The learning rate is further modulated by a step scheduler, decaying by a factor of $0.1$ for every 25 epochs. The training process consists of 100 epochs with a batch size of 8. The number of flow steps in inference is set to 20 for Det. and 1 for Loc.. For the detection task, the anomaly score is computed by averaging the top-$K$ anomaly scores in the anomaly map, with $K$=0.03 \cite{zhou2024msflow}. The WT mapping is implemented via a simple non-learnable LayerNorm. For a limited subset, we fine-tuned the input resolution and model feature size to adapt to more complex environments and targets (refer to \textit{Suppl. III}). All experiments are implemented with PyTorch 1.12 on a single NVIDIA RTX 3090 GPU.

\subsection{Evaluation on MVTec AD}

We compared our rFM and WT-Flow with previous flow-based and non-flow methods on the MVTec AD benchmark using their public implementations. All flow-based methods, except for DifferNet,  CSFlow and FastFlow, are evaluated under a single-scale setup with their reported optimal configurations. As summarized in Table \ref{tab1}, WT-Flow outperforms others in both detection and localization tasks, achieving leading mAUC across all categories of 99.05\% for detection and 97.95\% for localization. Without incorporating any external knowledge, our method surpasses several non-flow approaches that employ multi-scale designs and exhibits superior convergence stability. Our method closely approaches the benchmark results of state-of-the-art multi-scale flow method (99.05\%/97.95\% vs. 99.7\%/98.8\% \cite{zhou2024msflow}). The rationale for the exclusion of multi-scale configurations is further elaborated in Sec. \ref{sec:limt}.

\subsection{Evaluation on VisA}

To evaluate the generalization capability of WT-Flow, we extend our benchmark evaluations to the VisA dataset. As quantified in Table \ref{tab2}, WT-Flow outperforms prominent flow-based competitors, such as MSFlow, by margins of 1.46\% and 1.11\%, respectively. Notably, compared to CSFlow and FastFlow—which rely on default multi-scale architectures—WT-Flow exhibits superior generalization robustness across more complex industrial scenarios. Furthermore, our framework surpasses several multi-scale non-flow methodologies that leverage auxiliary external knowledge, such as RD and RD++. These consistent performance gains, achieved without multi-scale configurations or external priors, underscore the architectural superiority of Flow Matching over traditional Normalizing Flows. This evidence further highlights the intrinsic efficacy of the proposed WT mechanism in capturing diverse and complex anomalous patterns.

\subsection{Evaluation on MVTec LOCO}

We benchmark the efficacy of our method against purely vision-based IAD approaches on the MVTec LOCO AD dataset, which encompasses both structural and logical anomalies (Table \ref{tabloco}). Under the single-scale setting, WT-Flow outperforms previous state-of-the-art methods in anomaly detection. Specifically, our approach substantially surpasses MSFlow by 5.6\% in image-wise mAUC. It also achieves superior performance over methods specifically optimized for logical anomalies, such as EfficientAD, with a 0.2\% improvement. Compared to the rFM without WT module, the proposed WT-Flow achieves a substantial performance boost of 12.4\% (in terms of mAUC). To ensure methodological fairness, we exclude recent text-guided approaches (e.g., LogSAD \cite{zhang2025towards}) from this quantitative comparison, as they leverage additional textual annotations for semantic guidance, which deviates from the purely visual setting of our study.

\subsection{Evaluation on MPDD and MVTec 3D}

MPDD simulates challenging industrial scenarios, including randomized positions and orientations, overlapping instances, uneven illumination, strong reflections, and motion blur. Meanwhile, MVTec 3D covers objects with diverse geometric structures, visual appearances, and physical properties, ranging from textured parts to deformable components, supporting robust evaluation under realistic settings. vanilla rFM nearly fails on both datasets and yields randomized detection performance, achieving only 58.9\% and 57.8\% image-level mAUC, respectively. In contrast, our WT-Flow obtains clear improvements, reaching 94.2\% and 85.1\% image-level mAUC on MPDD and MVTec 3D. Such consistent gains highlight that the designed worst-transport mechanism effectively amplifies inherent centripetal tendencies, alleviates non-Lipschitz error accumulation, and enlarges feature separation between normal and anomalous samples. Moreover, WT-Flow remains competitive compared with existing baseline methods.

\begin{table}[h]
\centering
\footnotesize
\setlength{\arrayrulewidth}{0.8pt}
\begin{tabular}{c c c c c}
    \hline
    \hline
                  &UniAD\cite{you2022unified}    &SimpleNet\cite{liu2023simplenet}     &DeSTSeg\cite{zhang2023destseg}     &RD\cite{zhang2023destseg}                   \\
    \hline
           Det.   &82.2   &90.6  &93.0   &91.4     \\
           Loc.   &95.1   &97.1  &94.1   &97.5    \\
    \hline
                   &DiAD\cite{he2024diffusion}     &MSFlow$^\dagger$\cite{zhou2024msflow}     &rFM$^\ddagger$     &WT-Flow$^\ddagger$                             \\
    \hline
           Det.   &74.6   &92.6   &58.9   &\textbf{94.2}    \\
           Loc.   &93.0   &97.2   &93.3   &\textbf{98.3}    \\
    \hline
    \hline
\end{tabular}
\caption{Anomaly detection performance on the MPDD dataset with mAUC.}
\label{tabmpdd}
\end{table}

\begin{table}[h]
\centering
\footnotesize
\setlength{\arrayrulewidth}{0.8pt}
\begin{tabular}{c c c c c}
    \hline
    \hline
                   &UniAD\cite{you2022unified}    &SimpleNet\cite{liu2023simplenet}      &DeSTSeg\cite{zhang2023destseg}     &RD\cite{zhang2023destseg}                   \\
    \hline
           Det.   &78.9   &72.5  &79.6   &77.9     \\
           Loc.   &96.5   &93.6  &95.1   & 98.4     \\
    \hline
                   &DiAD\cite{he2024diffusion}    &MSFlow$^\dagger$\cite{zhou2024msflow}     &rFM$^\ddagger$     &WT-Flow$^\ddagger$                             \\
    \hline
           Det.   &84.6   &80.9  &57.8  &\textbf{85.1}    \\
           Loc.   &96.4   &96.3  &88.3  &\textbf{98.6}    \\
    \hline
    \hline  
\end{tabular}
\caption{Anomaly detection performance on the MVTec 3D dataset with mAUC.}
\label{tab3d}
\end{table}

\subsection{Ablation Study on WT-Flow}
Table \ref{tab3} summarizes the results of a comprehensive ablative analysis conducted to validate the core design choices of WT-Flow across the MVTec dataset. The investigated components include: (i) implementation variants of the WT map, (ii) the architectural scalability regarding the number of base channels ($NC$) in the U-Net, and (iii) the integration of positional encoding (PE). Our analysis demonstrates that scaling the model capacity from $NC=128$ to $NC=768$ yields a substantial mean gain of 21.7\% in detection and 6.9\% in localization mAUC. This improvement is primarily attributed to the increased model capacity, which effectively mitigates the estimation challenges induced by early-stage non-Lipschitz dynamics, thereby enhancing the approximation fidelity of the learned marginal vector field. The incorporation of the WT map demonstrated consistent improvement in mAUC in all comparative experiments, with particularly notable enhancements observed for the ``screw'' and the ``toothbrush'' categories (see \textit{Suppl.} for detailed results on subclasses). The LN-based WT implementation yields superior performance, reaching 99.05\% and 97.95\% mAUC for Det. and Loc., respectively. The positional encoding shows partial observable contribution to WT-Flow's effectiveness.

\subsection{Ablation Study on Flow Steps}

Table \ref{tab6} reveals a counterintuitive property of WT-Flow. Its single-step inference outperforms its multi-step counterpart. This phenomenon further validates our geometric rationale regarding the CPF at $t=0$. The discriminability originates from early-phase energy dissipation, rendering the single-step vector field worst for transport problem yet optimal for density proxy. This immediate efficacy sidesteps the requirement for computationally expensive likelihood estimation, positioning WT-Flow as an ideal candidate for real-time industrial applications where both accuracy and efficiency are paramount.

\begin{table}[h]
\centering
\footnotesize
\setlength{\arrayrulewidth}{0.8pt}
\begin{tabular}{l c c c c c c}
    \hline
    \hline
      steps         &1      &10     &20     &100  &200      &400   \\
    \hline
    rFM             &86.52  &96.02  &96.19  &95.86  &95.95  &95.90      \\
    WT(BN)          &97.53  &97.49  &97.45  &97.41 &97.38   &97.38     \\
    WT(LN)          &97.95  &97.88  &97.86  &97.86  &97.85  &97.85     \\
    \hline
    \hline
\end{tabular}
\caption{Experimental comparison between rFM and WTflow under different numbers of ODE steps with pixel-mAUC on MVTec.}
\label{tab6}
\end{table}

\subsection{Ablation Study on Different Geometric Design}

In Table~\ref{tabTCCM}, we evaluate three distinct target distributions as priors for rFM and report the performance of WT under each configuration. \textit{Single Point} adopts an origin contraction strategy (e.g., TCCM \cite{li2025scalable}), mapping data to a fixed origin. The \textit{Uniform Sphere} replaces the conventional Gaussian prior in rFM with a uniform distribution on the $\sqrt{D}$-sphere. The \textit{Gaussian} serves as the standard baseline using $\mathcal{N}(\mathbf{0}, \mathbf{I_D})$.

Results reveal that \textit{Single Point} targets induce severe performance degradation in high-dimensional domains. Even with WT regularization applied to the source distribution, a substantial performance gap persists. We attribute this failure to the inherent risk of mode collapse when mapping high-dimensional data onto a zero-dimensional point, which collapses the rich manifold structure of natural images into an overly constrained representation. In contrast, the \textit{Uniform Sphere} provides strong empirical validation for our geometric conjecture. The concentration of measure in high-dimensional Gaussians suggests that a spherical uniform distribution serves as an effective surrogate, offering a more geometrically natural target for rFM.

\begin{table}[h]
\centering
\footnotesize
\setlength{\arrayrulewidth}{0.8pt}
\begin{tabular}{c c c c c c c }
    \hline
    \hline
           Target        &\multicolumn{2}{c}{Single-Point}    &\multicolumn{2}{c}{Uniform-Sphere}    &\multicolumn{2}{c}{Gaussian}\\
    \hline
           WT     &\textbf{$\times$}  &\textbf{$\checkmark$}   &\textbf{$\times$} &\textbf{$\checkmark$}    &\textbf{$\times$}    &\textbf{$\checkmark$} \\
      \hline     
           Det.   &81.01  &84.37   &93.96   &99.02   &94.54  &99.05  \\
           Loc.   &94.64  &95.64   &96.31   &97.87   &96.55  &97.95  \\
    \hline
    \hline
\end{tabular}
\caption{Ablation study on the placement of the normalization layer. ``w/o'': excluding the layer; ``std'': the layer is applied to the input before it is fed into the model}
\label{tabTCCM}
\end{table}

\subsection{Inferencet Time Comparison}
\label{sec:as7}

As reported in Table~\ref{table8}, we employ a WideResNet50 to extract features from fixed resolution $512 \times 512$ inputs, with a latency of 6.5 ms. For CFlow and MSFlow, we set the number of flow blocks to 8. The batch size is 1 for both training and inference.
The latency comparison reveals that conventional discrete approaches are bottlenecked by the requirement of iterative solvers. For instance, rFM-768 requires at least 20 Euler steps to achieve performance comparable to WT-Flow, resulting in an inference latency of 330.1 ms. In contrast, our WT-Flow overcomes this limitation via its novel non-probabilistic modeling, enabling unprecedented one-step inference with only 6.7 ms. This is $49\times$ faster than multi-step methods and $4.7\times$ faster than state-of-the-art normalizing flow models. While the NF-based methods may possess fewer parameters, their invertible architectures lead to higher inference latency and limited scalability. Furthermore, to the best of our knowledge, similar reconstruction-based diffusion methods typically achieve an inference FPS close to 1, lagging behind the efficiency of our proposed method.

\begin{table}[h]
\centering
\footnotesize
\setlength{\arrayrulewidth}{0.8pt}
\begin{tabular}{lrcrrr}
    \hline 
    \hline 
             & Params.   &Flow step &Flow Lat.  & FPS.  &VRAM \\
    \clineB{6-6}{0.8}
             &           &          &           &      &train/test\\
    \hlineB{0.8}           
    CFlow    & 17.1M     &none      &48.6ms     &20.6  &3.2/1.1GB   \\
    MSFlow   & 56.7M     &none      &31.2ms     &26.5  &9.4/3.8GB   \\
    WT-512   & 0.34B     &20        &170.3ms    &5.7   &7.9/2.9GB    \\
    WT-768   & 0.76B     &20        &330.5ms    &3.0   &15.2/6.8GB     \\
    WT-512   & 0.34B     &1         &6.4ms      &77.5  &7.9/2.9GB     \\
    WT-768   & 0.76B     &1         &6.7ms      &75.6  &15.2/6.8GB       \\
    \hline
    \hline 
\end{tabular}
\caption{Comparison of Inference efficiency. ``512/768'' are base channel numbers;``Flow Lat.'' indicates the latency of the flow module. ``Flow Step'' denotes Euler solver iterations. ``FPS'' is calculated by both the encoder and the flow module.}
\label{table8}
\end{table}	

\subsection{Discussion: Is WT Mapping Just a Basic Norm Layer?}
\label{sec5.5}

Table \ref{tab4} reports the convergence loss and the corresponding mAUC under six configurations. The results indicate: (1) WT deliberately introduces optimization resistance (evidenced by higher loss) yet consistently enhances generalization; (2) LN-based WT achieves superior performance with higher loss than BN-based variants; (3) These trends persist across different NC settings. We emphasize that the higher loss is an intentional outcome of our design, which aims to prevent normal samples from reaching the Gaussian hypersphere. Normalized anomaly-free samples are subjected to stronger constraints from the potential field, thereby actively countering the original Flow Matching objective. This stands in stark contrast to the inherent purpose of a BN or a LN, which is to mitigate internal covariate shift and enhance training stability—typically manifesting as reduced convergence loss. It is noteworthy that despite the increased loss, our method achieves highly robust results (see the standard deviation in Table \ref{tab1}).

\begin{table}[h]
\centering
\footnotesize
\setlength{\arrayrulewidth}{0.8pt}
\begin{tabular}{c| c c c| c c c}
    \hline
    \hline
           NC        &w/o   &+BN     &+LN     &w/o   &+BN     &+LN                   \\
    \hline
           512   &0.802   &0.853  &1.011  &90.28   &96.68    &98.36  \\
           768   &0.687   &0.737  &0.906  &92.20   &97.79    &99.05  \\

    \hline
    \hline
\end{tabular}
\caption{Ablation study on the loss of WT map. (Left) Converged loss values of our method under six different settings are compared. The right part provides the corresponding image-wise mAUC on the MVTec dataset}
\label{tab4}
\end{table}

As shown in another ablation study in Table \ref{tab5}, positioning the WT module solely as a standard normalization (std) component  in front of the U-Net input leads to a clear performance drop. This configuration not only deteriorates both detection and localization accuracy but also increases the risk of instability.

\begin{table}[h]
\centering
\footnotesize
\setlength{\arrayrulewidth}{0.8pt}
\begin{tabular}{c c c c c }
    \hline
    \hline
                   &w/o   &+std     &+WT     &+WT\&std                   \\
    \hline
           Det.   &94.54$_{\pm0.14}$   &86.93$_{\pm0.32}$  &99.05$_{\pm0.01}$   &98.10$_{\pm0.02}$      \\
           Loc.   &96.55$_{\pm0.04}$   &93.96$_{\pm0.32}$  &97.95$_{\pm0.01}$   &97.71$_{\pm0.01}$       \\

    \hline
    \hline
\end{tabular}
\caption{Ablation study on the placement of the normalization layer. ``w/o'': excluding the layer; ``std'': the layer is applied to the input before it is fed into the model}
\label{tab5}
\end{table}

\subsection{Qualitative Results}

Figure \ref{fig9} presents a visual comparison of anomaly detection results by rFM, MSFlow, and WT-Flow under single-scale input. All three methods achieve promising recall. However, the precision of rFM and MSFlow degrades on complex categories involving object position shifts, rotation, texture variations, and diverse semantic structures. In contrast, WT-Flow effectively suppresses false-positive patterns in anomaly maps, maintaining robustness while producing fine-grained detection results.

\section{Related Works}
\label{sec:rw}


\noindent \textbf{Flow-based Anomaly Detection}: NFs model the empirical density $p^*(x)$ via a bijective transformation. While CFlow \cite{gudovskiy2022cflow} identified anomalies as low-likelihood ($p_\theta(x)\downarrow$) samples, later methods such as DifferNet \cite{rudolph2021same}, FastFlow \cite{yu2021fastflow}, CSFlow \cite{rudolph2022fully} and MSFlow \cite{zhou2024msflow} instead performed density proxy in the latent space $p_\theta(z)$. Progress in this direction has been limited due to their strict invertibility requirement, which constrains integration with modern architectures.

We frame density proxy estimation as a form of implicit one-class classification. Related approaches \cite{tax2004support, ruff2018deep} employ Support Vector Data Description (SVDD) to construct a spherical boundary around the dataset, outside of which data points are classified as anomalous. DO2HSC \cite{zhangdeep} explores graph-level anomaly detection in high-dimensional spaces. Recognizing that minimizing the sum of squared distances between the data point and a single center may distort the spherical boundary of SVDD. Thus, they propose to refine this boundary via deep orthogonal bi-hypersphere compression. Instead, flow-based density proxy methods aim to push data points toward an ideal hypersphere corresponding to a Gaussian annulus, with any points falling outside this surface being regarded as the anomalous.

We notice that work \cite{xie2025flow} highlights FM's dual potential for likelihood estimation and anomaly detection. The authors proved the ODE is well-posed given velocity field regularity. Our work complements this foundation by investigating ill-posed scenarios with non-regular velocity fields. We compared TCCM~\cite{li2025scalable}, another FM-based UAD method that achieves strong performance on low-dimensional tabular data. TCCM follows a strategy of predicting a contracting vector field, whose probability paths point from data toward the origin. Although this formulation greatly simplifies the geometric objective, it does not fully resolve the initial singularity issue. Table~\ref{tabTCCM} illustrates the significant performance gaps caused by different geometric designs in high-dimensional settings.


\noindent \textbf{High Dimensional Problem in OOD Detection}: Likelihood-based anomaly detection can be viewed as a form of out-of-distribution (OOD) detection, and has its connection to the underlying data manifolds revealed by a classic OOD paradox \cite{kirichenko2020normalizing}. Empirical studies have shown that models trained on relatively complex datasets tend to assign higher likelihood to simpler OOD samples \cite{caterini2022entropic}. Our geometric intuition stems from the following extension of this problem. \cite{bengio2013representation} proposed the manifold hypothesis, which \cite{fefferman2016testing} later formalized by proving that data manifolds can be embedded in high-dimensional spaces. \cite{johnsson2014low} found that for any high-dimensional manifold with a boundary, most of its volume concentrates near the boundary. Recent work \cite{brownverifying} introduced the manifold disjointness hypothesis, positing that data lie on a union of disconnected manifolds with varying intrinsic dimensions. \cite{kamkari2024geometricexplanation} extended this by showing that when two disjoint supports differ in dimension, the lower-dimensional support may exhibit higher density yet lower probability mass—–a critical insight for understanding OOD failures. AN-ODE \cite{dupont2019augmented} enumerates several low-dimensional topological examples where ODE modeling may fail. We posit that such cases are likely to be prevalent in high-dimensional settings.


\noindent \textbf{The Diffusion-based Anomaly Detection}: Diffusion-based methodologies \cite{zhou2024r3d} typically deviate from likelihood-based paradigms, using reverse diffusion for reconstruction and measuring the reconstruction distance to detect anomalies. This paradigm relies on hand-crafted anomaly priors that fall outside our design scope. DTE \cite{livernochediffusion} avoids such predefined assumptions by utilizing the diffusion time posterior as an anomaly proxy and incorporating manifold deviation, forming a loosely related precedent with ours. However, DTE tends to underperform on high-dimensional image data. To our knowledge, these methods are severely bottlenecked by the diffusion steps, leading to prohibitive inference latency.

\section{Limitations and Future Work}
\label{sec:limt}

Training such a high-dimensional input with the push-forward method is computationally and temporally expensive. Despite the model simplicity, accurately predicting the vector field requires significantly increasing the NC to enhance the model capacity, which limits the feasibility of multi-scale architectural scaling. Future work will explore more efficient frameworks. Nevertheless, WT-Flow still achieves unsatisfactory performance on categories with drastic object variations. This is likely because the large distribution shift between unseen normal samples and training data cannot be effectively constrained by our self-limiting mechanism. 

Furthermore, combining an indirect pushforward model with density proxy estimation represents a promising research direction. We observe that the noise paths in diffusion models appear invariant to manifold constraints, which may provide insight into the differences between ODE-driven and SDE-driven approaches. For instance, SDEs permit trajectory crossings through Brownian noise, thereby avoiding the formation of CPF. As a result, samples in energy-based models can diffuse freely into Gaussian annulus instead of being trapped.

\begin{figure*}[!t]
\centering
\includegraphics[width=1.94\columnwidth]{imgs/visual-all.pdf} 
\caption{Visualization of anomaly maps on the MVTec, VisA, MVTec LOCO, MVTec 3D, and MPDD benchmarks.}
\label{fig9}
\end{figure*}

\section{Conclusion}
\label{sec:conc}

This work systematically investigates the limitations of time-reversed flow matching for unsupervised image anomaly detection, with particular focus on inherent irreversibility and dynamic instability in high-dimensional spaces. Building upon these insights, we introduced the local WT mapping, which constructs a self-limiting potential well to selectively constrain normal samples. This mechanism enables a clean geometric separation between normal and anomalous data without explicit density estimation. Empirically, WT-Flow achieved competitive results on five benchmarks, while maintaining remarkable stability and interpretability. Our study not only demonstrates the practical utility of FM but also bridges a theoretical gap between manifold geometry, probability flow, and anomaly detection. We hope that this work inspires future research on density proxies, manifold-constrained dynamics, and their integration with other push-forward models for more robust and explainable anomaly detection.

\bibliographystyle{ieeetr}
\bibliography{main}


\addtocontents{toc}{\protect\setcounter{tocdepth}{-1}}

\begin{IEEEbiography}[{\includegraphics[width=1in,height=1.25in,clip,keepaspectratio]{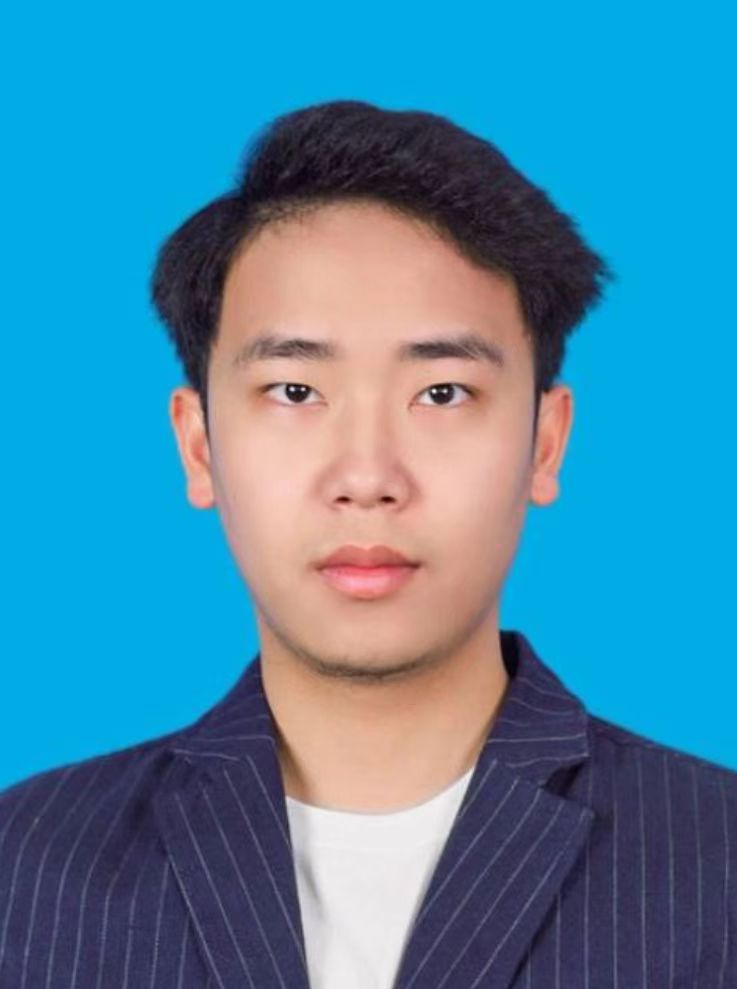}}]{Liangwei Li}
received the M.S. degree from the University of Electronic Science and Technology of China (UESTC), Chengdu, China, where he is currently working toward the Ph.D. degree. His research interests include likelihood estimation and likelihood-based deep generative models.
\end{IEEEbiography}

\begin{IEEEbiography}[{\includegraphics[width=1in,height=1.25in,clip,keepaspectratio]{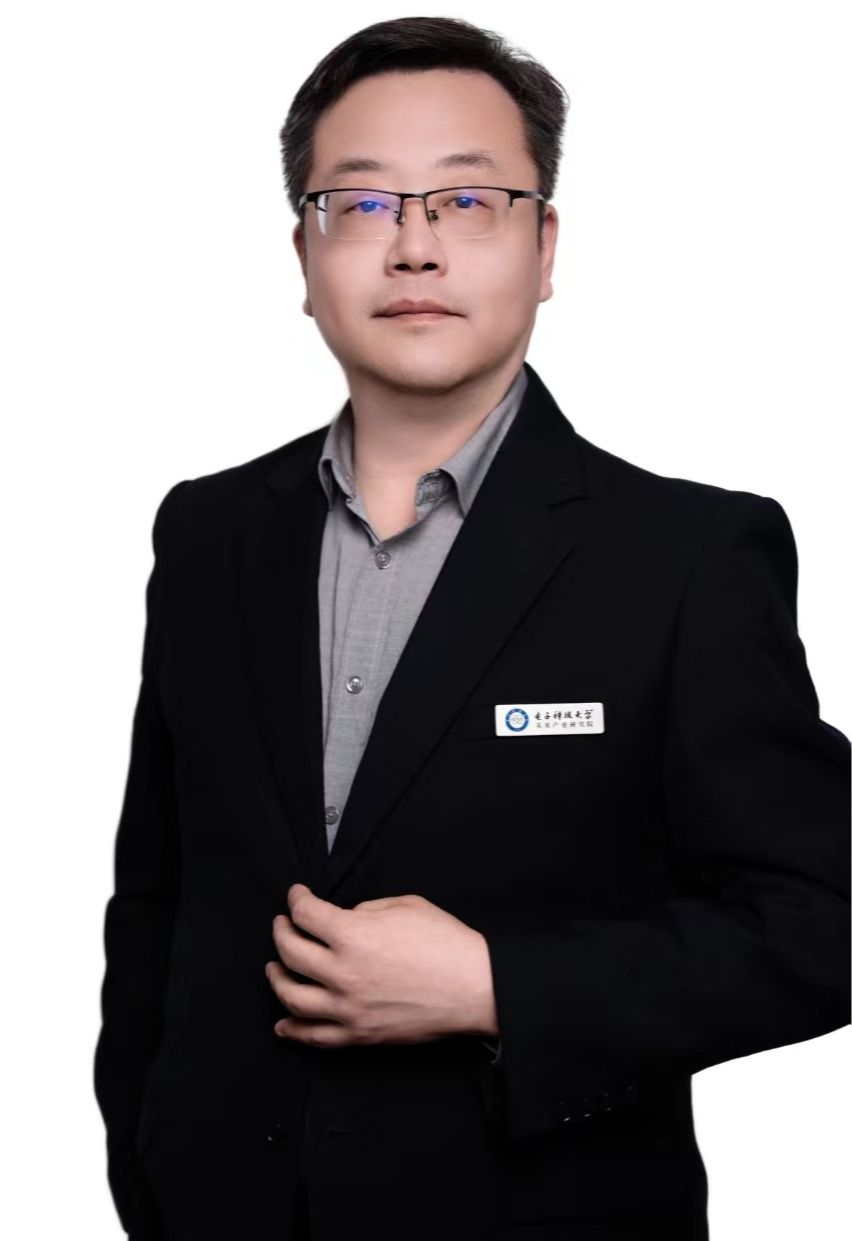}}]{Lin Liu}
received the B.S., M.S., and Ph.D. degrees in optical engineering from the University of Electronic Science and Technology of China (UESTC), Chengdu, China. He is currently a Professor with the School of Optoelectronic Science and Engineering, UESTC, where he serves as the Director of the Modern Opto-Electronic Measurement and Instrument Laboratory (MOEMIL Lab). 
\end{IEEEbiography}

\begin{IEEEbiography}[{\includegraphics[width=1in,height=1.25in,clip,keepaspectratio]{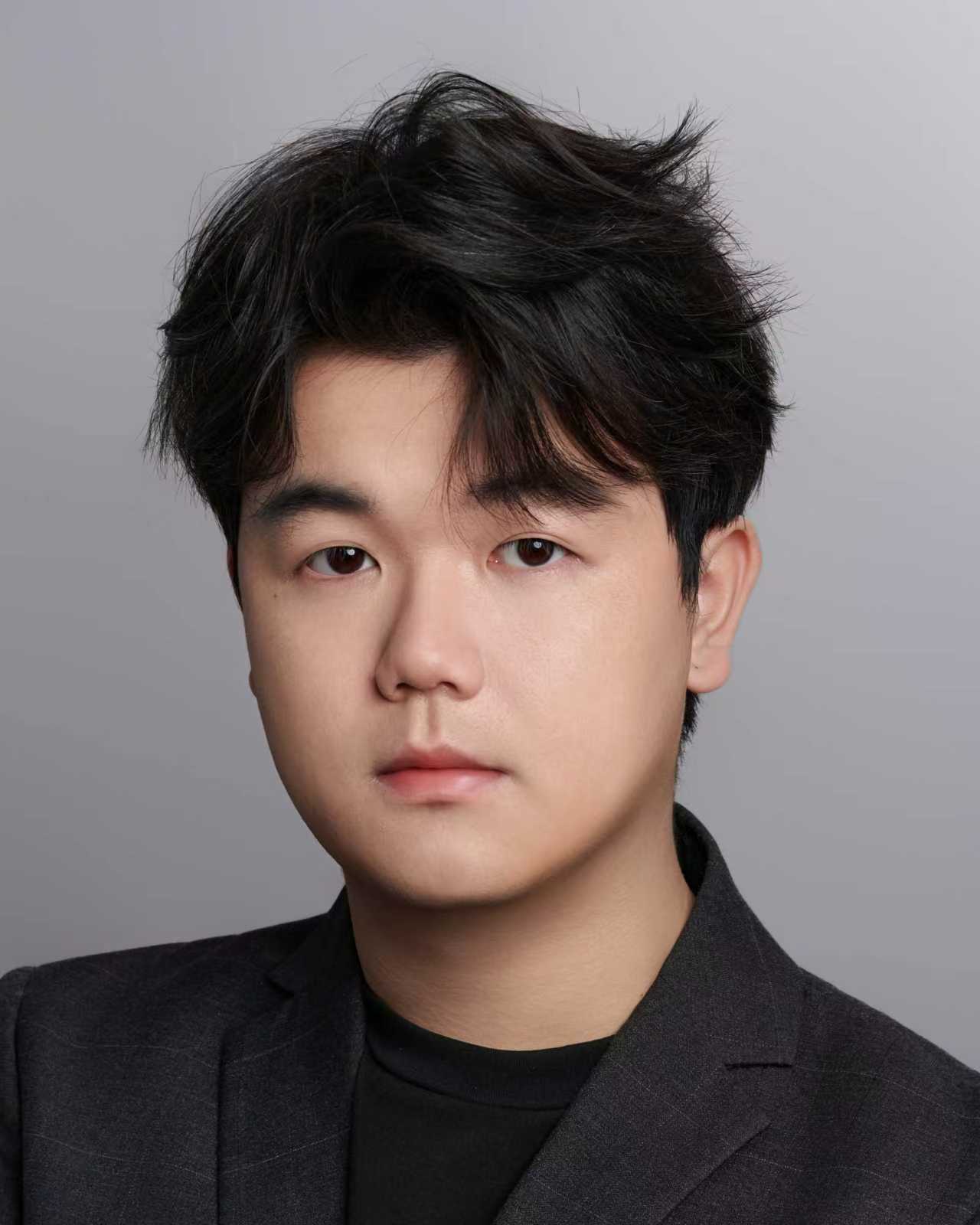}}]{Hanzhe Liang} (Student Member, IEEE) is currently visiting at Mohamed bin Zayed University of Artificial Intelligence~(MBZUAI), UAE. He is pursuing dual degrees, a Bachelor of Management at Shenzhen University, Shenzhen, China, and a Bachelor of Science at Audenica Business School, France. His research interests include computer vision and precision medicine.
\end{IEEEbiography}

\begin{IEEEbiography}[{\includegraphics[width=1in,height=1.25in,clip,keepaspectratio]{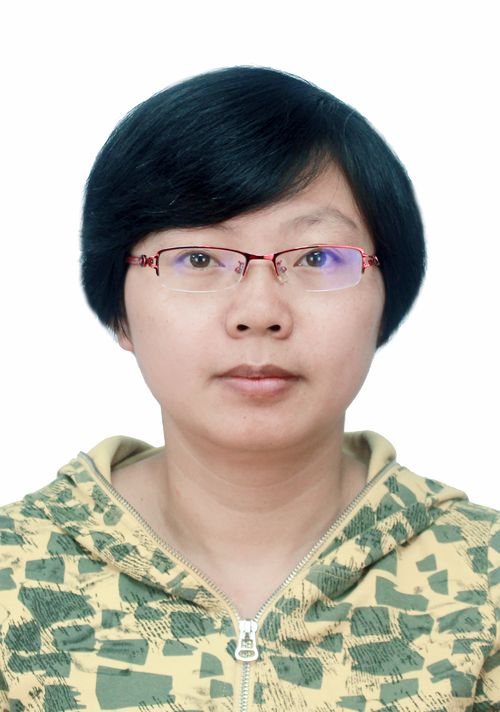}}]{Juanxiu Liu} received the master’s and Ph.D. degrees from the University of Electronic Science and Technology of China, Chengdu, China, in 2004 and 2010, respectively. She is currently an Associate Professor of optical engineering with UESTC. Her research focuses on machine vision and industrial precision automated inspection equipment, attending about the integrated research and application of ``optical, machinery, and software'' integration, and she developed various intelligent detections in the field of biological intelligence and industrial detection.
\end{IEEEbiography}

\begin{IEEEbiography}[{\includegraphics[width=1in,height=1.25in,clip,keepaspectratio]{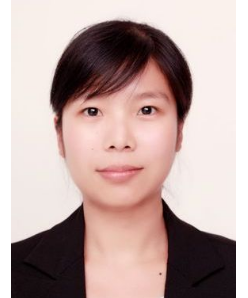}}]
{Liangwei Li} 
(Member, IEEE) received the Ph.D. degrees from the University of Electronic Science and Technology of China, Chengdu, China, in 2013. She is an Associate Professor of optical engineering with the University of Electronic Science and Technology of China. Her research focuses on  image processing and related technologies in the field automatic optical inspection system. The funds presided over by the research include the National Natural Science Foundation and the Central University Research Fund. She has published more than 20 articles in important publications, among which 15 articles have been included in SCI and EI. She has applied for more than 30 patents for inventions and has authorized eight items. She is the coeditor of the book ``The theory and technology of integrated optics''
\end{IEEEbiography}

\begin{IEEEbiography}[{\includegraphics[width=1in,height=1.25in,clip,keepaspectratio]{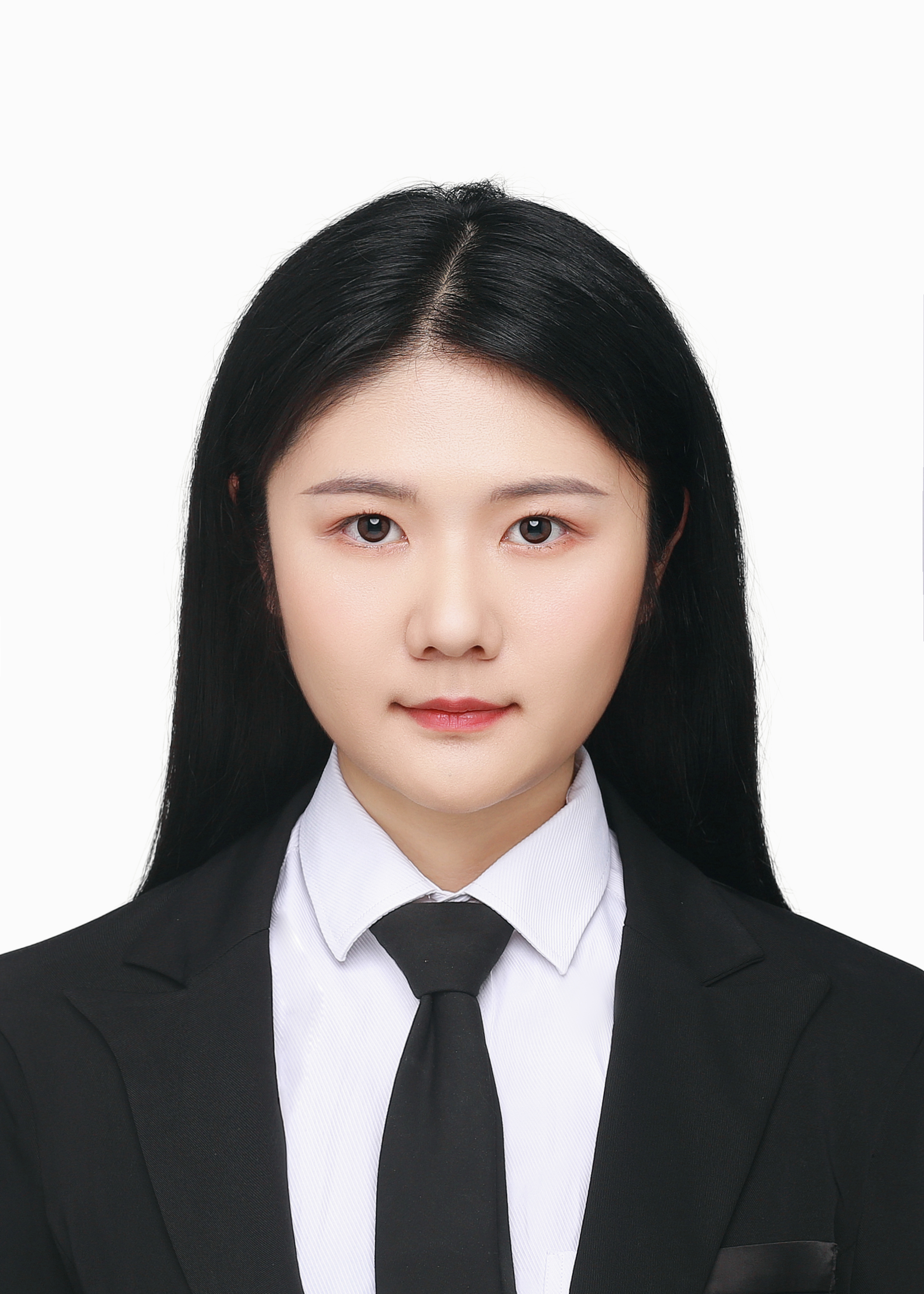}}]{Ruqian Hao} is currently a Postdoctoral Researcher with the School of Optoelectronic Science and Engineering, University of Electronic Science and Technology of China (UESTC), Chengdu, China. Her research interests include medical image processing and non-contact physiological monitoring.

\end{IEEEbiography}

\begin{IEEEbiography}[{\includegraphics[width=1in,height=1.25in,clip,keepaspectratio]{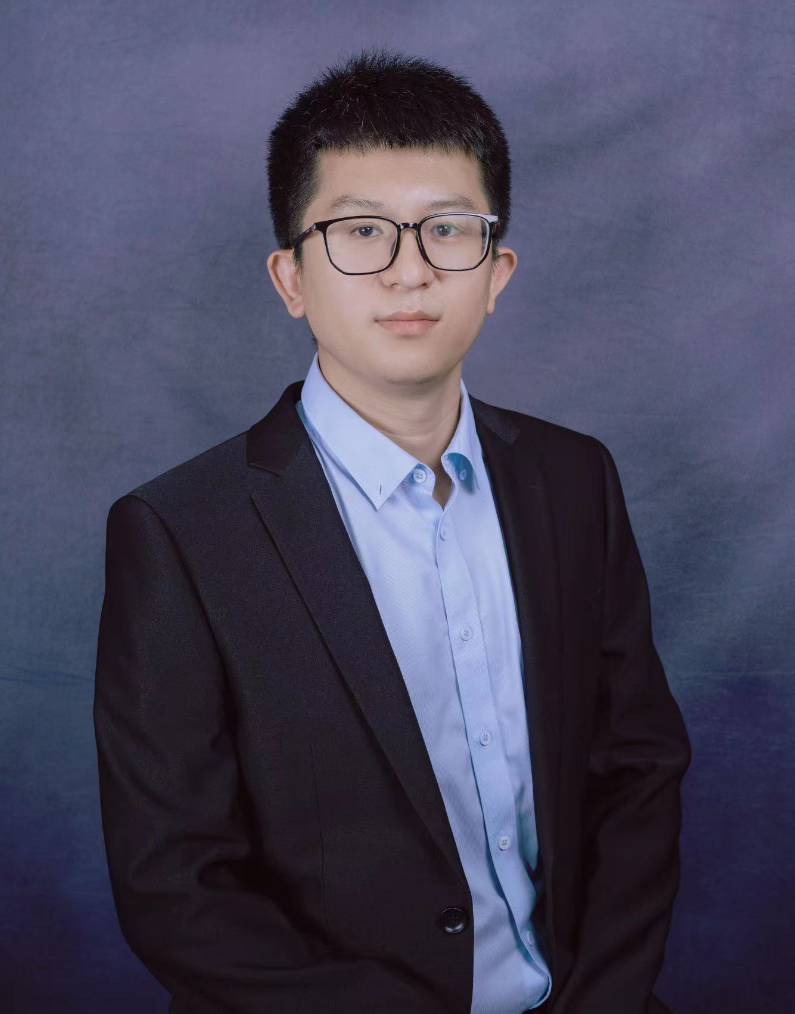}}]
{Xiaohui Du} 
received the D.E. degree from the School of Optoelectronic Science and Engineering, University of Electronic Science and Technology of China (UESTC), Chengdu, China, in 2019. He is currently conduct Post-Doctoral research in UESTC. His research interests include the object detection of microbiological cells, machine learning, and design of structured light optical system.
\end{IEEEbiography}

\begin{IEEEbiography}[{\includegraphics[width=1in,height=1.25in,clip,keepaspectratio]{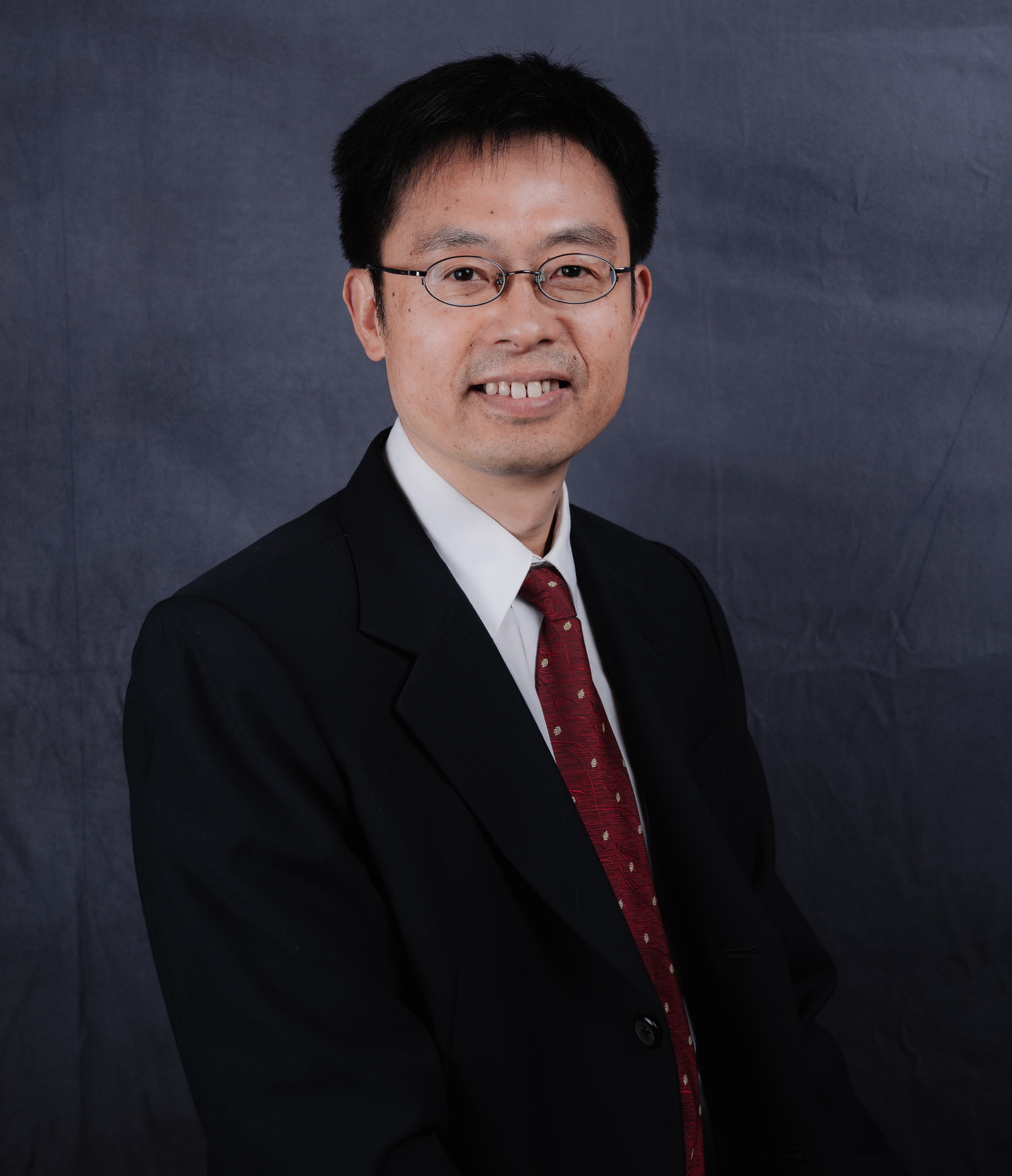}}]{Yong Liu} 
(Senior Member, IEEE) received the M.S. degree from the University of Electronic Science and Technology of China (UESTC), Chengdu, China, in 1994, and the Ph.D. degree from Eindhoven University of Technology, Eindhoven, the Netherlands, in 2004. Since 2007, he has been a Professor with UESTC, where he is currently the Dean of the School of Optoelectronic Science and Engineering. He has authored or coauthored more than 200 journal and conference papers, which have been cited more than 1,200 times (Web of Science). Dr. Liu was a recipient of the IEEE/LEOS (now IEEE Photonics Society) Graduate Student Fellowship in 2003.
\end{IEEEbiography}

\begin{IEEEbiography}[{\includegraphics[width=1in,height=1.25in,clip,keepaspectratio]{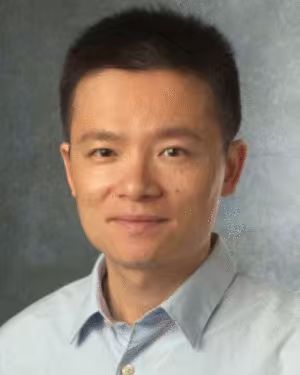}}]{Pan Li} (Fellow, IEEE) received the Ph.D. degree in electrical and computer engineering from the University of Florida, Gainesville, FL, USA, in 2009. He was a Faculty Member with Mississippi State University, Starkville, MS, USA and Case Western Reserve University, Cleveland, OH, USA. He is currently the Academic Vice President and Lixing Chair Professor with Hangzhou Dianzi University, Hangzhou, China. His research interests include artificial intelligence, cybersecurity, and the interplay between them. He was the Editor of nine internationally renowned journals such as IEEE Transactions on Mobile Computing, IEEE Transactions on Big Data, IEEE Transactions on Wireless Communications, IEEE Transactions on Network Science and Engineering, and on the organizing committee and technical program committee for flagship conferences like AAAI, IJCAI, USENIX Security, INFOCOM, and MobiHoc. He was the recipient of NSF CAREER Award and AAIA/AIIA Fellow.\end{IEEEbiography}

\addtocontents{toc}{\protect\setcounter{tocdepth}{2}}

\tableofcontents
\onecolumn

\section{Backgrounds}
\label{sec:bg}

\subsection{Flow-based Anomaly Detection}
\label{sec:bg1}

The primary objective of unsupervised anomaly detection is to identify subtle yet discriminative signals that capture and generalize the intrinsic patterns of normal training data, while distinguishing definable anomalies. Existing methodologies can be broadly categorized into: reconstruction-based \cite{zavrtanik2022dsr, yu2025attention}, knowledge distillation-based \cite{batzner2024efficientad, rudolph2023asymmetric}, pseudo label-Based \cite{zavrtanik2021draem, liu2023simplenet, bae2023pni}, and memory bank-based \cite{roth2022towards, li2024target} approaches. Notably, density-based methods \cite{gudovskiy2022cflow, zhou2024msflow, yu2021fastflow} constitute a particularly compelling theoretically grounded formalism due to the inherent correlation between anomalies and low-frequency observations. This paradigm contributes to understanding the underlying data structure by directly modeling the distribution of normal samples or their features for anomaly identification.

The CFlow \cite{gudovskiy2022cflow} pioneered conditional Normalizing Flows (NF) for unsupervised anomaly detection. The method adaptively models feature distributions at distinct spatial locations, enabling pixel-level anomaly localization through independently trained multi-scale NF-based models. CFlow explicitly computes the Jacobian determinant term and evaluates anomalies based on the likelihood $p_\theta(x)$. While not formally established, subsequent methods forgo direct computation of data's likelihood and instead adopt density proxies using $p_\theta(z)$ in their implementations. FastFlow\cite{yu2021fastflow} significantly improves inference efficiency by replacing fully connected layers with convolutional networks in NFs, allowing parallel processing of high-resolution features. Although CSFlow \cite{rudolph2022fully} improves the consistency of density estimation through cross-scale feature integration with a shared NF module, its application is restricted to image-level anomaly detection. MSFlow \cite{zhou2024msflow} unifies multi-scale features within a single flow model and adopts a hierarchical flow design to replace $1\times1$ convolutions, achieving the best performance in the NFs. 

Recent work \cite{xie2025flow} presents a mathematical framework for FM in generative modeling, exploring their density evolution governed by ODEs, and highlighting their dual potential for likelihood estimation and anomaly detection. The authors proved the ODE is well-posed given velocity field regularity. Our work complements this foundation by investigating ill-posed scenarios with non-regular velocity fields, thereby extending the theoretical understanding of flow-based models in degenerate cases.

\subsection{High-dimensional OOD Problems}
\label{sec:bg2}
Density-based anomaly detection can be viewed as a form of out-of-distribution (OOD) detection \cite{ren2019likelihood}. The connection between density-based models and the underlying data manifolds was first revealed through a classic paradox in OOD detection \cite{nalisnick2018deep, kirichenko2020normalizing}. Empirical studies have shown that models trained on relatively complex datasets tend to assign higher likelihood to simpler OOD samples. This counter-intuitive phenomenon has spurred massive theoretical investigations into the geometric characteristics of probability distributions in high-dimensional manifold spaces. Research on this issue mainly encompasses three aspects: (1) inherent limitations of Gaussian priors in high dimensions \cite{le2021perfect, nalisnick2019detecting}, (2) geometric complexity of data manifolds \cite{cornish2020relaxing, dupont2019augmented, fetayaunderstanding, brownverifying, kamkari2024geometricexplanation}, and (3) architectural constraints in push-forward models \cite{nalisnick2018deep, zhang2021understanding, salmona2022can, ross2024neural}. 

\subsection{Preliminaries of Conditional Flow Matching and Rectified Flow  }
\label{sec:pf2}

Rectified Flow (RF) \cite{liuflow2023} directly establishes a mapping from Gaussian distributions to data samples, operating under the premise that parameterized neural networks can provide smooth interpolation of the data distribution. However, when reversing the probability path, the singularity at the starting point theoretically invalidates this smooth approximation. Conditional Flow Matching (CFM) \cite{lipman2023flow} and OT-Flow \cite{lipman2023flow} circumvent discussion of terminal singularities by adding noise to the data, while diffusion models similarly direct predict noise-perturbed quantities.

\noindent \textbf{Preliminaries of Conditional Flow Matching}: Consider a continues time-dependent probability density function (PDF) $p_t(x)$, where $t \sim \mathcal{U}(0,1)$, and data $x=(x^1,x^2,\ldots,x^d) \in \mathbb{R^d}$. The PDF is driven by a ordinary differential equation (ODE) through a time-dependent vector field $u_t$:

\begin{equation}
    u_t(x_t)=\frac{d}{dt} x_t.
\end{equation}

In work \cite{lipman2023flow}, the $\phi_t(x)=x_t$ denotes a \textit{flow}, with $\phi_0(x)=x_0$. Therefore, the $p_t$ over time can be expressed by the push-forward equation:

\begin{equation}
    p_t=[\phi_t]_*p_0(x)=p_0(\phi_t^{-1}(x))\text{det}[\frac{\partial\phi_t^{-1}}{\partial x}(x)].
    \label{eq:21}
\end{equation}

Once a $v_{\theta,t}$ parameterized by $\theta$, generates $p_t$ with its $\phi_t$ satisfies the Eq. \eqref{eq:21}. We say the $v_{\theta,t}$ defines a probability density path $p_t$, which is a time-dependent diffeomorphic map. The Flow Matching (FM) objective is defined as

\begin{equation}
    \mathcal{L}_\text{FM}(\theta)=\mathbb{E}_{t,p_t(x)}\lVert v_{\theta}(x,t)-u_t(x) \rVert.
\end{equation}

The objective of FM is conceptually straightforward. However, the absence of closed-form solutions for $u_t$ and prior knowledge regarding the construction of $p_t$ renders it impractical for direct application. CFM constructs conditional probability path $p_t(x|x_1)$ by conditioning on specific samples $x_1\sim p^*(x_1)$, then obtains the marginal probability path $p_t(x)$ through marginalizing it by

\begin{equation}
    p_t(x)=\int p_t(x|x_1)p^*(x_1)dx_1,
\end{equation}

\noindent where $p^*(x_1)$ is the unknown data distribution. 

Specifically, when the conditional probability path is defined as a Gaussian probability path,
\begin{equation}
    p_t(x|x_1)=\mathcal{N}(x|\mu_t(x_1),\sigma_t(x_1)^2 I),
\end{equation}

\noindent we obtain the Optimal Transport (OT) flow through linear interpolation, where $\mu_t(x_1)=tx_1$, and $\sigma_t(x_1)=1-(1-\epsilon)t$. The $\epsilon$ is a sufficient small noise.

Thus, the push-forward equation is:

\begin{equation}
    \begin{aligned}
        \phi_t(x) = \mu_t(x_1) + \sigma_t(x_1)x_0 =tx_1+(1-(1-\epsilon)t)x_0.
    \end{aligned}
\end{equation}

\noindent where $x_0\sim \mathcal{N}(\mathbf{0},\mathbf{I})$. By Theorem 2 in \cite{lipman2023flow}, the marginal vector field generates the marginal probability path, due to

\begin{equation}
    \nabla_\theta \mathcal{L}_\text{FM}(\theta) = \nabla_\theta \mathcal{L}_\text{CFM}(\theta)
\end{equation}

Consequently, we can directly obtain the marginal vector field $p_t(x)$ through the estimation of the conditional vector field $p_t(x|x_1)$.

According to Continuity Equation \cite{lipman2023flow}, the conditional vector field $u_t(x|x_1)$ can be defined as

\begin{equation}
    \begin{aligned}
        u_t(x_t|x_1) = \frac{d}{dt} \phi_t(x) =\mu ^\prime _t(x_1) + \sigma^\prime _t(x_1)x,
        \label{eq:27}
    \end{aligned}
\end{equation}

We perform reparameterization of $x_0$ with $x_t$ by

\begin{equation}
    x_0=\frac{x_t-\mu_t(x_1)}{\sigma_t(x_1)}.
    \label{eq:28}
\end{equation}

Then, substituting Eq. \eqref{eq:28} in Eq. \eqref{eq:27}, we obtain

\begin{equation}
    \begin{aligned}
        u_t(x_t|x_1)&=\mu ^\prime _t(x_1) + \frac{\sigma_t^\prime(x_1)}{\sigma_t(x_1)}(x_t-\mu_t(x_1)), \\
        &=x_1+\frac{-(1-\epsilon)}{1-(1-\epsilon)t}(tx_1+(1-(1-\epsilon)t)x_0), \\
        &=x_1 - (1-\epsilon)x_0.
    \end{aligned}
\end{equation}

We formulate the OT-flow objective by first sampling $x_0$ from $\mathcal{N}(\mathbf{0},\mathbf{I})$ and pairing it with $x_1$, then employing a parameterized neural network $v_\theta$ to predict the conditional vector field $u_t(x|x_1)$. This yields the CFM loss function in the following form

\begin{equation}
    \mathcal{L}_\text{CFM}(\theta)=\mathbb{E}_{t,p(x_1),p(x_0)} \lVert x_1-(1-\epsilon)x_0-v_\theta(t,x_t) \rVert.
    \label{eq:30}
\end{equation}

\noindent \textbf{Preliminaries of Rectified Flow}: From Eq. \eqref{eq:30}, we observe that OT-Flow aims to model the trajectory from a noisy Gaussian $(1-\epsilon)x_0$ to the data $x_1$, whereas RF adopts a more direct approach. Specifically, RF eliminates the noise term, resulting in a straight flow path connecting $x_0$ directly to $x_1$, where the objective is

\begin{equation}
    \mathcal{L}_\text{RF}(\theta)=\mathbb{E}_{t,p(x_1),p(x_0)}\lVert x_1-x_0-v_\theta(t,x_t)\rVert.
\end{equation}

In their work, the temporal domain for RF is defined over the interval $t \in [0,a)$ where $a < 1$, with a well-regularized neural network $v_\theta$ serving as a smooth functional approximation to the flow map $\phi_t$.

\section{The Proofs}
\label{sec:pf}
\setcounter{equation}{10} 
\setcounter{figure}{7} 
\setcounter{table}{6} 

\subsection{Proofs for Non-Invertibility in Flow Matching}
\label{sec:pf1}


Given a data sample $x_0$ drawn from an unknown distribution $p^*(x)$, and a standard Gaussian variable $x_1 \sim \mathcal{N}(\mathbf{0}, \mathbf{I})$, we define a Gaussian conditional probability path $p_t(x|x_0)$ via linear interpolation, that

\begin{equation}
    p_t(x|x_0)=\mathcal{N}(x|\mu_t(x_0),\sigma_t(x_0)^2I),
\end{equation}

\noindent where $\mu_t(x_0)=(1-t)x_0$, and $\sigma_t=t$, with $t\sim \mathcal{U}(0,1)$. This construction can be formally interpreted as a linear time-continuous noising process, with $x_{t=0}=x_0$, $x_{t=1}=x_1$, and 

\begin{equation}
    \begin{aligned}
        x_t &= \mu_t(x_0) + \sigma_t(x_0)x_1, \\
            &=(1-t)x_0+tx_1.
    \end{aligned}
\end{equation}

The push-forward equation \cite{lipman2023flow} is defined as 

\begin{equation}
    [\phi_t(x)]_*p(x_0)=p_t(x|x_0)+C,
\end{equation}

\noindent with $\phi_t(x)=x_t$ and $C$ is a constant. Based on the Continuity Equation \cite{ben2022matching, chen2018neural}, we derive the instantaneous rate of change in continuous time, where the conditional vector field $u_t(x|x_0)$ is given by

\begin{equation}
    \begin{aligned}
        u_t(x_t|x_0) &= \frac{\mathrm{d}}{\mathrm{d}t}\phi_t(x), \\
                     &=\mu ^\prime _t(x_0) + \sigma^\prime _t(x_0)x_1.
                     \label{eq:13}
    \end{aligned}
\end{equation}

We perform reparameterization of $x_1$ with $x_t$ by

\begin{equation}
    x_1=\frac{x_t-\mu_t(x_0)}{\sigma_t(x_0)}.
    \label{eq:14}
\end{equation}

Then, substituting Eq.\eqref{eq:14}  into Eq. \eqref{eq:13}, we obtain

\begin{equation}
    \begin{aligned}
        u_t(x_t|x_0)&=\mu ^\prime _t(x_0) + \frac{\sigma_t^\prime(x_0)}{\sigma_t(x_0)}(x_t-\mu_t(x_0)), \\
        &=-x_0+\frac{1}{t}((1-t)x_0+tx_1-(1-t)x_0), \\
        &=\frac{tx_1}{t}-x_0.
    \end{aligned}
\end{equation}

The division-by-zero error at $t=0$ indicates that this path is undefined at this point. Thus, we have Theorem~\ref{th:singularity}

\begin{theorem}[Initial Singularity]
    The conditional vector field ${u}(x,t|x_0)$ of rFM possesses an initial singularity w.r.t. time variable ${t}$ at $t=0$, which leaves ${u}(x,0|x_0)$ undefined.
    \label{th:singularity}
\end{theorem}

\subsection{Proofs for the Lipschitzness of the Reversed Vector Field}
\label{sec:pflps}

\begin{lemmanum}{2}
The marginal vector field $u$ can be written as:

\begin{equation}
    u(x,t)=-\frac{1}{1-t} x -\frac{t}{1-t} \nabla_x \log p_t(x)
\end{equation}

\noindent where $p_t$ is the density of $X_t$, and $X_t = (1-t)X_0 + tX_1$.

\label{lm2}
\end{lemmanum}

\textit{Proof.} 

\begin{align}
        u(x,t) &= \mathbb{E}[X_1-X_0 \mid X_t=x]  \notag \\
        &= \mathbb{E}\left[\frac{tX_1+(1-t)X_0-(1-t)X_0}{t} -X_0 | X_t=x \right]  \notag \\
        &= \frac{1}{t}x- \frac{1}{t}\mathbb{E}[X_0|X_t=x]    \notag \\
        &= \frac{1}{t}x- \frac{1}{t} \int x_0 p_{0|t}(x_0|x) \text{d}x_0  \notag \\
        &= \frac{1}{t}x- \frac{1}{t} \int \frac{x_0 p_{t|0}(x|x_0)p_0(x_0)}{p_t(x)} \text{d}x_0  \notag \\
        &= \frac{1}{t}x- \frac{1}{t} \int \frac{1}{\sqrt{(2 \pi)^dt^{2d}}} \frac{x_0 \exp (-\frac{\lVert x-(1-t)x_0 \rVert^2}{2t^2})p_0(x_0)}{p_t(x)} \text{d}x_0 \notag \\
        &= \frac{1}{t}x- \frac{t}{1-t} \int \frac{1}{\sqrt{(2 \pi)^dt^{2d}}} \frac{\left(\frac{x}{t^2} -\frac{x-(1-t)x_0}{t^2}\right) \exp (-\frac{\lVert x-(1-t)x_0 \rVert^2}{2t^2})p_0(x_0)}{p_t(x)} \text{d}x_0 \notag \\
        &= (\frac{1}{t} - \frac{1}{t(1-t)})x- \frac{t}{1-t} \int \frac{1}{\sqrt{(2 \pi)^dt^{2d}}} \frac{\nabla_x \exp (-\frac{\lVert x-(1-t)x_0 \rVert^2}{2t^2})p_0(x_0)}{p_t(x)} \text{d}x_0 \notag \notag \\ 
        &= -\frac{1}{1-t} x -\frac{t}{1-t} \nabla_x \log p_t(x) \label{uxt18} \\ \notag 
\end{align}

We also have $\nabla_x \log p_t(x) =\frac{1-t}{t^2} \mathbb{E}[X_0|X_t=x] -\frac{1}{t^2}x$ from \eqref{uxt18}.

\begin{lemmanum}{3}
    The upper bound of the partial derivative of the marginal vector field $u$ w.r.t $t \in [T,1]$ is:

    \begin{equation}
        \sup \limits_{t \in [T,1]}  \sup \limits_{x \in [-R,R]^D} \lVert \partial_t u(x,t) \rVert=\mathcal{O}(\frac{D^2}{T^4}),
    \end{equation}
    \noindent where $t_{min}=T$ is the truncation time.
    \label{lm3}
\end{lemmanum}

\textit{Proof.} To ease notation, we define $\phi_t(x) := \int  A_t p_0(x_0)\dd x_0 :=\int \exp \left( -\frac{\Vert x-(1-t)x_0 \Vert^2}{2t^2} \right) p_0(x_0)\dd x_0$, which is the unnormalized version of $p_t(x)$. Note that $\nabla_x \log \phi_t(x) = \nabla_x \log p_t(x)$. By Lemma \ref{lm2}, we have

\begin{align}
    \partial_t u(x, t) &= -\frac{1}{(1-t)^2}x  - \frac{1}{(1-t)^2}\nabla_x \log p_t(x) - \frac{t}{1-t} \partial_t \nabla_x \log p_t(x) \notag \\
    &= -\frac{1}{(1-t)^2}x -\frac{1}{(1-t)^2} \left( \frac{1-t}{t^2} \mathbb{E}[X_0|X_t=x] -\frac{1}{t^2}x \right) - \frac{t}{1-t} \partial_t \frac{\nabla_x \phi_t(x)}{\phi_t(x)} \notag \\
    &= \frac{1+t}{t^2(1-t)}x -\frac{1}{t^2(1-t)}\mathbb{E}[X_0|X_t=x] - \frac{t}{1-t} \left( \frac{\partial_t \nabla_x \phi_t(x)}{\phi_t(x)} - \frac{\partial_t \phi_t(x)}{\phi_t(x)} \frac{\nabla_x \phi_t(x)}{\phi_t(x)}\right). \label{eq:putl2} \\ \notag
\end{align}

Then we focus on the last term in \eqref{eq:putl2}, where

\begin{align}
   \frac{\partial_t \nabla_x \phi_t(x)}{\phi_t(x)} &= \frac{\partial_t}{\phi_t(x) } \nabla_x \int \exp \left( -\frac{\Vert x-(1-t)x_0 \Vert^2}{2t^2} \right) p_0(x_0)\dd x_0 \notag \\
   &= \frac{\partial_t}{\phi_t(x) } \int \frac{(1-t)x_0-x}{t^2}\cdot A_t \cdot p_0(x_0)\dd x_0 \notag \\
   &= \frac{1}{\phi_t(x)} \int \left[ \frac{(t-2)x_0}{t^3} +\frac{2}{t^3}x +\frac{[(1-t)x_0-x] \cdot [(t-2)x_0^{\top}x+(1-t)\Vert x_0 \Vert^2+ \Vert x \Vert^2]}{t^5} \right]\cdot A_t \cdot  p_0(x_0)\dd x_0 \notag \\
   &= \frac{t-2}{t^3} \mathbb{E}[X_0|X_t=x] + \frac{2}{t^3}x  + \frac{(1-t)(t-2)}{t^5}\mathbb{E}[X_0X_0^{\top}|X_t=x]  \notag \\
   &+ \frac{(1-t)^2}{t^5} \mathbb{E}[\Vert X_0\Vert^2X_0|X_t=x] + \frac{(1-t)}{t^5} \mathbb{E}[X_0|X_t=x] \Vert X \Vert^2 \notag \\
   &- \frac{(t-2)}{t^5} \mathbb{E}[X_0^{\top}x|X_t=x]x - \frac{(1-t)}{t^5} \mathbb{E}[\Vert X_0 \Vert^2|X_t=x]x - \frac{1}{t^5} \Vert x\Vert^2x ,\label{eq:putl3} \\ \notag
\end{align}

and by some calculus, we have

\begin{align}
   \frac{\partial_t \phi_t(x)}{\phi_t(x)} &= \frac{\partial_t}{\phi_t(x) }  \int \exp \left( -\frac{\Vert x-(1-t)x_0\Vert ^2}{2t^2} \right) p_0(x_0)\dd x_0 \notag \\
   &= \frac{1}{\phi_t(x) }  \int A_t \cdot \frac{(t-2)x_0x+(1-t) \Vert x_0 \Vert^2 +  \Vert x \Vert^2}{t^3} p_0(x_0)\dd x_0 \notag \\
   &= \frac{t-2}{t^3} \mathbb{E}[X_0^{\top}x|X_t=x] + \frac{1-t}{t^3} \mathbb{E}[\Vert X_0 \Vert^2|X_t=x] +\frac{1}{t^3} \Vert x\Vert ^2 .\label{eq:putl4} \\ \notag
\end{align}

and

\begin{align}
   \frac{\nabla_x \phi_t(x)}{\phi_t(x)} &= \frac{\nabla_x}{\phi_t(x) } \int \exp \left( -\frac{\Vert x-(1-t)x_0 \Vert^2}{2t^2} \right) p_0(x_0)\dd x_0 \notag \\
   &= \frac{1}{\phi_t(x) }  \int A_t \cdot \frac{(1-t)x_0 - x}{t^2} p_0(x_0)\dd x_0 \notag \\
   &= -\frac{1}{t^2}x + \frac{1-t}{t^2}\mathbb{E}[X_0|X_t=x]  \label{eq:putl5} \\ \notag
\end{align}

Combining \eqref{eq:putl2}, \eqref{eq:putl3}, \eqref{eq:putl4} and \eqref{eq:putl5}, we obtain

\begin{align}
    \partial_t u(x, t) &= -\frac{1}{t^2}x + \frac{1}{t^2} \mathbb{E}[X_0|X_t=x] - \frac{t-2}{t^4} \text{Cov} [X_0|X_t=x]x  \notag \\ 
    &-\frac{1-t}{t^4} (\mathbb{E}[X_0\Vert X_0  \Vert^2|X_t=x] - \mathbb{E}[X_0|X_t=x]\mathbb{E}[ \Vert X_0  \Vert^2|X_t=x])   \label{eq:putl6} \\ \notag
\end{align}

Then we have 

\begin{align}
    \Vert \partial_t u(x, t) \Vert &\leq \frac{1}{t^2} \Vert x\Vert  + \frac{1}{t^2} \Vert \mathbb{E}[X_0|X_t=x]\Vert  + \frac{2-t}{t^4} \|\operatorname{Cov} [X_0|X_t=x] \|_{\operatorname{op}} \|x\| \notag \\ 
    &+\frac{1-t}{t^4} (\mathbb{E}[X_0\Vert X_0  \Vert^2|X_t=x] + \mathbb{E}[X_0|X_t=x]\mathbb{E}[ \Vert X_0  \Vert^2|X_t=x])  \label{eq:putl7} \\ \notag
\end{align}

\noindent where the operator norm $\|\cdot\|$ is defined of a matrix $A$ as $\|A\|_{\operatorname{op} } := \sup_{\|x\|\leq 1}  \|Ax\|$.

Note that $p_0$ is assumed to be supported on $[0, 1]^D$, we have $\| \mathbb{E} [X_0|X_t=x] \leq  \mathbb{E} [\|X_0\|^2 |X_t=x]^{1/2} \| \leq D^{1/2}  $  and $\| \mathbb{E} [X_0\Vert X_0  \Vert^2|X_t=x] \leq  \mathbb{E} [\|X_0\|^6 |X_t=x]^{1/2} \| \leq D^{3/2} $. And we have the following inequality for any $u \in \mathbb{R}^D$,

\begin{align}
    u^\top \operatorname{Cov}[X_0|X_t=x]u & = \mathbb{E}[ u^\top X_0 X_0^\top u | X_t=x] - \mathbb{E}[ u^\top X_0| X_t=x] \mathbb{E}[ u X_0^\top| X_t=x]\notag \\ 
    &=\mathbb{E}[(u^\top X_0)^2|X_t=x] - \mathbb{E}[ u^\top X_0| X_t=x]^2 \leq 2D \|u \|^2 \\ \notag
\end{align}

Hence we have $\|\operatorname{Cov} [X_0|X_t=x] \|_{\operatorname{op}} \leq 2D$. Using these inequalities above, we have

\begin{equation}
        \sup \limits_{t \in [T,1]}  \sup \limits_{x \in [-R,R]^D}  \lVert \partial_t u(x,t) \rVert \leq \frac{R\sqrt{D}}{T^2} + \frac{\sqrt{D}}{T^2} + \frac{2(2-T)}{T^4}D^{3/2}R+ \frac{2(1-T)}{T^4}D^{3/2},
\end{equation}

Considering Lemma \ref{lm:ln}, for sufficiently high dimensions $D$, we introduce the simplification below.

\begin{equation}
\sup \limits_{t \in [T,1]}  \sup \limits_{x \in [-R,R]^D}  \lVert \partial_t u(x,t) \rVert = \frac{D}{T^2} + \frac{\sqrt{D}}{T^2} + \frac{2(2-T)}{T^4}D^{2}+ \frac{2(1-T)}{T^4}D^{3/2}.
\end{equation}

Note that $T > 0$, the above inequality implies $\sup \limits_{t \in [T,1]}  \sup \limits_{x \in [-R,R]^D}  \lVert \partial_t u(x,t) \rVert = \mathcal{O}(\frac{D^{2}}{T^4})$.

\begin{lemmanum}{4}
    The upper bound of the gradient of the spatial variable of the marginal vector field $u$ w.r.t $x \in [-R,R]^D$ is:

    \begin{equation}
        \sup \limits_{t \in [T,1]}  \sup \limits_{x \in [-R,R]^D} \lVert  \nabla_x u(x,t) \rVert \leq \max \left\{ \frac{1}{T}, \frac{(1-T)D}{T^3}\right\}
    \end{equation}
    \label{lm:ubxvf}
\end{lemmanum}

\textit{Proof.} By Lemma \ref{lm2}, we have

\begin{equation}
       \nabla_x {u}(x, t) = -\frac{1}{1-t} I_D -\frac{t}{1-t} \nabla^2_x \log p_t(x),
       \label{eq:nx59}
\end{equation}

Then, we define $x^{\otimes 2} := xx^\top$, and the Hessian $\nabla^2_x \log p_t(x)$ can be computed as

\begin{align}
        \nabla^2_x \log p_t(x) &= \nabla_x (\nabla_x\log \phi_t(x))  \notag  \\ 
        &= \nabla_x \left(\frac{\nabla_x \phi_t(x) }  {\phi_t(x)} \right) \notag  \\ 
        &= \nabla_x \left( \frac{ \int \frac{(1-t)x_0-x}{t^2} A_t p_0(x_0) \dd x_0}{\int A_t p_0(x_0) \dd x_0}\right)  \notag \\
        &= -\frac{\int A_t p_0(x_0) \dd x_0)^{\otimes2} }{t^2(\int A_t p_0(x_0) \dd x_0)^{\otimes2}}  \notag \\
        &+ \frac{[\int (1-t)x_0-x)\nabla_x A_t p_0(x_0) \dd x_0] \cdot[ \int A_t p_0(x_0) \dd x_0 ]^\top}{t^2(\int A_t p_0(x_0) \dd x_0)^{\otimes2}} \notag \\
        &- \frac{[\int (1-t)x_0-x) A_t p_0(x_0) \dd x_0 ] \cdot[ \int \nabla_x A_t p_0(x_0) \dd x_0]^\top}{t^2(\int A_t p_0(x_0) \dd x_0)^{\otimes2}} \notag \\
        &= -\frac{1}{t^2} I_D + \frac{\int \left (\frac{(1-t)x_0-x}{t^2} \right)^{\otimes 2} A_t p_0(x_0) \dd x_0}{\int A_t p_0(x_0) \dd x_0} -\left (\frac{\int \frac{(1-t)x_0-x}{t^2}  A_t p_0(x_0) \dd x_0}{\int A_t p_0(x_0) \dd x_0}\right)^{\otimes 2} \notag \\
        &= -\frac{1}{t^2} I_D + \mathbb{E}\left[\left(\frac{(1-t)X_0-x}{t^2}\right)^{\otimes 2}|X_t=x \right] - \mathbb{E}\left[\frac{(1-t)X_0-x}{t^2}|X_t=x \right] ^{\otimes 2} \notag \\
        &= -\frac{1}{t^2} I_D + \text{Cov}\left[\frac{(1-t)X_0-x}{t^2}|X_t=x \right] \notag \\
        &= -\frac{1}{t^2} I_D + \frac{(1-t)^2}{t^4}\text{Cov}[X_0|X_t=x]. \label{eq:nx60}
\end{align}

Combine \eqref{eq:nx59} and \eqref{eq:nx60}, we have $\nabla_x {u}(x, t) = \frac{1}{t} I_D - \frac{1-t}{t^3}\text{Cov}[X_0|X_t=x]$. Since we assume the target distribution $p_0$ is supported on $[0, 1]^D$ , we have the following evaluation of the covariance matrix $0  \preceq  \operatorname{Cov}[X_1|X_t = x]  \preceq  DI_D$, and

\begin{equation}
       \left( \frac{1}{T} - \frac{(1-T)^D}{(T)^3} \right) I_D \preceq \nabla_x {u}(x, t) \preceq \frac{1}{T} I_D.
\end{equation}

If Assumption 1 holds, $u(x,t)$ is $\xi$-Lipschitz continuous w.r.t. $x \in \mathbb{R}^D \times [T,1]$, where $\xi \leq  \operatorname{max}\{ \frac{1}{T}, \frac{(1-T)D}{T^3} \}$. Then, Theorem \eqref{th:lps-rfm} can be directly derived from Lemmas \eqref{lm3} and \eqref{lm:ubxvf}.

\begin{theorem}[Lipschitzness of rFM]
    If the Assumption 1 hold, then with truncation $t \in [T,1]$, the marginal vector field ${u}(x,t)$ of rFM is jointly $\xi$-Lipschitz continuous w.r.t. $(x,t)$ on $\mathbb{R}^D \times [T,1]$, where $\xi \leq \operatorname{max}\{ \frac{1}{T},  \frac{(1-T)D}{T^3}, \mathcal{O} (\frac{D^2}{T^4})  \} $.
    \label{th:lps-rfm}
\end{theorem}

\subsection{Proofs for Transport Degeneration under Manifold Coincidence}
\label{sec:pf3}

We first prove that the LayerNorm (LN) projects D-dimen-sional inputs onto the Gaussian spherical surface corresponding to their dimensionality. We then demonstrate that while BatchNorm (BN) outputs match the Gaussian radius in expectation, their instantiated positions depend on inter-sample variance and generally fall below the theoretical radius due to feature sparsity. Finally, we establish that LN-normalized samples exhibit normal transport degeneracy.

\begin{lemmanum}{5}
Assume the Assumptions 1 and 2 hold and the data are non-trivial. Then LayerNorm \cite{ba2016layer} maps the input data $x \in \mathbb{R}^{C \times H \times W} $ onto a Gaussian annulus with same dimension $D$.

    \begin{equation}
        \lVert \text{LN}(x) \rVert _2=R=\sqrt{D}.
    \end{equation}
    \label{lm:ln}
where $D = C \times H \times W $ represents the total dimension.
\label{lm5}
\end{lemmanum}

\textit{Proof.} Consider a single feature sample represented as a tensor $x \in \mathbb{R}^{C \times H \times W}$. Let $D := CHW$, and denote the vectorized entries as $x = (x_1, \dots, x_D)^\top \in \mathbb{R}^D$. The sample mean $\mu$ and the variance $\sigma^2$ are defined as:
\begin{equation}
\mu = \frac{1}{D} \sum_{i=1}^D x_i, \qquad \sigma^2 = \frac{1}{D} \sum_{i=1}^D (x_i - \mu)^2.
\end{equation}
The Layer Normalization operation transforms each entry $x_i$ into:
\begin{equation}
\text{LN}(x)_i = \tilde{x}_i = \frac{x_i - \mu}{\sqrt{\sigma^2}}, \qquad i = 1, \dots, D.
\end{equation}
Consequently, the squared Frobenius norm of the normalized output $\text{LN}(x)$ yields:
\begin{equation}
\lVert \text{LN}(x) \rVert^2 = \sum_{i=1}^D \tilde{x}_i^2 = \sum_{i=1}^D \frac{(x_i - \mu)^2}{\sigma^2} = \frac{1}{\sigma^2} \sum_{i=1}^D (x_i - \mu)^2 = D.
\end{equation}
This derivation establishes the algebraic identity. In practical implementations, a small constant $\epsilon > 0$ is introduced for numerical stability:
\begin{equation}
\tilde{x}_i = \frac{x_i - \mu}{\sqrt{\sigma^2 + \epsilon}}, \qquad i = 1, \dots, D,
\end{equation}
which theoretically causes the normalized norm to be slightly less than the Gaussian radius $R = \sqrt{D}$. Specifically, the squared norm is scaled by the factor:
\begin{equation}
\sum_{i=1}^D \tilde{x}_i^2 = D \cdot \frac{\sigma^2}{\sigma^2 + \epsilon}.
\end{equation}
Furthermore, the choice of variance estimator (e.g., using $D-1$ as the denominator for unbiased estimation) may shift the identity to $\sqrt{D-1}$. Regardless of such implementation nuances and internal channel correlations, Layer Normalization inherently ensures that each sample is constrained to a fixed hyperspherical manifold defined by the dimensionality of the feature space.

\begin{lemmanum}{6}
Assume Assumptions 1 and 2 hold. For a batch of samples $\mathcal{B} = \{x^{(1)}, \dots, x^{(N)}\}$, BatchNorm \cite{ioffe2015batch} maps the data distribution such that the expectation of the squared Frobenius norm of any sample $x \in \mathcal{B}$ is asymptotically governed by the Gaussian radius $R=\sqrt{D}$, where $D=CHW$.
\begin{equation}
    \mathbb{E} [ \lVert \operatorname{BN}(x) \rVert^2_F ] \approx D.
\end{equation}
\label{lm:bn}
\end{lemmanum}

\textit{Proof.} Let $x_{n,c,h,w}$ denote the feature value at batch index $n$, channel $c$, and spatial location $(h,w)$. Unlike Layer Normalization, BatchNorm computes the mean $\mu_c$ and variance $\sigma_c^2$ across the batch $N$ and spatial dimensions $H, W$ for each channel $c$ independently:

\begin{equation}
\mu_c = \frac{1}{NHW} \sum_{n=1}^N \sum_{h=1}^H \sum_{w=1}^W x_{n,c,h,w}, \quad \sigma_c^2 = \frac{1}{NHW} \sum_{n,h,w} (x_{n,c,h,w} - \mu_c)^2.
\end{equation}

The normalized data is defined as $\hat{x}_{n,c,h,w} = (x_{n,c,h,w} - \mu_c) / \sqrt{\sigma_c^2 + \epsilon}$. For a specific sample $x^{(n)}$, the squared Frobenius norm is given by:

\begin{equation}
\lVert \text{BN}(x^{(n)}) \rVert^2_F = \sum_{c=1}^C \sum_{h=1}^H \sum_{w=1}^W \frac{(x_{n,c,h,w} - \mu_c)^2}{\sigma_c^2 + \epsilon}.
\end{equation}

Taking the average across the entire batch $N$:

\begin{equation}
\frac{1}{N} \sum_{n=1}^N \lVert \text{BN}(x^{(n)}) \rVert^2_F = \sum_{c=1}^C \left( \frac{1}{N} \sum_{n,h,w} \frac{(x_{n,c,h,w} - \mu_c)^2}{\sigma_c^2 + \epsilon} \right) = \sum_{c=1}^C \left( HW \cdot \frac{\sigma_c^2}{\sigma_c^2 + \epsilon} \right).
\end{equation}

As $\epsilon \to 0$, the batch-averaged squared norm converges to $CHW = D$. However, for an individual sample $x^{(n)}$, $\lVert \text{BN}(x^{(n)}) \rVert^2_F$ is a random variable whose realization depends on the batch statistics. This implies that while LN imposes a rigid algebraic constraint on each sample, BN only enforces a statistical concentration around the Gaussian annulus, leading to the observed stochastic fluctuations in radial displacement. Within the high-dimensional latent space extracted by Wide-ResNet50, this discrepancy is further exacerbated by feature sparsity and the low intra-class variance inherent to simple industrial textures. For such samples, the batch-wise scaling factors tend to compress their intrinsic deviations from the mean, pulling the realized initial position $x_0$ closer to the origin.

In the following, we elucidate the key conceptual frameworks and terminology established in the primary manuscript.

\begin{propositionnum}{1} 
    (Centripetal Potential Field) If the Assumption 1 and 2 hold, the marginal vector field ${u}(x,t)$ of rFM points toward $\textbf{0}$ at $t=0$, and its magnitude is positively correlated with the sample's initial distance $d= \lVert {x_0} \rVert _2$.
    \label{pp:cpf}
\end{propositionnum}

\textit{Proof.} By the definition of the marginal vector field for linear probability paths in Lemma \ref{lm2}, we have 

\begin{equation}
    u(x, t) = \mathbb{E}[X_1 - X_0 \mid X_t = x] = \mathbb{E}[X_1 \mid X_t = x] - \mathbb{E}[X_0 \mid X_t = x].   
\end{equation}


As $t \to 0^+$, the linear interpolation $X_t = (1-t)X_0 + tX_1$ converges to $X_0$ in distribution. Thus, the conditioning $X_t = x$ asymptotically implies $X_0 = x$. Applying this limit to the second term yields

\begin{equation}
    \lim_{t \to 0^+} \mathbb{E}[X_0 \mid X_t = x] = \mathbb{E}[X_0 \mid X_0 = x] = x.
\end{equation}


For the first term, we invoke the independent coupling assumption ($X_1 \perp X_0$), which simplifies the conditional expectation to the marginal expectation of the prior,

\begin{equation}
    \lim_{t \to 0^+} \mathbb{E}[X_1 \mid X_t = x] = \mathbb{E}[X_1 \mid X_0 = x] = \mathbb{E}[X_1].
\end{equation}

Given the zero-centered Gaussian prior $\mathbb{E}[X_1] = \mathbf{0}$, substituting these limits back into the decomposition concludes:

\begin{equation}
    \lim_{t \to 0^+} u(x, t) = 0 - x = -x
\end{equation}

The identity $u(x,0) = -x$ indicates that the vector field at the initial boundary acts as a linear centripetal potential, directing samples toward the origin with magnitude $\|x_0\|_2$.






\begin{theorem}[Benamou–Brenier Theorem \cite{benamou1999numerical}]
     the Wasserstein-2 distance $W_2$ between a source distribution $\mu_0$ and a target distribution $\mu_1$ can be formulated as an action minimization problem over a time-dependent density $p_t$ and a vector field $u_t$:
     
     $$W_2^2(\mu_0, \mu_1) = \inf_{(p_t, v_t)} \int_0^1 \int_{\mathbb{R}^D} \frac{1}{2} \|u_t(x)\|^2 p_t(x) dx dt$$subject to the continuity equation $\partial_t p_t + \nabla \cdot (p_t u_t) = 0$, with boundary conditions $p_0 = \mu_0$ and $p_1 = \mu_1$.

     \label{th:BB}
\end{theorem}

The Benamou–Brenier theorem reformulates the geometric optimal transport problem of minimizing the Wasserstein-2 distance ($W_2$) into a dynamic energy minimization framework. It embeds the transport process into a continuous-time evolution governed by the continuity equation. In this formulation, the objective functional (or action) represents the cumulative kinetic energy of a time-varying velocity field $v_t$, weighted by the evolving density $p_t$. Minimizing this action yields the geodesic in the Wasserstein space, thereby recovering the optimal trajectory of mass transport.


\noindent \textbf{Local Worst transport mapping}: Let $\mathcal{M} \subset \mathbb{R}^D$ be a smooth $k$-dimensional embedded submanifold, and let $\mu, \nu$ be Borel probability measures supported exactly on $\mathcal{M}$, i.e., $\text{supp}(\mu), \text{supp}(\nu) \subseteq \mathcal{M}$. Consider the Kantorovich optimal transport problem with the quadratic cost $c(x,z) = \frac{1}{2}\|x-z\|^2$:

$$\mathcal{T}_c(\mu, \nu) = \inf_{\pi \in \Pi(\mu,\nu)} \int_{\mathcal{M} \times \mathcal{M}} \frac{1}{2}\|x-z\|^2 d\pi(x,z)$$

where $\Pi(\mu,\nu)$ denotes the set of joint distributions with marginals $\mu$ and $\nu$.

Degenerate Kantorovich Problem: The Kantorovich optimal transport problem with the extrinsic Euclidean cost $\|x-z\|^2$ asymptotically degenerates in the normal direction of $\mathcal{M}$. The optimal transport plan rearrange mass essentially along the tangent space, locally reducing the cost to the intrinsic Riemannian geodesic distance $d_\mathcal{M}(x,z)^2$.

Proof: Let $x, z \in \mathcal{M}$ be two sufficiently close points on the manifold. We can decompose the extrinsic displacement vector $z - x \in \mathbb{R}^D$ into its tangential and normal components relative to the tangent space $T_x\mathcal{M}$ and normal space $N_x\mathcal{M}$ at $x$:

$$z - x = v_{\parallel} + v_{\perp}$$

where $v_{\parallel} \in T_x\mathcal{M}$ and $v_{\perp} \in N_x\mathcal{M}$. By the Pythagorean theorem, the extrinsic squared cost is:

$$\|z - x\|^2 = \|v_{\parallel}\|^2 + \|v_{\perp}\|^2$$

To establish the relationship between these components and the intrinsic geometry, let $\gamma(s)$ be the unit-speed geodesic on $\mathcal{M}$ connecting $x$ and $z$, such that $\gamma(0) = x$ and $\gamma(d) = z$, where $d = d_\mathcal{M}(x,z)$ is the intrinsic geodesic distance. The initial velocity is a unit tangent vector $\gamma'(0) = u \in T_x\mathcal{M}$.

Applying the Taylor expansion to $\gamma(s)$ around $s=0$, we have:

$$z = \gamma(d) = x + d \cdot \gamma'(0) + \frac{d^2}{2} \gamma''(0) + \mathcal{O}(d^3)$$

Since $\gamma$ is a geodesic, its acceleration vector $\gamma''(0)$ has no tangential component and is completely determined by the Second Fundamental Form $II_x$ of the manifold: $\gamma''(0) = II_x(u, u) \in N_x\mathcal{M}$.Substituting this into the displacement equation yields:

$$z - x = \underbrace{d \cdot u + \mathcal{O}(d^3)_{\parallel}}_{v_{\parallel}} + \underbrace{\frac{d^2}{2} II_x(u,u) + \mathcal{O}(d^3)_{\perp}}_{v_{\perp}}$$

Evaluating the squared norms of the tangential and normal components, we obtain:

$$\|v_{\parallel}\|^2 = d^2 \|u\|^2 + \mathcal{O}(d^4) = d_\mathcal{M}(x,z)^2 + \mathcal{O}(d_\mathcal{M}(x,z)^4)$$

$$\|v_{\perp}\|^2 = \frac{d^4}{4} \|II_x(u,u)\|^2 + \mathcal{O}(d^5) = \mathcal{O}(d_\mathcal{M}(x,z)^4)$$This derivation explicitly reveals a critical geometric property: the normal displacement $v_{\perp}$ is a second-order effect driven by the manifold's curvature $II_x$, and its squared magnitude $\|v_{\perp}\|^2$ is of order $\mathcal{O}(d^4)$.

For the Kantorovich problem over any local coupling $\pi$, the expected cost can be decomposed as:$$\mathbb{E}_{(x,z)\sim\pi}[\|x-z\|^2] = \mathbb{E}[\|v_{\parallel}\|^2] + \mathbb{E}[\|v_{\perp}\|^2]$$

Because $\|v_{\perp}\|^2 = \mathcal{O}(\|v_{\parallel}\|^4)$, as $x \to z$, the contribution of the normal component vanishes at a significantly faster rate than the tangential component. Therefore, for measures supported strictly on $\mathcal{M}$, the normal cost contributes negligibly to the total transport cost asymptotically. The optimization problem intrinsically reduces to minimizing the transport cost over the tangent bundle:

$$\inf_{\pi \in \Pi(\mu,\nu)} \int_{\mathcal{M} \times \mathcal{M}} d_\mathcal{M}(x,z)^2 d\pi(x,z)$$

This demonstrates that the transport degenerates along the normal direction, mathematically precluding the formation of a normal-oriented displacement (such as the Centripetal Potential Field) under pure Optimal Transport dynamics on a shared manifold support.

\section{The Analyses}
\label{sec:as}

\subsection{Analysis on forward Process in FM}
\label{sec:as1}
The forward process maintains distinct advantages. Although Gaussian samples are still uniformly distributed on the spherical surface, their trajectories terminate in a more compact low-dimensional structure \cite{bengio2013representation}. The probability paths remain undisturbed, enabling reliable transport during generation. The OT-Flow circumvents terminal point handling by adding a minor noise perturbation into the target distribution, yielding the probability path
\begin{equation}
    p_t(x|x_1)=\mathcal{N}(x|tx_1,(1-(1-\epsilon)t)^2 I).
\end{equation}

This construction provides an $\epsilon$-approximation to the target solution. The RF circumvents the singularity at $t=1$ by using a smooth neural network function to approximate $x_1$.

\subsection{Quantitative analysis of trajectories}
\label{sec:as2}
Training in Flow Matching is a continuous-time process that involves sampling a time step $t \sim \mathcal{U}(0,1)$ and a target vector $x_1 \sim \mathcal{N}(\mathbf{0}, \mathbf{I})$ of the same dimension as the data $x_0$. The sample position at time $t$ is computed as $x_t = t x_0 + (1-t) x_1$. 

\noindent \textbf{The Target $\mathbf{x_1}$:} In this work, we extract features using a WideResNet-50 pretrained on ImageNet, yielding feature maps of size $(8, 1024, 16, 16)$. The corresponding Gaussian radius for this dimension, according to the Gaussian Annulus Theorem, is:

\begin{equation}
    R=\sqrt{CHW}=512.
\end{equation}

\noindent where $(B, C, H, W)$ denote batchsize, number of channels, height, and width, respectively. Note that for the ``transistor'' category, which uses $256\times256$ image inputs, the corresponding Gaussian radius is $256$, which is standard practice. The above radius value can be empirically verified by sampling a tensor of dimension $(8, 1024, 16, 16)$ using NumPy and computing its average Frobenius norm. 

\noindent \textbf{The Start $\mathbf{x_0}$:} In Sec. \ref{sec:pf3}, we analyze the effect of LayerNorm and BatchNorm on the data manifold.  BN's expectation of $\sqrt{d}$, though robust in mean, exhibits sample-wise stochasticity that strongly depends on channel correlations, while LN's deterministic  Frobenius norm is an algebraic fact for any correlation structure. As analyzed in Section \ref{sec:as4}, the feature maps extracted from WideResNet-50 reside within a hyperspherical shell of radius $R$, satisfying $d = ||x_0||_2 < R$. LN-normalized samples reside almost exactly on the Gaussian spherical surface (see \ref{sec:pf3} for analysis of their slight deviation below 512), whereas BN-produced samples show initial positions correlated with inter-sample variance, falling significantly below the Gaussian radius. The "Transistor" category uses a smaller input size, for which the corresponding Gaussian radius is 256, as shown in Table \ref{table11}.

\noindent \textbf{The Trajectory $\mathbf{x_t}$:} Tables \ref{table9}, \ref{table10} and \ref{table11} document the temporal evolution of axial positions for samples under rFM and WT-Flow, respectively. The first five categories in the table represent simple texture types. Under both configurations, the samples exhibit an initial tendency to move toward the origin, indicating that they are constrained by the geometric relationship as described, resulting in motion opposite to the optimal transport direction. Furthermore, we observe that this constraining effect is more pronounced in high-dimensional spaces compared to low-dimensional ones, with some samples remaining trapped near the source distribution even after a certain number of steps.

\subsection{Analysis of the Behaviors of Anomalies}
\label{sec:as8}

\begin{figure}[H]
\centering
\includegraphics[width=0.5\columnwidth]{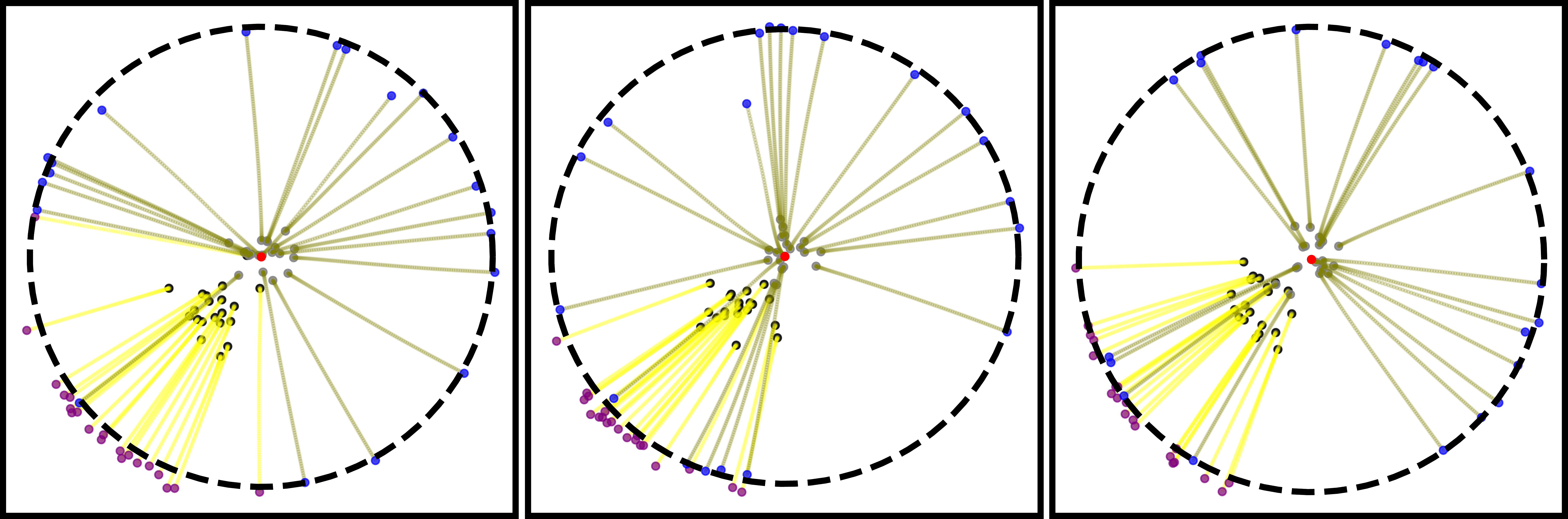} 
\caption{Outliers detection with intersecting supports, during different epoch. The oliver and the yellow lines represent the trajectories of the normals$\sim \mathcal{N}(0,I)$ and the abnormals$\sim \mathcal{N}((-1,-1), 0.3I)$, respectively.}
\label{fig7}
\end{figure}

This study categorizes anomalies into three types based on their positions in the manifold space: disjoint support, intersecting support, and overlapping support types. Theoretically, our method can effectively identify anomalous samples located outside the manifold support. The WT mapping is self-limiting, with its influence confined to the vicinity of the training distribution manifold; consequently, even in interleaved support scenarios, WT-Flow lacks the capacity to enforce constraints (Fig. \ref{fig7}). Similar to existing analyses \cite{zhang2021understanding}, our method also fails to identify overlapping support anomalies without explicit outlier definitions. However, by clearly defining the constrained objects, our approach offers a more flexible discrimination mechanism compared to traditional probabilistic modeling.

\subsection{Hyperparameters}
\label{sec:as3}
The training data are normalized using the mean and standard deviation values from ImageNet. In both the MVTec AD and VisA, the input image size was uniformly set to 512$\times$512 for all categories except ``transistor,'' which used a resolution of 256$\times$256, consistent with the experimental settings of other comparative methods. For the feature extractor, we employ a WideResNet50 model pre-trained on ImageNet, specifically utilizing the feature maps from its third block. For the ``macaroni2,'' we instead use feature maps from the second stage. During the training phase, the AdamW optimizer with a learning rate of 2e-4 and a weight decay of 0.1 was employed. The scheduler was adopted, with learning rate decay occurring every 20 epochs and a decay factor gamma of 0.1. The training process consisted of 100 training epochs, with a batch size of 8. The flow step during the inference phase was set to 20 for image-wise detection and 1 for pixel-wise localization. Performance metrics were computed by averaging the results over the last 10 epochs. As for the image-wise anomaly detection task, we calculate the anomaly score by averaging the largest $K$ anomaly scores in a anomaly map, choosing the $K=0.03$ of topK. Our WT mapping is implemented through a simple non-learnable LayerNorm. We report the mean and standard deviation over the final 10 training epochs to ensure statistical reliability and mitigate performance fluctuations.

\subsection{Quantitative analysis of Feature Extractor}
\label{sec:as4}
\begin{figure}[H]
\centering
\includegraphics[width=0.9\columnwidth]{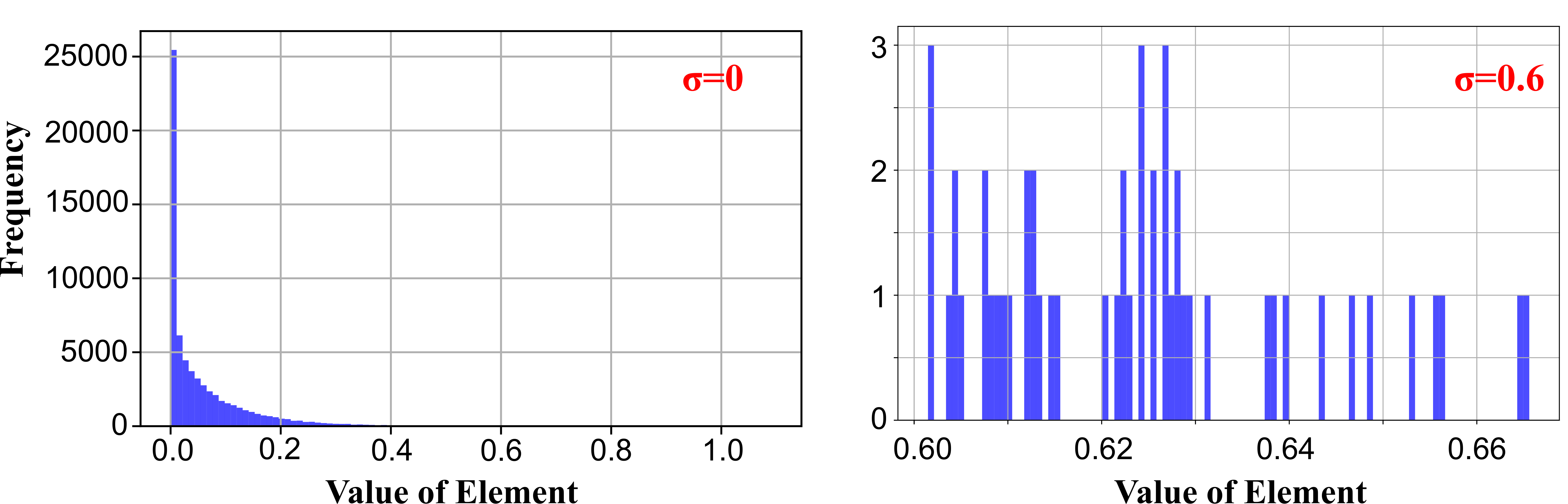} 
\caption{Activation histogram of a WideResNet-50 feature map. (top) Histogram with activation $\sigma=0.6$, demonstrating a distinct long-tail distribution with sparse non-zero activations. (bottom) Histogram with activation $\sigma=0$.}
\label{fig8}
\end{figure}

When defining the starting point $x_0$ and endpoint $x_1$ on the probability path, each sample's feature map is randomly paired with a (1024, 16, 16) Gaussian noise tensor, where the Gaussian radius is corresponding to 512. Crucially, even in the most extreme scenario where samples consist entirely of maximum activations (as illustrated in Fig.\ref{fig8} (bottom)), their spatial positions remain well within the Gaussian radius boundary.

\begin{table}[h]
\footnotesize
\centering

\setlength{\arrayrulewidth}{1pt}
\begin{tabular}{lcc}
    \hline 
    \hline
    Backbone           & WT-Flow(BN)       & WT-Flow(LN)   \\
    \hline
    ResNet18           &97.90/97.71     &98.24/97.72        \\
    ResNet34           &97.81/97.58     &98.31/97.62       \\
    WideResnet50       &98.57/97.68     &99.05/97.95         \\
    \hline
    \hline
\end{tabular}
\caption{Ablation study of feature extractor, with Det./Loc. mAUC on MVTec AD.}
\label{table7}
\end{table}

Our experiments demonstrate robustness to different backbone architectures, with all models (ResNet18/34 and WideResNet50) achieving comparable performance (detection: 98.24-99.05\%, localization: 97.72-97.95\%). The maximal variation ($\sigma$ $<$ 0.8\%) confirms the insensitivity of our method to the feature extractor choice.

\subsection{Supplementary Tables}

\label{sec:as9}

Comparative results for anomaly detection and localization on MVTec AD are shown in Tables \ref{table12} and \ref{table13}, using image-level and pixel-level AUROC metrics respectively, with each entry representing the average of the final 10 epochs per sub-category. Tables \ref{table14} and \ref{table15} present corresponding results on the VisA dataset, also using image-wise and pixel-wise AUROC for detection and localization, with each entry being the average of the last 10 epochs per sub-category. Tables \ref{table9} to \ref{table11} provide high-dimensional empirical evidence regarding the evolution of radial displacement $\lVert x_t \rVert_2$ across the 15 MVTec classes. Tables \ref{tab16} and \ref{tab17} provide detailed per-category image-level classification and pixel-level localization results in the ablation study.

\begin{table*}[h]
\centering
\footnotesize
\setlength{\arrayrulewidth}{1pt}
\begin{tabular}{ccccccccc}
    \hline
    \hline
        &RD++$^\dagger$\cite{tien2023revisiting} &CSFlow$^\dagger$\cite{rudolph2022fully}  & FastFlow$^\dagger$\cite{yu2021fastflow}     & CFlow$^\ddagger$\cite{gudovskiy2022cflow}   & MSFlow$^\ddagger$\cite{zhou2024msflow}        & rFM  & WT-Flow(BN) & WT-Flow(LN)	    \\
    \hline
        Carpet      &98.78$\pm$0.12    &100.0$\pm$0.00   &94.32$\pm$1.76  &99.31$\pm$0.01  &97.11$\pm$0.82  &100.0$\pm$0.03  &100.0$\pm$0.01  &99.92$\pm$0.00     \\
        Grid        &99.46$\pm$1.60    &97.74$\pm$0.45   &99.81$\pm$0.06  &96.30$\pm$0.01  &99.58$\pm$0.13  &99.92$\pm$0.07  &99.75$\pm$0.00  &99.42$\pm$0.04     \\
        Leather     &100.0$\pm$0.00    &100.0$\pm$0.00   &100.0$\pm$0.00  &100.0$\pm$0.00  &100.0$\pm$0.00  &100.0$\pm$0.00  &100.0$\pm$0.00  &100.0$\pm$0.00     \\
        Tile        &99.04$\pm$0.25    &99.20$\pm$0.04   &99.98$\pm$0.02  &98.71$\pm$0.01  &99.71$\pm$0.14  &99.86$\pm$0.03  &100.0$\pm$0.00  &100.0$\pm$0.00     \\
        Wood        &99.13$\pm$0.16    &99.50$\pm$0.17   &99.56$\pm$0.25  &98.42$\pm$0.00  &87.28$\pm$1.22  &99.39$\pm$0.08  &99.30$\pm$0.00  &99.39$\pm$0.00     \\
        Bottle      &99.99$\pm$0.03    &99.48$\pm$0.04   &99.97$\pm$0.08  &100.0$\pm$0.00  &100.0$\pm$0.00  &99.92$\pm$0.04  &100.0$\pm$0.00  &100.0$\pm$0.00     \\
        Cable       &97.46$\pm$0.36    &95.96$\pm$0.42   &95.39$\pm$0.43  &88.58$\pm$0.01  &98.89$\pm$0.14  &95.33$\pm$0.08  &98.24$\pm$0.01  &99.18$\pm$0.02     \\
        Capsule     &97.41$\pm$0.37    &92.78$\pm$0.41   &98.80$\pm$0.14  &95.16$\pm$0.01  &98.64$\pm$0.14  &89.07$\pm$0.17  &98.17$\pm$0.04  &99.00$\pm$0.02     \\
        Hazelnut    &100.0$\pm$0.00    &97.27$\pm$0.18   &96.96$\pm$0.22  &98.64$\pm$0.01  &99.93$\pm$0.02  &99.75$\pm$0.00  &100.0$\pm$0.00  &100.0$\pm$0.00     \\
        Metal Nut   &100.0$\pm$0.00    &96.50$\pm$0.24   &99.14$\pm$0.73  &98.46$\pm$0.01  &100.0$\pm$0.00  &97.02$\pm$0.28  &99.85$\pm$0.00  &99.95$\pm$0.00     \\
        Pill        &97.23$\pm$0.38    &97.37$\pm$0.15   &98.22$\pm$0.21  &88.84$\pm$0.01  &99.07$\pm$0.21  &92.69$\pm$0.14  &97.30$\pm$0.02  &97.35$\pm$0.05     \\
        Screw       &97.78$\pm$0.43    &95.40$\pm$0.46   &89.89$\pm$1.23  &84.06$\pm$0.01  &97.75$\pm$0.20  &57.84$\pm$0.52  &94.55$\pm$0.06  &94.67$\pm$0.04     \\
        Toothbrush  &97.24$\pm$1.44    &88.98$\pm$0.97   &99.63$\pm$0.28  &89.72$\pm$0.00  &100.0$\pm$0.00  &91.67$\pm$0.32  &92.22$\pm$0.00  &97.50$\pm$0.00     \\
        Transistor  &96.49$\pm$0.39    &96.62$\pm$0.24   &98.74$\pm$0.28  &95.59$\pm$0.01  &99.75$\pm$0.05  &96.42$\pm$0.22  &99.58$\pm$0.00  &99.83$\pm$0.00     \\
        Zipper      &98.79$\pm$0.33    &99.07$\pm$0.10   &97.77$\pm$0.10  &99.08$\pm$0.00  &99.79$\pm$0.05  &99.21$\pm$0.08  &99.58$\pm$0.01  &99.55$\pm$0.02     \\
        \hline
        Det.       &98.59$\pm0.39$     &97.06$\pm$0.26    &97.88$\pm0.39$      &95.39$\pm0.01$   &98.50$\pm0.21$     &94.54$\pm0.14$  &98.57$\pm0.01$ &\textbf{99.05}$\pm$\textbf{0.01}    \\
    \hline
    \hline
\end{tabular}
\caption{Anomaly detection performance on the MVTec AD dataset with mean AUROC per subclass over final 10 epochs. Keys: $\dagger$ (evaluated under their multi-scale setting), $\ddagger$ (our single-scale setup).}
\label{table12}
\end{table*}

\begin{table*}[h]
\centering
\footnotesize
\setlength{\arrayrulewidth}{1pt}
\begin{tabular}{ccccccccc}
    \hline
    \hline
        &RD++$^\dagger$\cite{tien2023revisiting}   &CSFlow$^\dagger$\cite{rudolph2022fully}  & FastFlow$^\dagger$\cite{yu2021fastflow}     & CFlow$^\ddagger$\cite{gudovskiy2022cflow}   & MSFlow$^\ddagger$\cite{zhou2024msflow}        & rFM  & WT-Flow(BN)  & WT-Flow(LN)	    \\
    \hline
        Carpet      &99.00$\pm$0.00   &-   &98.10$\pm$0.80  &98.55$\pm$0.00  &98.08$\pm$0.03  &98.63$\pm$0.01  &98.86$\pm$0.01  &98.95$\pm$0.01     \\
        Grid        &98.29$\pm$1.44   &-   &99.26$\pm$0.02  &96.11$\pm$0.01  &98.05$\pm$0.03  &97.69$\pm$0.04  &98.18$\pm$0.00  &98.22$\pm$0.01     \\
        Leather     &99.40$\pm$0.00   &-   &99.60$\pm$0.01  &98.99$\pm$0.00  &98.53$\pm$0.02  &98.57$\pm$0.01  &98.77$\pm$0.01  &98.86$\pm$0.01     \\
        Tile        &95.74$\pm$0.05   &-   &96.17$\pm$0.15  &96.36$\pm$0.00  &95.76$\pm$0.08  &93.57$\pm$0.07  &95.90$\pm$0.01  &96.16$\pm$0.01     \\
        Wood        &95.57$\pm$0.10   &-   &95.52$\pm$0.53  &93.48$\pm$0.00  &86.29$\pm$0.08  &91.51$\pm$0.14  &92.34$\pm$0.02  &93.77$\pm$0.02     \\
        Bottle      &98.70$\pm$0.07   &-   &98.37$\pm$0.32  &97.49$\pm$0.00  &97.31$\pm$0.01  &97.66$\pm$0.02  &97.91$\pm$0.00  &97.91$\pm$0.01     \\
        Cable       &97.52$\pm$0.07   &-   &96.63$\pm$0.21  &95.75$\pm$0.01  &97.82$\pm$0.01  &94.77$\pm$0.08  &97.04$\pm$0.01  &98.17$\pm$0.00     \\
        Capsule     &98.69$\pm$0.06   &-   &98.67$\pm$0.04  &98.10$\pm$0.00  &98.68$\pm$0.01  &98.07$\pm$0.01  &98.50$\pm$0.01  &98.76$\pm$0.00     \\
        Hazelnut    &98.94$\pm$0.05   &-   &96.87$\pm$0.27  &98.21$\pm$0.00  &97.85$\pm$0.02  &97.37$\pm$0.01  &98.19$\pm$0.01  &98.43$\pm$0.00     \\
        Metal Nut   &97.40$\pm$0.07   &-   &98.18$\pm$0.21  &96.58$\pm$0.01  &97.47$\pm$0.02  &93.87$\pm$0.05  &96.73$\pm$0.01  &97.32$\pm$0.01     \\
        Pill        &98.32$\pm$0.07   &-   &96.96$\pm$0.13  &97.20$\pm$0.01  &98.53$\pm$0.03  &95.87$\pm$0.07  &98.21$\pm$0.03  &98.59$\pm$0.03     \\
        Screw       &99.60$\pm$0.00   &-   &97.98$\pm$0.17  &96.33$\pm$0.00  &98.70$\pm$0.03  &95.89$\pm$0.06  &98.77$\pm$0.02  &98.84$\pm$0.02     \\
        Toothbrush  &99.10$\pm$0.00   &-   &97.93$\pm$0.05  &98.24$\pm$0.00  &98.24$\pm$0.02  &98.60$\pm$0.01  &98.71$\pm$0.00  &98.83$\pm$0.00     \\
        Transistor  &92.42$\pm$0.47   &-   &96.40$\pm$0.22  &91.27$\pm$0.01  &96.38$\pm$0.04  &98.26$\pm$0.03  &98.22$\pm$0.01  &98.22$\pm$0.01     \\
        Zipper      &98.47$\pm$0.10   &-   &98.53$\pm$0.02  &97.59$\pm$0.00  &98.29$\pm$0.02  &97.99$\pm$0.07  &98.27$\pm$0.02  &98.24$\pm$0.01     \\
        \hline
        Loc.        &97.81$\pm$0.17   &-        &97.68$\pm0.21$  &96.81$\pm0.01$  &97.06$\pm0.03$  &96.55$\pm0.05$  &97.64$\pm0.02$  &\textbf{97.95}$\pm$\textbf{0.01}    \\
    \hline
    \hline
\end{tabular}
\caption{Anomaly localization performance on the MVTec AD dataset with mean AUROC per subclass over final 10 epochs. Keys: $\dagger$ (evaluated under their multi-scale setting), $\ddagger$ (our single-scale setup).}
\label{table13}
\end{table*}

\begin{table*}[t]
\centering
\footnotesize
\setlength{\arrayrulewidth}{1pt}
\begin{tabular}{cccccccccc}
    \hline
    \hline
                   &RD$^\dagger$\cite{deng2022anomaly}   &RD++$^\dagger$\cite{tien2023revisiting}       &Fastflow$^\dagger$\cite{yu2021fastflow}       & CFlow$^\ddagger$\cite{gudovskiy2022cflow}   & MSFlow$^\ddagger$\cite{zhou2024msflow}        & rFM  & WT-Flow(BN)  & WT-Flow(LN)    \\
    \hline
    candle         &95.24$\pm$0.70   &95.92$\pm$0.37   &97.20$\pm$0.79   &90.12$\pm$0.01   &96.15$\pm$0.38     &95.62$\pm$0.09  &98.09$\pm$0.29  &98.07$\pm$0.09   \\
    capsules       &90.16$\pm$1.19   &88.76$\pm$0.52   &81.54$\pm$2.14   &93.61$\pm$0.01   &89.54$\pm$0.44     &79.32$\pm$0.28  &89.72$\pm$0.09  &93.12$\pm$0.15   \\
    cashhew        &95.92$\pm$2.47   &97.09$\pm$0.31   &94.51$\pm$2.84   &95.02$\pm$0.03   &95.27$\pm$0.29     &96.13$\pm$0.13  &96.26$\pm$0.12  &96.68$\pm$0.18   \\
    chewinggum     &98.16$\pm$0.64   &98.69$\pm$0.47   &99.41$\pm$0.37   &97.32$\pm$0.01   &98.64$\pm$0.12     &98.88$\pm$0.03  &99.41$\pm$0.03  &99.58$\pm$0.04   \\
    fryum          &91.23$\pm$6.29   &94.95$\pm$0.37   &95.71$\pm$1.19   &91.60$\pm$0.01   &98.08$\pm$0.17     &91.58$\pm$0.10  &97.22$\pm$0.09  &97.46$\pm$0.09   \\
    marcaron1      &98.30$\pm$0.33   &96.31$\pm$0.25   &88.00$\pm$2.71   &82.18$\pm$0.01   &93.92$\pm$0.37     &84.81$\pm$0.61  &95.84$\pm$0.06  &97.65$\pm$0.19   \\
    marcaron2      &88.03$\pm$2.33   &87.43$\pm$0.66   &76.66$\pm$3.46   &59.75$\pm$0.01   &82.49$\pm$0.45     &68.84$\pm$0.26  &80.39$\pm$0.06  &85.35$\pm$0.16   \\
    pcb1           &96.99$\pm$0.24   &96.46$\pm$0.19   &91.65$\pm$0.98   &91.17$\pm$0.01   &97.38$\pm$0.12     &95.69$\pm$0.33  &96.67$\pm$0.02  &97.44$\pm$0.01   \\
    pcb2           &96.37$\pm$1.43   &96.02$\pm$0.28   &90.13$\pm$0.57   &85.41$\pm$0.01   &95.35$\pm$0.41     &92.28$\pm$0.21  &96.41$\pm$0.03  &96.47$\pm$0.07   \\
    pcb3           &96.87$\pm$0.29   &96.06$\pm$0.24   &87.39$\pm$1.23   &80.82$\pm$0.01   &96.24$\pm$0.22     &87.82$\pm$0.32  &97.34$\pm$0.05  &97.44$\pm$0.05   \\
    pcb4           &99.86$\pm$0.05   &99.77$\pm$0.04   &98.11$\pm$0.26   &89.37$\pm$0.02   &98.38$\pm$0.20     &99.53$\pm$0.01  &99.66$\pm$0.01  &99.67$\pm$0.02   \\
    pipefryum      &99.26$\pm$0.23   &99.64$\pm$0.10   &97.96$\pm$0.19   &94.80$\pm$0.03   &99.26$\pm$0.08     &96.98$\pm$0.11  &99.26$\pm$0.04  &99.36$\pm$0.08   \\
    \hline
       Det.        &95.53$\pm$1.35   &95.59$\pm$0.32   &91.52$\pm1.39$   &87.60$\pm$0.01   &95.06$\pm0.27$     &90.62$\pm0.21$  &95.52$\pm$0.07  &\textbf{96.52}$\pm$0.09    \\
    \hline
    \hline
\end{tabular}
\caption{Anomaly detection performance on the VisA dataset with mean AUROC per subclass over final 10 epochs. Keys: $\dagger$ (evaluated under their multi-scale setting), $\ddagger$ (our single-scale setup).}
\label{table14}
\end{table*}

\begin{table*}[t]
\centering
\footnotesize
\setlength{\arrayrulewidth}{1pt}
\begin{tabular}{cccccccccc}
    \hline
    \hline
                   &RD$^\dagger$\cite{deng2022anomaly}   &RD++$^\dagger$\cite{tien2023revisiting}       &Fastflow$^\dagger$\cite{yu2021fastflow}       & CFlow$^\ddagger$\cite{gudovskiy2022cflow}   & MSFlow$^\ddagger$\cite{zhou2024msflow}        & rFM  & WT-Flow(BN)  & WT-Flow(LN)    \\
    \hline
    candle         &98.84$\pm$0.07   &98.53$\pm$0.06    &98.27$\pm$0.19   &97.42$\pm$0.00   &98.43$\pm$0.04  &98.65$\pm$0.01  &98.86$\pm$0.00  &99.12$\pm$0.00   \\
    capsules       &99.40$\pm$0.00   &99.33$\pm$0.03    &97.99$\pm$1.03   &68.24$\pm$0.02   &98.94$\pm$0.05  &97.04$\pm$0.06  &99.04$\pm$0.02  &99.53$\pm$0.02   \\
    cashhew        &94.63$\pm$0.67   &94.80$\pm$0.31    &98.35$\pm$0.85   &99.10$\pm$0.00   &98.88$\pm$0.02  &99.06$\pm$0.03  &98.50$\pm$0.07  &98.66$\pm$0.07   \\
    chewinggum     &97.60$\pm$0.11   &97.70$\pm$0.05    &98.87$\pm$0.09   &98.32$\pm$0.00   &98.33$\pm$0.01  &99.12$\pm$0.01  &99.07$\pm$0.00  &99.09$\pm$0.00   \\
    fryum          &96.37$\pm$0.39   &96.02$\pm$0.13    &93.23$\pm$2.96   &96.02$\pm$0.00   &90.13$\pm$0.46  &96.07$\pm$0.02  &94.81$\pm$0.10  &95.44$\pm$0.10   \\
    marcaron1      &99.37$\pm$0.11   &99.24$\pm$0.09    &98.68$\pm$0.57   &98.07$\pm$0.00   &98.76$\pm$0.05  &98.50$\pm$0.08  &99.15$\pm$0.01  &99.83$\pm$0.01   \\
    marcaron2      &99.31$\pm$0.06   &99.17$\pm$0.03    &96.13$\pm$1.23   &95.54$\pm$0.00   &96.41$\pm$0.23  &95.86$\pm$0.16  &98.46$\pm$0.01  &99.60$\pm$0.01   \\
    pcb1           &99.70$\pm$0.00   &99.67$\pm$0.02    &98.91$\pm$0.12   &98.96$\pm$0.00   &99.64$\pm$0.01  &99.33$\pm$0.04  &99.66$\pm$0.00  &99.66$\pm$0.00   \\
    pcb2           &98.52$\pm$0.11   &98.68$\pm$0.03    &97.24$\pm$0.12   &96.78$\pm$0.01   &98.28$\pm$0.04  &98.12$\pm$0.05  &98.78$\pm$0.01  &98.79$\pm$0.01   \\
    pcb3           &99.20$\pm$0.00   &99.14$\pm$0.01    &97.36$\pm$0.58   &96.98$\pm$0.01   &98.64$\pm$0.03  &98.43$\pm$0.02  &98.89$\pm$0.00  &98.92$\pm$0.00   \\
    pcb4           &98.41$\pm$0.13   &98.30$\pm$0.08    &95.86$\pm$0.10   &98.06$\pm$0.00   &98.20$\pm$0.05  &97.67$\pm$0.02  &98.84$\pm$0.00  &98.85$\pm$0.00   \\
    pipefryum      &98.90$\pm$0.07   &99.05$\pm$0.03    &99.03$\pm$0.02   &99.21$\pm$0.00   &98.89$\pm$0.01  &99.34$\pm$0.01  &99.00$\pm$0.01  &99.28$\pm$0.01   \\
    \hline
       Loc.        &98.35$\pm0.14$  &98.30$\pm0.07$   &97.49$\pm0.68$   &95.23$\pm$0.00   &97.79$\pm0.08$  &98.10$\pm0.04$  &98.59$\pm0.02$  &\textbf{98.90}$\pm$0.03    \\
    \hline
    \hline
\end{tabular}
\caption{Anomaly localization performance on the VisA dataset with mean AUROC per subclass over final 10 epochs. Keys: $\dagger$ (evaluated under their multi-scale setting), $\ddagger$ (our single-scale setup)}
\label{table15}
\end{table*}

\clearpage

\begin{table*}[t]
\centering
\footnotesize
\setlength{\arrayrulewidth}{1pt}
\begin{tabular}{cccccccccccc}
    \hline
    ODE Step            &0      &5     &10    &15   &20   &25    &30	 &35  &40  &45    &50\\
    \hline
    Carpet      &54.9 &49.8 &44.9 &40.7 &37.1 &34.4 &32.7 &\textbf{32.2} &32.6 &33.6 &36.0    \\
    Grid        &53.7 &49.3 &45.6 &43.6 &\textbf{43.4} &44.3 &45.0 &46.6 &49.2 &52.9 &57.6    \\
    Leather     &52.3 &47.9 &43.9 &\textbf{41.4} &41.6 &44.0 &46.5 &48.7 &51.5 &55.3 &59.8    \\
    Tile        &48.0 &44.0 &40.2 &36.5 &33.1 &30.1 &27.4 &25.4 &24.0 &\textbf{23.5} &25.4    \\
    Wood        &43.5 &40.4 &\textbf{39.2} &42.5 &48.3 &54.5 &59.0 &63.2 &67.8 &72.9 &78.0    \\
    \hline
    Bottle      &45.6 &42.6 &39.8 &37.2 &35.1 &33.5 &32.4 &\textbf{32.1} &32.4 &33.1 &35.2    \\
    Cable       &50.1 &47.4 &\textbf{45.8} &46.1 &49.2 &54.0 &58.4 &62.0 &65.7 &70.0 &74.7    \\
    Capsule     &42.8 &40.0 &38.3 &\textbf{37.8} &39.4 &42.3 &46.2 &49.9 &53.7 &58.0 &62.6    \\
    Hazelnut    &44.6 &\textbf{42.8} &43.3 &47.7 &56.0 &65.6 &72.4 &79.9 &88.4 &97.8 &108.3    \\
    Metal Nut   &47.8 &45.4 &\textbf{44.0} &44.5 &47.4 &52.5 &57.7 &62.0 &66.2 &70.7 &75.7    \\
    Pill        &44.2 &41.8 &\textbf{40.3} &40.6 &44.0 &50.7 &57.5 &63.0 &68.1 &74.0 &80.8    \\
    Screw       &40.9 &39.2 &\textbf{39.2} &42.2 &48.0 &53.2 &57.0 &61.2 &66.2 &72.0 &78.5    \\
    Toothbrush  &43.2 &40.3 &37.6 &35.0 &32.6 &30.5 &28.8 &27.5 &26.8 &\textbf{26.5} &27.8    \\
    Transistor  &24.4 &22.6 &21.1 &20.2 &\textbf{20.0} &20.4 &21.0 &21.7 &22.6 &23.7 &25.0     \\
    Zipper      &47.9 &44.0 &40.4 &\textbf{39.0} &39.8 &41.8 &45.1 &48.2 &51.6 &55.1 &58.9     \\
    \hline
\end{tabular}
\caption{(rFM) The mean values of $||x_t||_2$ over different ODE steps for the ``good'' class in the test set of each category in the MVTec dataset, with the minimum values highlighted in bold, indicating closer proximity to the origin. Except for ``Transistor''. The corresponding Gaussian radius is $\sqrt{D}=512$.}
\label{table9}
\end{table*}

\begin{table*}[t]
\centering
\footnotesize
\setlength{\arrayrulewidth}{1pt}
\begin{tabular}{cccccccccccc}
    \hline
    ODE Step    &0      &5     &10    &15   &20   &25    &30	 &35  &40  &45    &50\\
    \hline
    Carpet      &90.6   &\textbf{89.3}   &89.8   &93.5   &100.7  &109.9  &118.7  &126.7  &134.1  &141.5  &149.5   \\
    Grid        &114.0  &107.5  &100.7  &95.1   &91.5   &\textbf{89.9}   &90.1   &91.1   &93.1   &95.7   &99.1   \\
    Leather     &80.5   &\textbf{79.5}   &79.6   &82.4   &88.0   &95.5   &103.3  &110.4  &116.6  &122.6  &129.0   \\
    Tile        &80.0   &\textbf{79.6}   &80.9   &84.4   &90.5   &98.4   &106.8  &114.9  &122.3  &128.9  &135.6   \\
    Wood        &83.3   &82.6   &\textbf{82.5}   &84.6   &89.0   &95.2   &101.9  &108.2  &113.7  &119.0  &124.4   \\
    \hline
    Bottle      &112.8  &104.3  &97.1   &91.7   &88.8   &\textbf{88.5}   &89.9   &92.2   &94.5   &97.6   &102.1   \\
    Cable       &133.5  &126.7  &121.7  &\textbf{119.3}  &119.9  &123.6  &129.4  &136.2  &143.2  &150.6  &158.4   \\
    Capsule     &105.9  &98.7   &93.2   &89.9   &\textbf{89.5}   &92.2   &96.6   &101.1  &105.3  &109.5  &114.6   \\
    Hazelnut    &265.2  &249.2  &234.7  &223.0  &215.3  &\textbf{212.3}  &213.7  &219.0  &227.6  &239.2  &253.5    \\
    Metal Nut   &130.0  &122.6  &116.7  &112.9  &\textbf{112.3}  &114.7  &119.3  &125.2  &131.2  &137.3  &144.0   \\
    Pill        &148.2  &139.4  &133.3  &\textbf{130.6}  &132.1  &137.7  &146.1  &156.3  &166.9  &177.7  &188.9   \\
    Screw       &173.8  &162.8  &153.6  &147.2  &\textbf{144.8}  &146.8  &152.5  &160.7  &170.5  &181.2  &192.7   \\
    Toothbrush  &46.1   &44.3   &42.6   &41.2   &39.9   &38.9   &38.0   &37.3   &\textbf{36.9}   &37.0   &38.4   \\
    Transistor  &54.4   &51.1   &47.9   &45.4   &43.8   &43.1   &\textbf{43.1}   &43.6   &44.6   &46.0   &47.8    \\
    Zipper      &116.5  &107.3  &99.9   &95.1   &\textbf{93.8}   &95.8   &100.3  &105.6  &112.2  &118.9  &126.2    \\
    \hline
\end{tabular}
\caption{(WT-Flow(BN)) The mean values of $||x_t||_2$ over different ODE steps for the ``good'' class in the test set of each category in the MVTec dataset, with the minimum values highlighted in bold, indicating closer proximity to the origin. Except for ``Transistor''. The corresponding Gaussian radius is $\sqrt{D}=512$.}
\label{table10}
\end{table*}

\begin{table*}[t]
\centering
\footnotesize
\setlength{\arrayrulewidth}{1pt}
\begin{tabular}{cccccccccccc}
    \hline
    ODE Step    &0      &5     &10    &15   &20   &25    &30	 &35  &40  &45    &50\\
    \hline
    Carpet      &509.3  &498.7  &490.9  &486.7  &\textbf{486.4}  &490.2  &498.1  &510.0  &525.8  &545.3  &563.7   \\
    Grid        &509.5  &500.5  &494.9  &\textbf{493.5}  &496.6  &504.3  &516.5  &533.1  &553.9  &578.5  &601.1  \\
    Leather     &507.6  &488.9  &472.3  &458.8  &449.0  &443.2  &\textbf{441.9}  &444.9  &452.3  &464.3  &477.4   \\
    Tile        &509.0  &498.0  &490.1  &\textbf{486.1}  &486.3  &490.7  &499.5  &512.6  &529.6  &550.7  &570.4   \\
    Wood        &509.9  &502.7  &\textbf{499.3}  &500.7  &507.1  &518.5  &534.5  &554.9  &579.4  &607.6  &632.7   \\
    \hline
    Bottle      &506.9  &484.8  &465.6  &450.6  &440.3  &\textbf{435.1}  &435.5  &441.4  &452.8  &469.3  &485.9   \\
    Cable       &509.0  &498.4  &\textbf{494.0}  &496.5  &505.6  &521.5  &543.8  &572.4  &606.3  &645.1  &678.8   \\
    Capsule     &508.0  &492.0  &479.0  &469.9  &465.1  &\textbf{464.9}  &469.4  &478.6  &492.3  &510.5  &528.2   \\
    Hazelnut    &507.3  &487.7  &470.6  &457.2  &448.0  &\textbf{443.7}  &444.7  &450.9  &462.4  &478.7  &495.2    \\
    Metal Nut   &507.5  &488.4  &472.3  &460.4  &453.3  &\textbf{451.3}  &454.7  &463.6  &477.9  &497.0  &515.5   \\
    Pill        &507.5  &488.7  &472.7  &460.4  &452.4  &\textbf{449.0}  &450.6  &457.2  &468.7  &485.0  &501.3   \\
    Screw       &508.5  &495.2  &485.3  &479.8  &\textbf{479.1} &483.1  &492.0  &505.4  &523.2  &545.3  &566.0   \\
    Toothbrush  &507.9  &491.6  &478.3  &469.2  &\textbf{464.5}  &464.7  &469.7  &479.6  &494.2  &513.5  &532.1  \\
    Transistor  &254.7  &250.3  &247.9  &\textbf{247.9}  &250.4  &255.4  &262.7  &272.4  &284.1  &297.9  &310.2   \\
    Zipper      &509.2  &498.6  &491.7  &\textbf{489.2}  &491.5  &498.6  &510.5  &526.9  &547.6  &572.4  &595.1    \\
    \hline
\end{tabular}
\caption{((WT-Flow(LN)) The mean values of $||x_t||_2$ over different ODE steps for the ``good'' class in the test set of each category in the MVTec dataset, with the minimum values highlighted in bold, indicating closer proximity to the origin. Except for ``Transistor''. The corresponding Gaussian radius is $\sqrt{D}=512$.}
\label{table11}
\end{table*}

\clearpage

\begin{table*}[ht]
\centering
\footnotesize
\setlength{\arrayrulewidth}{1pt}
\begin{tabular}{c|c|c|c|c|c|c|c|c|c|c|c|c|c|c}
    \hline
    \hline
    WT map        &\multicolumn{5}{c|}{w/o}   &\multicolumn{3}{c|}{BatchNorm}                     &\multicolumn{6}{c}{LayerNorm}\\     
    
    \hline
    NC          &128    &256  &512  &\multicolumn{2}{c|}{768}   &512  &\multicolumn{2}{c|}{768}   &128    &256  &\multicolumn{2}{c|}{512}  &\multicolumn{2}{c}{768}    \\
    
    \hline
    pos         &\multicolumn{4}{c|}{}   &$\bm{\checkmark}$   &\multicolumn{2}{c|}{}  &$\bm{\checkmark}$  &\multicolumn{3}{c|}{}    &$\bm{\checkmark}$  & &$\bm{\checkmark}$   \\ 
    \hline
    \multicolumn{1}{c}{Carpet    }   &\multicolumn{1}{c}{100.}  &\multicolumn{1}{c}{100.}  &\multicolumn{1}{c}{100.}  &\multicolumn{1}{c}{100.}  &\multicolumn{1}{c}{100.}  &\multicolumn{1}{c}{99.9}  &\multicolumn{1}{c}{99.9}  &\multicolumn{1}{c}{100.}  &\multicolumn{1}{c}{99.9}  &\multicolumn{1}{c}{99.8}  &\multicolumn{1}{c}{99.9}  &\multicolumn{1}{c}{99.7}  &\multicolumn{1}{c}{99.9}  &\multicolumn{1}{c}{99.9}    \\
    \multicolumn{1}{c}{Grid      }   &\multicolumn{1}{c}{83.6}  &\multicolumn{1}{c}{96.1}  &\multicolumn{1}{c}{99.0}  &\multicolumn{1}{c}{99.5}  &\multicolumn{1}{c}{99.9}  &\multicolumn{1}{c}{99.4}  &\multicolumn{1}{c}{99.7}  &\multicolumn{1}{c}{99.8}  &\multicolumn{1}{c}{96.5}  &\multicolumn{1}{c}{98.4}  &\multicolumn{1}{c}{99.1}  &\multicolumn{1}{c}{99.2}  &\multicolumn{1}{c}{99.4}  &\multicolumn{1}{c}{99.5}    \\
    \multicolumn{1}{c}{Leather   }   &\multicolumn{1}{c}{100.}  &\multicolumn{1}{c}{100.}  &\multicolumn{1}{c}{100.}  &\multicolumn{1}{c}{100.}  &\multicolumn{1}{c}{100.}  &\multicolumn{1}{c}{100.}  &\multicolumn{1}{c}{100.}  &\multicolumn{1}{c}{100.}  &\multicolumn{1}{c}{99.9}  &\multicolumn{1}{c}{100.}  &\multicolumn{1}{c}{100.}  &\multicolumn{1}{c}{100.}  &\multicolumn{1}{c}{100.}  &\multicolumn{1}{c}{100.}    \\
    \multicolumn{1}{c}{Tile      }   &\multicolumn{1}{c}{100.}  &\multicolumn{1}{c}{100.}  &\multicolumn{1}{c}{100.}  &\multicolumn{1}{c}{99.9}  &\multicolumn{1}{c}{99.9}  &\multicolumn{1}{c}{100.}  &\multicolumn{1}{c}{100.}  &\multicolumn{1}{c}{100.}  &\multicolumn{1}{c}{100.}  &\multicolumn{1}{c}{100.}  &\multicolumn{1}{c}{100.}  &\multicolumn{1}{c}{100.}  &\multicolumn{1}{c}{100.}  &\multicolumn{1}{c}{100.}    \\
    \multicolumn{1}{c}{Wood      }   &\multicolumn{1}{c}{98.0}  &\multicolumn{1}{c}{98.3}  &\multicolumn{1}{c}{99.2}  &\multicolumn{1}{c}{99.3}  &\multicolumn{1}{c}{99.4}  &\multicolumn{1}{c}{99.3}  &\multicolumn{1}{c}{99.3}  &\multicolumn{1}{c}{99.3}  &\multicolumn{1}{c}{99.4}  &\multicolumn{1}{c}{99.4}  &\multicolumn{1}{c}{99.4}  &\multicolumn{1}{c}{99.4}  &\multicolumn{1}{c}{99.4}  &\multicolumn{1}{c}{99.3}    \\
    \multicolumn{1}{c}{Bottle    }   &\multicolumn{1}{c}{98.3}  &\multicolumn{1}{c}{99.3}  &\multicolumn{1}{c}{99.4}  &\multicolumn{1}{c}{99.6}  &\multicolumn{1}{c}{99.9}  &\multicolumn{1}{c}{100.}  &\multicolumn{1}{c}{100.}  &\multicolumn{1}{c}{100.}  &\multicolumn{1}{c}{99.4}  &\multicolumn{1}{c}{99.6}  &\multicolumn{1}{c}{100.}  &\multicolumn{1}{c}{100.}  &\multicolumn{1}{c}{100.}  &\multicolumn{1}{c}{100.}    \\
    \multicolumn{1}{c}{Cable     }   &\multicolumn{1}{c}{42.7}  &\multicolumn{1}{c}{66.2}  &\multicolumn{1}{c}{91.6}  &\multicolumn{1}{c}{93.3}  &\multicolumn{1}{c}{95.3}  &\multicolumn{1}{c}{98.0}  &\multicolumn{1}{c}{98.3}  &\multicolumn{1}{c}{98.2}  &\multicolumn{1}{c}{93.8}  &\multicolumn{1}{c}{97.1}  &\multicolumn{1}{c}{98.8}  &\multicolumn{1}{c}{99.0}  &\multicolumn{1}{c}{99.2}  &\multicolumn{1}{c}{99.2}    \\
    \multicolumn{1}{c}{Capsule   }   &\multicolumn{1}{c}{42.5}  &\multicolumn{1}{c}{56.1}  &\multicolumn{1}{c}{85.4}  &\multicolumn{1}{c}{82.0}  &\multicolumn{1}{c}{89.1}  &\multicolumn{1}{c}{94.7}  &\multicolumn{1}{c}{97.9}  &\multicolumn{1}{c}{98.2}  &\multicolumn{1}{c}{82.4}  &\multicolumn{1}{c}{89.7}  &\multicolumn{1}{c}{97.8}  &\multicolumn{1}{c}{98.1}  &\multicolumn{1}{c}{99.0}  &\multicolumn{1}{c}{98.9}    \\
    \multicolumn{1}{c}{Hazelnut  }   &\multicolumn{1}{c}{87.6}  &\multicolumn{1}{c}{97.3}  &\multicolumn{1}{c}{99.1}  &\multicolumn{1}{c}{99.7}  &\multicolumn{1}{c}{99.8}  &\multicolumn{1}{c}{100.}  &\multicolumn{1}{c}{100.}  &\multicolumn{1}{c}{100.}  &\multicolumn{1}{c}{98.6}  &\multicolumn{1}{c}{99.9}  &\multicolumn{1}{c}{100.}  &\multicolumn{1}{c}{100.}  &\multicolumn{1}{c}{100.}  &\multicolumn{1}{c}{100.}    \\
    \multicolumn{1}{c}{Metal Nut }   &\multicolumn{1}{c}{32.1}  &\multicolumn{1}{c}{54.7}  &\multicolumn{1}{c}{88.5}  &\multicolumn{1}{c}{95.3}  &\multicolumn{1}{c}{97.0}  &\multicolumn{1}{c}{99.6}  &\multicolumn{1}{c}{99.7}  &\multicolumn{1}{c}{99.9}  &\multicolumn{1}{c}{94.4}  &\multicolumn{1}{c}{98.6}  &\multicolumn{1}{c}{99.8}  &\multicolumn{1}{c}{99.7}  &\multicolumn{1}{c}{100.}  &\multicolumn{1}{c}{100.}    \\
    \multicolumn{1}{c}{Pill      }   &\multicolumn{1}{c}{64.6}  &\multicolumn{1}{c}{84.3}  &\multicolumn{1}{c}{89.8}  &\multicolumn{1}{c}{92.0}  &\multicolumn{1}{c}{92.7}  &\multicolumn{1}{c}{95.8}  &\multicolumn{1}{c}{97.1}  &\multicolumn{1}{c}{97.3}  &\multicolumn{1}{c}{89.7}  &\multicolumn{1}{c}{93.7}  &\multicolumn{1}{c}{96.4}  &\multicolumn{1}{c}{96.0}  &\multicolumn{1}{c}{97.4}  &\multicolumn{1}{c}{97.3}    \\
    \multicolumn{1}{c}{Screw     }   &\multicolumn{1}{c}{39.4}  &\multicolumn{1}{c}{40.8}  &\multicolumn{1}{c}{49.5}  &\multicolumn{1}{c}{52.7}  &\multicolumn{1}{c}{57.8}  &\multicolumn{1}{c}{89.3}  &\multicolumn{1}{c}{93.9}  &\multicolumn{1}{c}{94.6}  &\multicolumn{1}{c}{35.9}  &\multicolumn{1}{c}{57.2}  &\multicolumn{1}{c}{88.4}  &\multicolumn{1}{c}{88.1}  &\multicolumn{1}{c}{94.7}  &\multicolumn{1}{c}{95.0}    \\
    \multicolumn{1}{c}{Toothbrush}   &\multicolumn{1}{c}{64.4}  &\multicolumn{1}{c}{61.1}  &\multicolumn{1}{c}{65.0}  &\multicolumn{1}{c}{72.2}  &\multicolumn{1}{c}{91.7}  &\multicolumn{1}{c}{75.6}  &\multicolumn{1}{c}{81.9}  &\multicolumn{1}{c}{92.2}  &\multicolumn{1}{c}{93.1}  &\multicolumn{1}{c}{93.1}  &\multicolumn{1}{c}{96.7}  &\multicolumn{1}{c}{96.4}  &\multicolumn{1}{c}{97.5}  &\multicolumn{1}{c}{97.2}    \\
    \multicolumn{1}{c}{Transistor}   &\multicolumn{1}{c}{58.7}  &\multicolumn{1}{c}{83.8}  &\multicolumn{1}{c}{88.6}  &\multicolumn{1}{c}{97.8}  &\multicolumn{1}{c}{96.4}  &\multicolumn{1}{c}{99.4}  &\multicolumn{1}{c}{99.5}  &\multicolumn{1}{c}{99.6}  &\multicolumn{1}{c}{99.4}  &\multicolumn{1}{c}{99.4}  &\multicolumn{1}{c}{99.8}  &\multicolumn{1}{c}{99.7}  &\multicolumn{1}{c}{99.8}  &\multicolumn{1}{c}{99.8}    \\
    \multicolumn{1}{c}{Zipper    }   &\multicolumn{1}{c}{46.1}  &\multicolumn{1}{c}{93.8}  &\multicolumn{1}{c}{99.1}  &\multicolumn{1}{c}{99.1}  &\multicolumn{1}{c}{99.2}  &\multicolumn{1}{c}{99.2}  &\multicolumn{1}{c}{99.6}  &\multicolumn{1}{c}{99.6}  &\multicolumn{1}{c}{98.7}  &\multicolumn{1}{c}{98.7}  &\multicolumn{1}{c}{99.5}  &\multicolumn{1}{c}{99.3}  &\multicolumn{1}{c}{99.6}  &\multicolumn{1}{c}{99.6}    \\

    \hline
    
    \multicolumn{1}{c}{Det.} &\multicolumn{1}{c}{70.5} &\multicolumn{1}{c}{82.1} &\multicolumn{1}{c}{90.3} &\multicolumn{1}{c}{92.2} &\multicolumn{1}{c}{94.5} &\multicolumn{1}{c}{96.7} &\multicolumn{1}{c}{97.8} &\multicolumn{1}{c}{98.6} &\multicolumn{1}{c}{92.1} &\multicolumn{1}{c}{95.0}  &\multicolumn{1}{c}{98.5}  &\multicolumn{1}{c}{98.3}  &\multicolumn{1}{c}{99.1}  &\multicolumn{1}{c}{99.0}   \\

    \hline
    \hline
\end{tabular}
\caption{Ablation study of the proposed WT-Flow on the MVTec AD dataset is presented with image-level AUROC per subclass. ``NC'': num of base channel in U-Net; ``WT map'' and ``pos'': wether to use the WT and positional embedding.}
\label{tab16}
\end{table*}

\begin{table*}[ht]
\centering
\footnotesize
\setlength{\arrayrulewidth}{1pt}
\begin{tabular}{c|c|c|c|c|c|c|c|c|c|c|c|c|c|c}
    \hline
    \hline
    WT map        &\multicolumn{5}{c|}{w/o}   &\multicolumn{3}{c|}{BatchNorm}                     &\multicolumn{6}{c}{LayerNorm}\\     
    
    \hline
    NC          &128    &256  &512  &\multicolumn{2}{c|}{768}   &512  &\multicolumn{2}{c|}{768}   &128    &256  &\multicolumn{2}{c|}{512}  &\multicolumn{2}{c}{768}    \\
    
    \hline
    pos         &\multicolumn{4}{c|}{}   &$\bm{\checkmark}$   &\multicolumn{2}{c|}{}  &$\bm{\checkmark}$  &\multicolumn{3}{c|}{}    &$\bm{\checkmark}$  & &$\bm{\checkmark}$   \\ 
    \hline
    \multicolumn{1}{c}{Carpet    }   &\multicolumn{1}{c}{99.0}  &\multicolumn{1}{c}{99.1}  &\multicolumn{1}{c}{99.1}  &\multicolumn{1}{c}{99.0}  &\multicolumn{1}{c}{98.6}  &\multicolumn{1}{c}{98.9}  &\multicolumn{1}{c}{98.9}  &\multicolumn{1}{c}{98.9}  &\multicolumn{1}{c}{98.9}  &\multicolumn{1}{c}{98.8}  &\multicolumn{1}{c}{98.6}  &\multicolumn{1}{c}{98.6}  &\multicolumn{1}{c}{99.0}  &\multicolumn{1}{c}{98.6}    \\
    \multicolumn{1}{c}{Grid      }   &\multicolumn{1}{c}{92.1}  &\multicolumn{1}{c}{96.7}  &\multicolumn{1}{c}{98.0}  &\multicolumn{1}{c}{98.1}  &\multicolumn{1}{c}{97.7}  &\multicolumn{1}{c}{98.1}  &\multicolumn{1}{c}{98.1}  &\multicolumn{1}{c}{98.2}  &\multicolumn{1}{c}{97.3}  &\multicolumn{1}{c}{97.9}  &\multicolumn{1}{c}{98.0}  &\multicolumn{1}{c}{98.2}  &\multicolumn{1}{c}{98.2}  &\multicolumn{1}{c}{98.2}    \\
    \multicolumn{1}{c}{Leather   }   &\multicolumn{1}{c}{98.8}  &\multicolumn{1}{c}{98.8}  &\multicolumn{1}{c}{98.9}  &\multicolumn{1}{c}{98.9}  &\multicolumn{1}{c}{98.6}  &\multicolumn{1}{c}{98.8}  &\multicolumn{1}{c}{98.8}  &\multicolumn{1}{c}{98.8}  &\multicolumn{1}{c}{98.8}  &\multicolumn{1}{c}{98.8}  &\multicolumn{1}{c}{98.7}  &\multicolumn{1}{c}{98.7}  &\multicolumn{1}{c}{98.9}  &\multicolumn{1}{c}{98.6}    \\
    \multicolumn{1}{c}{Tile      }   &\multicolumn{1}{c}{96.3}  &\multicolumn{1}{c}{96.1}  &\multicolumn{1}{c}{96.2}  &\multicolumn{1}{c}{96.3}  &\multicolumn{1}{c}{93.6}  &\multicolumn{1}{c}{96.0}  &\multicolumn{1}{c}{95.9}  &\multicolumn{1}{c}{95.9}  &\multicolumn{1}{c}{95.8}  &\multicolumn{1}{c}{95.8}  &\multicolumn{1}{c}{95.4}  &\multicolumn{1}{c}{95.3}  &\multicolumn{1}{c}{96.7}  &\multicolumn{1}{c}{94.9}    \\
    \multicolumn{1}{c}{Wood      }   &\multicolumn{1}{c}{86.8}  &\multicolumn{1}{c}{87.4}  &\multicolumn{1}{c}{92.8}  &\multicolumn{1}{c}{91.9}  &\multicolumn{1}{c}{91.5}  &\multicolumn{1}{c}{93.4}  &\multicolumn{1}{c}{93.0}  &\multicolumn{1}{c}{92.3}  &\multicolumn{1}{c}{93.5}  &\multicolumn{1}{c}{93.5}  &\multicolumn{1}{c}{93.7}  &\multicolumn{1}{c}{93.7}  &\multicolumn{1}{c}{93.8}  &\multicolumn{1}{c}{93.5}    \\
    \multicolumn{1}{c}{Bottle    }   &\multicolumn{1}{c}{94.6}  &\multicolumn{1}{c}{96.1}  &\multicolumn{1}{c}{97.4}  &\multicolumn{1}{c}{97.7}  &\multicolumn{1}{c}{97.7}  &\multicolumn{1}{c}{97.9}  &\multicolumn{1}{c}{98.0}  &\multicolumn{1}{c}{97.9}  &\multicolumn{1}{c}{97.7}  &\multicolumn{1}{c}{97.9}  &\multicolumn{1}{c}{97.8}  &\multicolumn{1}{c}{97.9}  &\multicolumn{1}{c}{97.9}  &\multicolumn{1}{c}{97.9}    \\
    \multicolumn{1}{c}{Cable     }   &\multicolumn{1}{c}{83.1}  &\multicolumn{1}{c}{90.1}  &\multicolumn{1}{c}{94.3}  &\multicolumn{1}{c}{94.7}  &\multicolumn{1}{c}{94.8}  &\multicolumn{1}{c}{96.8}  &\multicolumn{1}{c}{97.1}  &\multicolumn{1}{c}{97.0}  &\multicolumn{1}{c}{95.4}  &\multicolumn{1}{c}{97.1}  &\multicolumn{1}{c}{97.9}  &\multicolumn{1}{c}{98.1}  &\multicolumn{1}{c}{98.2}  &\multicolumn{1}{c}{98.1}    \\
    \multicolumn{1}{c}{Capsule   }   &\multicolumn{1}{c}{92.2}  &\multicolumn{1}{c}{94.2}  &\multicolumn{1}{c}{97.4}  &\multicolumn{1}{c}{97.6}  &\multicolumn{1}{c}{98.1}  &\multicolumn{1}{c}{98.5}  &\multicolumn{1}{c}{98.5}  &\multicolumn{1}{c}{98.5}  &\multicolumn{1}{c}{97.8}  &\multicolumn{1}{c}{98.3}  &\multicolumn{1}{c}{98.5}  &\multicolumn{1}{c}{98.6}  &\multicolumn{1}{c}{98.8}  &\multicolumn{1}{c}{98.7}    \\
    \multicolumn{1}{c}{Hazelnut  }   &\multicolumn{1}{c}{93.8}  &\multicolumn{1}{c}{96.3}  &\multicolumn{1}{c}{97.0}  &\multicolumn{1}{c}{97.4}  &\multicolumn{1}{c}{97.4}  &\multicolumn{1}{c}{98.2}  &\multicolumn{1}{c}{98.1}  &\multicolumn{1}{c}{98.2}  &\multicolumn{1}{c}{97.5}  &\multicolumn{1}{c}{98.3}  &\multicolumn{1}{c}{98.3}  &\multicolumn{1}{c}{98.4}  &\multicolumn{1}{c}{98.4}  &\multicolumn{1}{c}{98.3}    \\
    \multicolumn{1}{c}{Metal Nut }   &\multicolumn{1}{c}{74.5}  &\multicolumn{1}{c}{82.2}  &\multicolumn{1}{c}{92.0}  &\multicolumn{1}{c}{94.2}  &\multicolumn{1}{c}{93.9}  &\multicolumn{1}{c}{97.1}  &\multicolumn{1}{c}{96.7}  &\multicolumn{1}{c}{96.7}  &\multicolumn{1}{c}{95.0}  &\multicolumn{1}{c}{96.9}  &\multicolumn{1}{c}{97.1}  &\multicolumn{1}{c}{97.1}  &\multicolumn{1}{c}{97.3}  &\multicolumn{1}{c}{97.1}    \\
    \multicolumn{1}{c}{Pill      }   &\multicolumn{1}{c}{88.6}  &\multicolumn{1}{c}{94.5}  &\multicolumn{1}{c}{95.9}  &\multicolumn{1}{c}{96.8}  &\multicolumn{1}{c}{95.9}  &\multicolumn{1}{c}{98.0}  &\multicolumn{1}{c}{98.1}  &\multicolumn{1}{c}{98.2}  &\multicolumn{1}{c}{96.9}  &\multicolumn{1}{c}{97.7}  &\multicolumn{1}{c}{98.4}  &\multicolumn{1}{c}{98.3}  &\multicolumn{1}{c}{98.6}  &\multicolumn{1}{c}{98.5}    \\
    \multicolumn{1}{c}{Screw     }   &\multicolumn{1}{c}{90.9}  &\multicolumn{1}{c}{91.8}  &\multicolumn{1}{c}{93.8}  &\multicolumn{1}{c}{94.8}  &\multicolumn{1}{c}{95.9}  &\multicolumn{1}{c}{98.3}  &\multicolumn{1}{c}{98.7}  &\multicolumn{1}{c}{98.8}  &\multicolumn{1}{c}{93.1}  &\multicolumn{1}{c}{96.5}  &\multicolumn{1}{c}{98.5}  &\multicolumn{1}{c}{98.4}  &\multicolumn{1}{c}{98.8}  &\multicolumn{1}{c}{98.8}    \\
    \multicolumn{1}{c}{Toothbrush}   &\multicolumn{1}{c}{89.8}  &\multicolumn{1}{c}{93.9}  &\multicolumn{1}{c}{95.8}  &\multicolumn{1}{c}{97.3}  &\multicolumn{1}{c}{98.6}  &\multicolumn{1}{c}{97.7}  &\multicolumn{1}{c}{98.1}  &\multicolumn{1}{c}{98.7}  &\multicolumn{1}{c}{98.6}  &\multicolumn{1}{c}{98.6}  &\multicolumn{1}{c}{98.8}  &\multicolumn{1}{c}{98.8}  &\multicolumn{1}{c}{98.8}  &\multicolumn{1}{c}{98.8}    \\
    \multicolumn{1}{c}{Transistor}   &\multicolumn{1}{c}{84.1}  &\multicolumn{1}{c}{83.8}  &\multicolumn{1}{c}{88.0}  &\multicolumn{1}{c}{98.0}  &\multicolumn{1}{c}{98.3}  &\multicolumn{1}{c}{98.3}  &\multicolumn{1}{c}{98.3}  &\multicolumn{1}{c}{98.2}  &\multicolumn{1}{c}{98.1}  &\multicolumn{1}{c}{98.1}  &\multicolumn{1}{c}{98.2}  &\multicolumn{1}{c}{98.1}  &\multicolumn{1}{c}{98.2}  &\multicolumn{1}{c}{98.1}    \\
    \multicolumn{1}{c}{Zipper    }   &\multicolumn{1}{c}{83.3}  &\multicolumn{1}{c}{91.7}  &\multicolumn{1}{c}{98.2}  &\multicolumn{1}{c}{98.2}  &\multicolumn{1}{c}{98.0}  &\multicolumn{1}{c}{98.2}  &\multicolumn{1}{c}{98.3}  &\multicolumn{1}{c}{98.3}  &\multicolumn{1}{c}{98.2}  &\multicolumn{1}{c}{98.2}  &\multicolumn{1}{c}{98.4}  &\multicolumn{1}{c}{98.3}  &\multicolumn{1}{c}{98.2}  &\multicolumn{1}{c}{98.2}    \\

    \hline
    \multicolumn{1}{c}{Loc.}       &\multicolumn{1}{c}{89.9}   &\multicolumn{1}{c}{92.8}  &\multicolumn{1}{c}{95.7}  &\multicolumn{1}{c}{96.7}  &\multicolumn{1}{c}{96.6}  &\multicolumn{1}{c}{97.6}  &\multicolumn{1}{c}{97.6}  &\multicolumn{1}{c}{97.6}  &\multicolumn{1}{c}{96.8}  &\multicolumn{1}{c}{97.5}  &\multicolumn{1}{c}{97.7}  &\multicolumn{1}{c}{97.8}  &\multicolumn{1}{c}{98.0}  &\multicolumn{1}{c}{97.8}   \\

    \hline
    \hline
\end{tabular}
\caption{Ablation study of the proposed WT-Flow on the MVTec AD dataset is presented with pixel-level AUROC per subclass. ``NC'': num of base channel in U-Net; ``WT map'' and ``pos'': wether to use the WT and positional embedding.}
\label{tab17}
\end{table*}

\end{document}